\documentclass[11pt]{article}

\usepackage[preprint]{acl}

\usepackage{times}
\usepackage{latexsym}
\usepackage{amsmath}
\usepackage[table]{xcolor}
\usepackage{booktabs,longtable,array}
\usepackage[T1]{fontenc}
\usepackage[utf8]{inputenc}

\usepackage{microtype}

\usepackage{inconsolata}

\usepackage{graphicx}
\usepackage{algorithm}
\usepackage{algpseudocode}
\usepackage{listings}
\usepackage{booktabs}
\usepackage{multirow}
\usepackage{pifont}
\usepackage{svg}

\title{Beyond Prompt-Based Planning: MCP‑Native Graph Planning-based Biomedical Agent System}

\author{
  \textbf{Zhangtianyi Chen\textsuperscript{1}}\thanks{~~Equal contribution.},
  \textbf{Florensia Widjaja\textsuperscript{1}}\footnotemark[1],
  \textbf{Wufei Dai\textsuperscript{1}},
  \textbf{Xiangjun Zhang\textsuperscript{1}},
  \textbf{Yuhao Shen\textsuperscript{1}},
  \textbf{Juexiao Zhou\textsuperscript{1}}\thanks{~~Correspondence: \href{mailto:juexiao.zhou@gmail.com}{juexiao.zhou@gmail.com}},
\\
\\
\textsuperscript{1}The Chinese University of Hong Kong, Shenzhen
}
\usepackage[most]{tcolorbox}
\usepackage{fvextra}
\newtcolorbox{promptbox}[1]{
    breakable,
    colback=gray!3,
    colframe=gray!55,
    coltitle=black,
    fonttitle=\bfseries,
    title=#1,
    boxrule=0.6pt,
    arc=2pt,
    left=4pt,
    right=4pt,
    top=4pt,
    bottom=4pt
}
\begin{document}
\maketitle
\begin{abstract}
Biomedical agents promise to automate complex biological workflows, yet current systems face two fundamental bottlenecks: bioinformatics tools are highly heterogeneous in interfaces and execution environments, while agent planning still relies on flat prompt-retrieved tool descriptions. As biomedical software ecosystems grow, this coupling between tool coverage and context size leads to tool confusion, unstable planning, and inefficient execution. We introduce \textbf{BioManus}, an MCP-native biomedical agent built on \emph{graph-scaffolded planning} over structured biological capabilities. BioManus first introduces the \textbf{BioinfoMCP Compiler}, which converts heterogeneous bioinformatics software into standardized MCP servers, yielding a large executable MCP ecosystem. It then organizes this ecosystem as a typed heterogeneous MCP graph over tools, operations, datatypes, and workflow stages. At inference time, BioManus retrieves compact task-specific subgraphs, synthesizes operation-level workflow scaffolds. This design decouples planning complexity from raw tool inventory size, achieving a context compression ratio of \(\Theta\!\left(N/(h\bar{m})\right)\) under high-recall retrieval, where \(N\) is the total tool count, \(h\) is the workflow horizon, and \(\bar{m}\ll N\) is the average number of candidate tools per operation. Experiments on BioAgentBench and LAB-Bench show that BioManus improves execution accuracy, workflow validity, and context efficiency over advanced biomedical agent baselines. This work suggests a paradigm shift: scalable biomedical reasoning requires structured executable capability graphs rather than increasingly larger prompt-level tool retrieval.
\end{abstract}

\begin{figure}[!htb]
    \centering
    \includegraphics[width=0.5\textwidth]{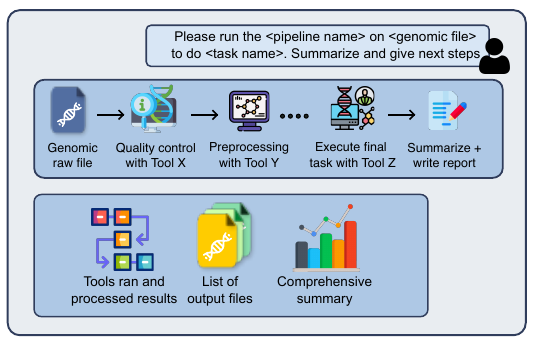}
    \caption{Staged genomic data processing pipeline: Triggered by simple instructions, it automatically completes quality control, preprocessing, core analysis, and report generation}
    \label{fig:intro_pipeline}
\end{figure}
\section{Introduction}
Foundation-model agents are reshaping computational biology, promising a future in which researchers express scientific intent in natural language and agents autonomously retrieve tools, compose workflows, and synthesize results~\cite{yao2022react,qin2023toolllmfacilitatinglargelanguage, patil2023gorillalargelanguagemodel,Jin_2024, wang2024geneagentselfverificationlanguageagent, xiao2024cellagentllmdrivenmultiagentframework, zhou2024ai,pickard2025automatic,wang2025spatialagent}. Yet as this vision scales, a structural bottleneck has become unavoidable: \emph{the dominant paradigm of prompt-based tool retrieval does not scale with the biomedical action space.} State-of-the-art biomedical systems such as Biomni~\cite{huang2025biomni} integrate large collections of tools, databases, and software into agentic workflows, while general tool-use agents often rely on retrieving API or tool documentation into the model context before planning over candidate actions~\cite{qin2023toolllmfacilitatinglargelanguage, patil2023gorillalargelanguagemodel, wang2025mcpbenchbenchmarkingtoolusingllm}. This works for small tool inventories, but in biology, where new command-line programs, R/Bioconductor packages, and containerized pipelines emerge continuously~\cite{leipzig2017review,gruning2018bioconda, wratten2021reproducible,reiter2021streamlining}, recall demands ever-larger retrieval windows, while context dilution, tool confusion, and planning instability grow correspondingly~\cite{liu-etal-2024-lost, li2026complexmcp}. The agent is asked to do combinatorial search in a flat semantic space whose size grows linearly with the field.

As shown in Figure~\ref{fig:intro_pipeline}, biological analysis naturally decomposes into staged workflows involving preprocessing, quality control, downstream analysis, and iterative summarization. Each step exchanges typed biological data, and valid execution requires respecting compatibility constraints across operations, formats, and stages. Modern bioinformatics ecosystems already encode this structure through workflow systems such as CWL, Galaxy, Snakemake, and Nextflow~\cite{amstutz2016common,goecks2010galaxy, koster2012snakemake,di2017nextflow,wratten2021reproducible}. Yet current LLM agents largely discard this rich execution structure, reducing tools to interchangeable text snippets retrieved into prompts.

We therefore argue that the core bottleneck is not retrieval, but \emph{representation}. Biomedical tools do not exist in a flat semantic space: they implement structured operations over typed inputs and outputs, compose through constrained execution dependencies, and form implicit workflow graphs.

\paragraph{Contributions.} We introduce \textbf{BioManus}, an MCP-native biomedical agent built around graph-scaffolded planning, a new paradigm in which the agent reasons over an explicit typed capability graph instead of unstructured prompt-retrieved tool descriptions. Two components make this possible.

First, the \textbf{BioinfoMCP Compiler} automatically lifts heterogeneous bioinformatics tools, including CLIs, Python packages, R libraries, and containerized workflows, into standardized MCP servers with unified interface contracts. This produces an executable ecosystem of \textbf{910 MCP servers and 3{,}500 callable tools spanning eight biological domains}, substantially larger than prior biomedical agent environments.

Second, BioManus constructs a \textbf{typed capability graph} whose nodes represent servers, tools, operations, datatypes, and workflow stages, while edges encode hosting relations, execution dependencies, datatype consumption/production, and stage transitions. At inference time, BioManus does not expose the full tool inventory to the LLM. Instead, it retrieves a compact, task-specific subgraph, synthesizes biologically valid operation paths, binds them to executable MCP tools, and dynamically registers only the required servers for execution.

This reframes the interface between LLMs and biomedical software ecosystems.
Instead of scaling with retrieved tool descriptions, BioManus performs
structured graph reasoning over executable biological capabilities, leading to
more efficient planning, more valid workflows, and stronger performance on
large-scale biomedical agent benchmarks.
\section{Related Work}

\paragraph{Agent systems for end-to-end bioinformatics analysis.}
Recent biomedical foundation-model agents have explored end-to-end biological
workflow automation through tool orchestration, database interaction, and
multi-step scientific reasoning. Systems such as
GeneAgent~\cite{wang2024geneagentselfverificationlanguageagent},
CellAgent~\cite{xiao2024cellagentllmdrivenmultiagentframework},
SpatialAgent~\cite{wang2025spatialagent}, and
Biomni~\cite{huang2025biomni} demonstrate that LLM-based agents can assist
complex bioinformatics analysis pipelines by integrating biological databases,
analytical software, and iterative reasoning. Concurrently, recent biomedical workflow and orchestration systems such as
PromptBio~\cite{zhang2025promptbio},
BioAgents~\cite{mehandru2025bioagents},
STELLA~\cite{jin2025stella}, and
PoSyMed~\cite{suwer2026biomedical} have increasingly emphasized scalable,
reproducible, and agent-driven execution over heterogeneous bioinformatics
ecosystems. These systems highlight the growing transition from isolated tool
execution toward integrated orchestration of large-scale computational biology
workflows.

\paragraph{Tool-using in agentic workflows for bioinformatics analysis.}
Recent advances in tool-using AI agents have enabled LLM systems to iteratively
interact with external APIs, software environments, and computational
tools~\cite{yao2022react,qin2023toolllmfacilitatinglargelanguage,
patil2023gorillalargelanguagemodel}. In bioinformatics analysis, these
paradigms are increasingly adopted to orchestrate biological databases,
computational pipelines, and heterogeneous analytical software within unified
agentic workflows~\cite{wang2024geneagentselfverificationlanguageagent,
xiao2024cellagentllmdrivenmultiagentframework,
wang2025spatialagent,
huang2025biomni,
zhou2024ai,
pickard2025automatic}. Existing systems typically retrieve candidate tools,
software APIs, or workflow descriptions directly into the model context and
perform planning over textual tool representations~\cite{qin2023toolllmfacilitatinglargelanguage,
patil2023gorillalargelanguagemodel,
huang2025biomni,
wang2025mcpbenchbenchmarkingtoolusingllm}. While effective in relatively small
tool spaces, this prompt-based retrieval paradigm becomes increasingly unstable
in large biomedical ecosystems due to context dilution, retrieval ambiguity,
and combinatorial planning complexity~\cite{liu-etal-2024-lost,
li2026complexmcp}. 
\section{Problem Formulation and Diagnosis}

\subsection{Prompt-based Tool Retrieval in Biomedical Agents}

Recent tool-augmented large language model (LLM) agents for biological analysis
increasingly adopt a \emph{prompt-based tool retrieval} paradigm, in which the
agent first retrieves a subset of candidate tools or APIs and then performs
planning directly over their textual descriptions. Let

\begin{equation}
\mathcal{T}=\{t_1,t_2,\dots,t_N\}
\end{equation}

denote the complete biomedical tool ecosystem and let \(q\) denote a user
query. Existing systems can be abstracted as:

\begin{equation}
q
\rightarrow
\mathcal{T}_k
\rightarrow
P(q,\mathcal{T}_k)
\rightarrow
\Pi
\end{equation}

where \(\mathcal{T}_k \subset \mathcal{T}\) denotes the retrieved candidate
tools, \(P(q,\mathcal{T}_k)\) denotes the constructed planning prompt, and
\(\Pi\) denotes the final execution trajectory generated by the LLM.

This paradigm underlies many modern tool-augmented agents, including
Toolformer~\cite{schick2023toolformerlanguagemodelsteach},
Gorilla~\cite{patil2023gorillalargelanguagemodel},
CodeAct~\cite{wang2024executablecodeactionselicit},
and biomedical systems such as Biomni. 

\subsection{Scalability Diagnosis}

\paragraph{Problem 1: Context Scaling in Large Tool Ecosystems.}

Prompt-based retrieval directly injects retrieved tool descriptions into the
reasoning context of the LLM. Consequently, planning complexity grows together
with the number of retrieved candidate tools. Let \(\ell_t\) denote the average
token length of a tool description and let \(k\) denote the number of retrieved
tools. The resulting planning context can be approximated as:

\begin{equation}
C_{\text{prompt}}
=
C_q
+
k\ell_t
+
C_{\text{instruction}}
\label{eq:prompt_context_consuming}
\end{equation}

where \(C_q\) denotes user-query tokens and
\(C_{\text{instruction}}\) denotes system and planning instructions.
Therefore,

\begin{equation}
C_{\text{prompt}} = O(k\ell_t)
\end{equation}

As biomedical tool ecosystems continue expanding, maintaining retrieval recall
often requires retrieving increasingly larger candidate tool sets. This creates
a direct coupling between ecosystem scale and planning-context size, leading to
higher inference cost, increased context pollution, and greater planning
instability in long-context reasoning.

\paragraph{Problem 2: Tool Heterogeneity and Execution Fragmentation.}

Biomedical software ecosystems are inherently heterogeneous. Modern biological
analysis pipelines span command-line tools, Python frameworks,
R/Bioconductor packages, workflow systems, database APIs, and containerized
execution environments. These tools operate across incompatible dependency
stacks, runtime environments, and interface conventions.

As a result, agent systems must simultaneously solve not only high-level
reasoning and planning, but also low-level execution orchestration problems,
including dependency conflicts, package installation failures, runtime
incompatibilities, and cross-language environment management. Existing
prompt-based agents typically expose these execution burdens directly to the
LLM, forcing the model to reason over fragmented software environments without
an explicit representation of execution structure or compatibility constraints.

\subsection{Empirical Observations of Context Scaling}

\begin{figure}[!htb]
    \includegraphics[width=0.5\textwidth]{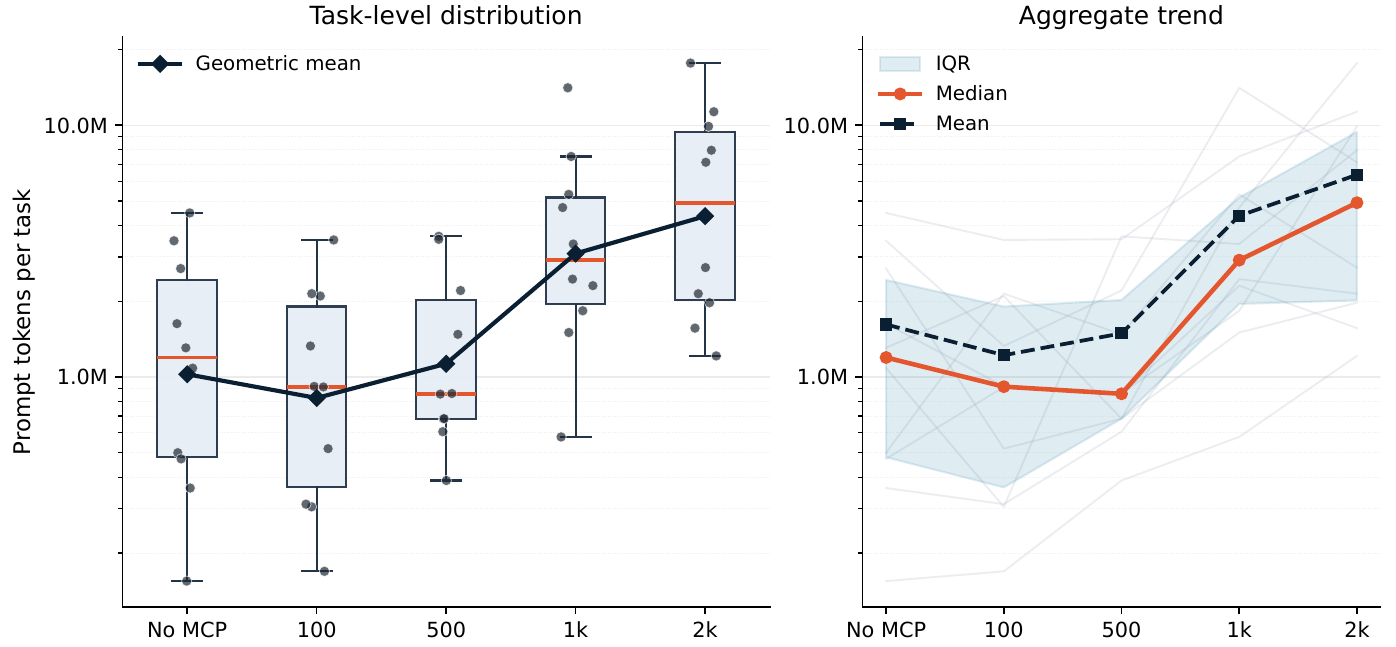}
    \caption{
    Prompt context growth with biomedical tool scale. Each scale contains 10
    BioAgentBench tasks. Points denote individual tasks, boxes summarize quartiles,
    and lines report aggregate trends.
    }
    \label{fig:context_growth}
\end{figure}

To empirically evaluate the scalability of prompt-based retrieval, we measure
the prompt-token consumption of Biomni under increasing
MCP tool inventory sizes. We include a No-MCP setting and randomly sample
\(100\), \(500\), \(1{,}000\), and \(2{,}000\) tools from
Bioconda~\cite{gruning2018bioconda}. For each setting, we run the same
10 BioAgentBench tasks~\cite{fa2026bioagentbenchaiagent} and record the
average prompt tokens consumed during planning.

Figure~\ref{fig:context_growth} shows both task-level distributions and
aggregate trends. Three observations emerge. First, prompt-based retrieval can
reduce context consumption in small tool spaces: moving from No-MCP to
100 tools slightly decreases prompt usage by filtering irrelevant actions.
Second, this benefit quickly diminishes as the tool ecosystem grows. From
No-MCP to \(2{,}000\) tools, the geometric mean of prompt tokens increases by
\(4.2\times\), indicating rapidly growing context overhead under large-scale
retrieval. Third, larger tool inventories also introduce substantially higher
variance and long-tail behavior, with several tasks exceeding \(10^7\) prompt
tokens.
\begin{figure*}[!htb]
    \centering
    \includegraphics[width=\textwidth]{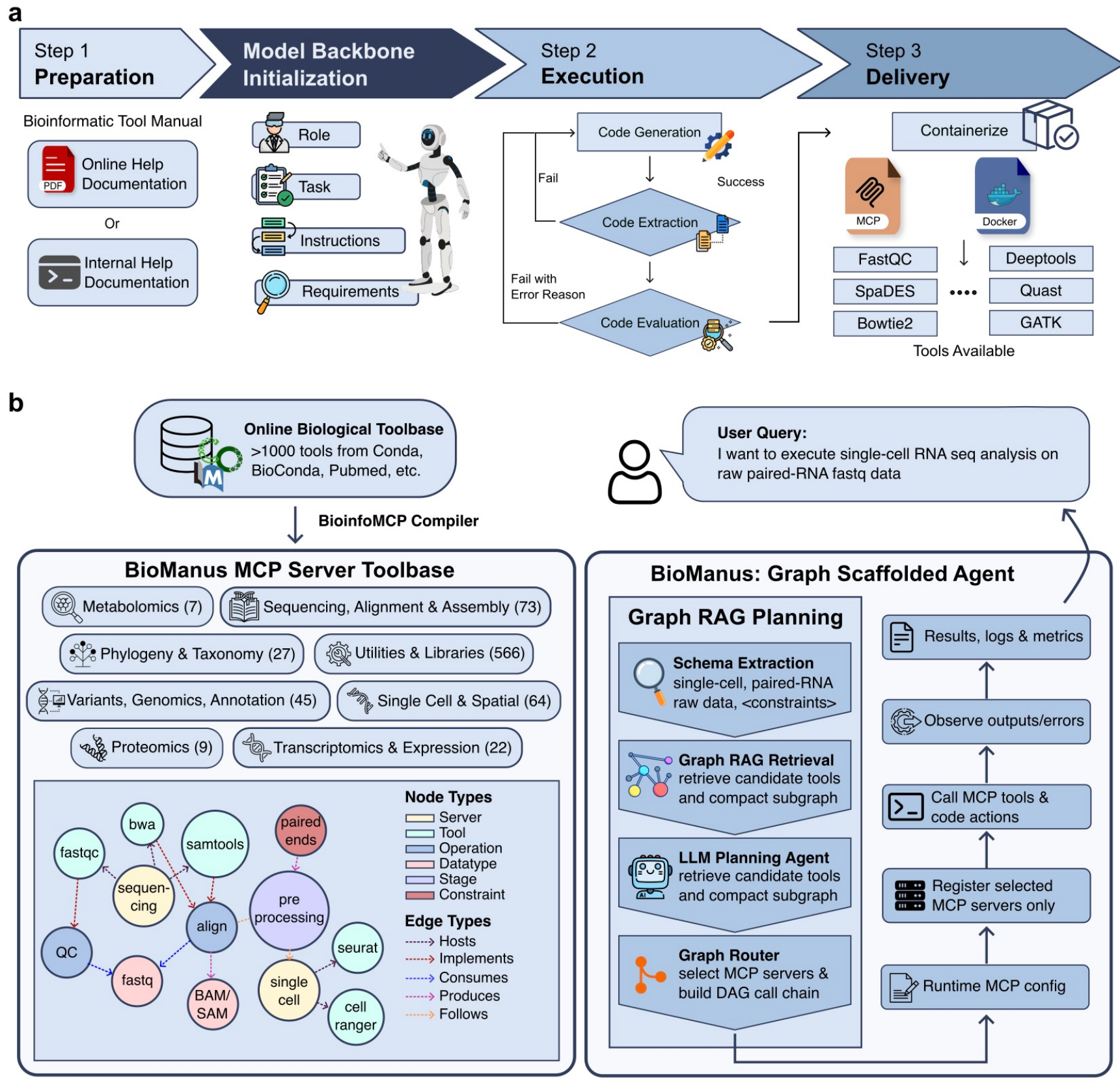}
    \caption{
\textbf{Overview of BioManus.}
(a) \textbf{BioinfoMCP Compiler} automatically converts heterogeneous
bioinformatics resources into standardized MCP servers
with unified callable interfaces and executable schemas.
(b) \textbf{BioManus} organizes the generated MCP ecosystem into a typed
capability graph and performs graph-scaffolded planning
}
\label{fig:biomanus_overview}
    \label{fig:BioManus}
\end{figure*}
Together, these results support our central diagnosis: prompt-based retrieval
is effective at small scales, but becomes increasingly unstable as biomedical
tool ecosystems expand because retrieval recall and planning-context size
remain tightly coupled.

\section{Methodology}

\subsection{Overview of BioManus}
BioManus is a graph-scaffolded, MCP-native biomedical agent designed to address
the scalability limitations of prompt-based biomedical agents. As illustrated
in Figure~\ref{fig:biomanus_overview}, the system consists of four components:
MCP conversion, typed capability graph construction, graph-scaffolded planning,
and dynamic execution. First, the \textbf{BioinfoMCP Compiler} converts
heterogeneous bioinformatics resources into standardized MCP servers with
unified callable interfaces (Appendix~\ref{app:mcp_server_catalog}). These
servers are organized into a heterogeneous \textbf{Typed Capability Graph}
whose nodes represent tools, operations, datatypes, and workflow stages.
Given a natural-language biological query, BioManus extracts structured task
semantics and performs GraphRAG-based retrieval over the global MCP graph to
obtain a compact task-specific subgraph together with candidate tools and
operation-path hints. An LLM-based planner then synthesizes an operation-level
workflow scaffold, after which only the selected MCP servers are dynamically
registered for execution. By reasoning over compact graph-scaffolded workflows
rather than flat prompt-retrieved tool descriptions, BioManus improves
scalability, context efficiency, and workflow validity in large biomedical
software ecosystems.
\subsection{BioinfoMCP Compiler: Automated Tool-to-MCP Conversion}
The BioinfoMCP Compiler is a framework that automatically converts heterogeneous bioinformatics tools into executable MCP servers through a three-stage pipeline: preparation, execution, and delivery. In the preparation stage, the compiler harvests tool documentation and command-line interfaces from diverse sources, including manuals, PDF documentation, and help flags, and extracts the semantic and execution specifications required for downstream code generation. In the execution stage, an LLM backbone synthesizes MCP server implementations directly from the extracted specifications; the generated code undergoes automatic parsing and validation for structural correctness, encompassing Python syntax checking and MCP interface compliance, with failed generations iteratively refined until a valid executable server is produced. Finally, during delivery, the resulting MCP server is packaged along with dependency specifications and Docker-based execution environments as complementary runtime artifacts. Such packaging enables portable and reproducible deployment across heterogeneous computing infrastructure. The overall framework is illustrated in Figure~\ref{fig:BioManus}(a), with detailed Chain-Of-Thought provided in the Appendix (Figure \ref{fig:rtir}).

\subsection{Biomedical Tool-Capability Graph}

BioManus organizes the MCP ecosystem as a heterogeneous typed graph
\(\mathcal{G}=(\mathcal{V},\mathcal{E})\), which serves as the structural
backbone for graph-scaffolded planning. Rather than exposing raw tool
descriptions directly to the LLM, the graph explicitly models the semantic and
execution relationships among tools, operations, datatypes, workflow stages,
and MCP servers.

\paragraph{Node and Edge Types.}
The graph contains six node categories:
tool nodes (\(\mathcal{V}_T\)),
operation nodes (\(\mathcal{V}_O\)),
datatype nodes (\(\mathcal{V}_D\)),
capability nodes (\(\mathcal{V}_C\)),
stage nodes (\(\mathcal{V}_S\)),
and MCP server nodes (\(\mathcal{V}_R\)).
Operation nodes abstract over semantically equivalent tools, while datatype and
stage nodes encode workflow compatibility and execution order priors.

Edges represent executable and semantic relations, including:
\texttt{implements} (operation \(\rightarrow\) tool),
\texttt{consumes}/\texttt{produces} (datatype \(\leftrightarrow\) operation),
\texttt{hosts} (server \(\rightarrow\) tool),
and soft workflow-transition edges between compatible operations.
Together, these relations transform the biomedical tool ecosystem into an
executable graph over typed biological transformations.

\paragraph{Graph Construction.}
The graph is constructed automatically from the generated MCP ecosystem.
For each MCP server, BioManus parses tool schemas, parameter specifications,
and natural-language descriptions. An LLM-based annotator then infers:
(i) input/output datatypes,
(ii) semantic operation categories,
and (iii) workflow-stage annotations.
Tools with similar semantic functions are merged into shared operation nodes,
while datatype-compatible operations are connected through soft workflow edges.
The resulting graph encodes both executable tool bindings and biologically
plausible workflow structure. The detailed statistics of the typed capability graph are displayed in the Appendix~\ref{app:graph_statistics}.

\subsection{Graph-Scaffolded Workflow Planning}

Given a natural-language biological query \(q\), BioManus first converts the
query into a structured task specification containing biological entities,
capabilities, datatypes, and workflow constraints. These semantic anchors are
projected onto the typed capability graph through GraphRAG-style retrieval.

Starting from inferred input datatypes and task objectives, the retriever
extracts a compact task-specific subgraph together with candidate operations
and workflow-path hints. Importantly, retrieval occurs at the \emph{operation
level} rather than the raw-tool level: the planner first reasons over abstract
biological transformations and only later binds them to executable MCP tools.

Based on the retrieved graph context, an LLM planner synthesizes an
operation-level execution route
\begin{equation}
P=(o_1,o_2,\dots,o_h),
\end{equation}
where each operation \(o_i\) is associated with a compact candidate tool set.
Only the MCP servers involved in the selected workflow are dynamically
registered and executed at runtime.

This graph-scaffolded design decouples workflow planning from global tool
inventory size by retrieving compact executable subgraphs instead of large
collections of raw tool descriptions.
\subsection{Complexity Analysis of Graph-Scaffolded Planning}

Most existing biomedical agents rely on prompt-based retrieval, where the LLM
plans directly over retrieved textual tool descriptions. Let
\begin{equation}
\mathcal{T}=\{t_1,t_2,\dots,t_N\}
\end{equation}
denote the global tool ecosystem, and let \(k\) be the number of retrieved
candidate tools. If \(\ell_t\) denotes the average tool-description length,
the planning context scales approximately as
\begin{equation}
C_{\text{prompt}} = O(k\ell_t).
\end{equation}

As the biomedical tool ecosystem grows, maintaining retrieval recall typically
requires increasing \(k\), causing planning context, inference cost, and tool
confusion to grow correspondingly.

In contrast, BioManus performs retrieval at the operation level. Given an
operation path
\begin{equation}
P=(o_1,o_2,\dots,o_h),
\end{equation}
each operation \(o_i\) is associated with only a compact local candidate set
\(\mathcal{T}(o_i)\). Let \(\bar{m}\) denote the average number of candidate
tools per operation. The graph-scaffolded planning context therefore scales as
\begin{equation}
C_{\text{graph}}
=
O\bigl(h(\ell_o+\bar{m}\ell_s)\bigr),
\end{equation}
where \(h\) is the workflow horizon, \(\ell_o\) is the operation-scaffold
length, and \(\ell_s\) is the average selected-tool schema length.

Crucially, \(C_{\text{graph}}\) depends primarily on workflow complexity rather
than the total ecosystem size \(N\). In high-recall settings where
\(k=\Theta(N)\), prompt-based planning grows linearly with the global tool
inventory, whereas graph-scaffolded planning grows only with the number of
workflow-relevant operations and their local executable neighborhoods. This
enables substantially more scalable and stable planning in large biomedical
tool ecosystems.

\section{Experiments}

\subsection{Experiment Setup}

\paragraph{Benchmarks.}
We evaluate BioManus on two representative biomedical agent benchmarks.
LAB-Bench~\cite{laurent2024labbenchmeasuringcapabilitieslanguage}
evaluates biological analysis and tool use through DbQA (database question
answering), SeqQA (sequence question answering) and CloningScenarios (molecular cloning scenario planning). BioAgentBench~\cite{fa2026bioagentbenchaiagent}
evaluates realistic end-to-end bioinformatics workflows

\paragraph{Baselines.}
We compare BioManus against both general-purpose tool-use agents and
biomedical foundation agents, including
ReAct-Code,
Biomni, Biomni-ReAct, and the base
LLM backbone DeepSeek-V4~\cite{deepseekai2026deepseekv4}. To analyze the scalability of prompt-based retrieval, we additionally
evaluate Biomni under progressively larger tool inventories, including
100, 500, 1k, and 2k. All methods use the DeepSeek-V4 backbone and execution environment.

\paragraph{Implementation details.}
All methods use a unified prompting template and experimental setting.
Additional implementation details and generation settings are provided in the Appendix~\ref{app:implementation}; the prompt templates are provided in the Appendix~\ref{app:prompts}.
\subsection{Benchmarking BioinfoMCP Compiler: Performance in Tool Onboarding}
To evaluate the scalability and robustness of BioinfoMCP Compiler, we first
analyze the resulting MCP ecosystem. As summarized in
Table~\ref{tab:compiler_ecosystem}(A), BioinfoMCP converts heterogeneous
bioinformatics software into an ecosystem spanning 910 MCP servers and  3,500 executable MCP tools across diverse biological domains, including
genomics, transcriptomics, metagenomics, and biomedical databases.

We then compare BioinfoMCP Compiler against direct prompting through 3,041
paired MCP-generation evaluations. Table~\ref{tab:compiler_ecosystem}(B)
reports three metrics: \textit{Structured Return}, which measures whether the
generated output conforms to a valid executable MCP interface;
\textit{Multi-Tool Rate}, which measures the fraction of MCP servers exposing
multiple callable tools; and \textit{Generation Tokens}, which measures average
generation cost. Compared with direct prompting, BioinfoMCP substantially
improves structured MCP generation (\(77.3\%\rightarrow99.4\%\)) and
multi-tool generation (\(10.4\%\rightarrow63.7\%\)), indicating that the
compiler not only improves robustness but also exposes richer executable
capabilities. Although generation cost increases moderately
(\(5{,}380\rightarrow7{,}306\) tokens), the resulting MCP servers are
substantially more reliable and deployable.
\begin{table}[t]
\centering
\scriptsize
\setlength{\tabcolsep}{3.5pt}
\renewcommand{\arraystretch}{1.06}

\caption{
Statistics and robustness analysis of the BioinfoMCP ecosystem and compiler.
}

\label{tab:compiler_ecosystem}

\begin{tabular}{lccc}
\toprule

\multicolumn{4}{c}{\textbf{(A) BioinfoMCP Ecosystem}} \\

\midrule

Domain
&
\# Servers
&
\# Tools
&
\\

\midrule

Genomics & 142 & 621 & \\
Transcriptomics & 128 & 544 & \\
Metagenomics & 84 & 337 & \\
Single-cell Biology & 73 & 286 & \\
Proteomics & 65 & 249 & \\
Evolutionary Biology & 58 & 211 & \\
Biomedical Databases & 149 & 603 & \\
Workflow Utilities & 211 & 649 & \\

\midrule

Total
&
\textbf{910}
&
\textbf{3,500}
&
\\

\midrule

\multicolumn{4}{c}{\textbf{(B) Compiler Robustness}} \\

\midrule

Metric
&
Prompting
&
Compiler
&
\\

\midrule

Structured Return$\uparrow$
&
77.3\%
&
\textbf{99.4\%}
{\scriptsize \textcolor{green!60!black}{$\uparrow$22.1}}
&
\\

Generation Tokens$\downarrow$
&
5,380
&
7,306
{\scriptsize \textcolor{red!70!black}{$\uparrow$35.8\%}}
&
\\

Multi-Tool Rate$\uparrow$
&
10.4\%
&
\textbf{63.7\%}
{\scriptsize \textcolor{green!60!black}{$\uparrow$53.3}}
&
\\

\midrule

\multicolumn{4}{c}{\textbf{(C) Backbone Generalization}} \\

\midrule

Backbone
&
Baseline Parse Success$\uparrow$
&
Compiler Parse Success$\uparrow$
\\

\midrule

Gemini 3.1
&
99.5\%
&
99.5\%
{\scriptsize \textcolor{green!60!black}{$\uparrow$0.0}}
\\

GPT-4.1-mini
&
97.6\%
&
100.0\%
{\scriptsize \textcolor{green!60!black}{$\uparrow$2.4}}
\\

Kimi 2.6
&
96.8\%
&
100.0\%
{\scriptsize \textcolor{green!60!black}{$\uparrow$3.2}}
\\

\bottomrule
\end{tabular}
\end{table}

Table~\ref{tab:compiler_ecosystem}(C) further shows that the compiler generalizes across heterogeneous LLM backbones, including Gemini 3.1, GPT-4.1-mini, and Kimi 2.6, consistently achieving near-perfect parse success rates (\(99.5\%\sim100.0\%\)). This suggests that BioinfoMCP acts as a stabilizing interface layer for MCP generation rather than relying on a single frontier model.
\begin{table*}[t]
\centering
\small
\setlength{\tabcolsep}{4.5pt}
\renewcommand{\arraystretch}{1.15}

\caption{
Main experimental results on BioAgentBench and LAB-Bench.
Higher is better for all metrics.
Bold indicates the best result, underline indicates the second-best result,
and gray highlights BioManus. $^\dagger$ Human expert results are reported from the LAB-Bench benchmark.
}
\label{tab:main_results}

\begin{tabular}{lccccc}
\toprule

\multirow{2}{*}{Method}
&
\multicolumn{2}{c}{\textbf{BioAgentBench}~\cite{fa2026bioagentbenchaiagent}}
&
\multicolumn{3}{c}{\textbf{LAB-Bench}~\cite{laurent2024labbenchmeasuringcapabilitieslanguage}}
\\

\cmidrule(lr){2-3}
\cmidrule(lr){4-6}

&
Mean Score$\uparrow$
&
Pass Count$\uparrow$
&
DbQA$\uparrow$
&
SeqQA$\uparrow$
&
CloningScenarios$\uparrow$
\\

\midrule

\multicolumn{6}{l}{\textit{Human Reference}} \\

Human Expert$^\dagger$
&
---
&
---
&
74.70\%
&
78.80\%
&
60.00\%
\\

\midrule

\multicolumn{6}{l}{\textit{Base LLMs}} \\

GPT-5.5~
&
---
&
---
&
23.70\%
&
16.96\%
&
21.21\%
\\

DeepSeek-V4~\cite{deepseekai2026deepseekv4}
&
---
&
---
&
27.30\%
&
49.84\%
&
39.39\%
\\

\midrule

\multicolumn{6}{l}{\textit{Biomedical Agent Systems}} \\

ReAct-Code~\cite{yao2022react}
&
39.25\%
&
\underline{4/10}
&
27.62\%
&
43.49\%
&
36.36\%
\\

Biomni-ReAct~\cite{huang2025biomni}
&
42.10\%
&
3/10
&
28.11\%
&
74.60\%
&
24.24\%
\\

Biomni~\cite{huang2025biomni}
&
39.67\%
&
\underline{4/10}
&
57.46\%
&
84.76\%
&
\underline{78.79\%}
\\

Biomni-100
&
43.45\%
&
\underline{4/10}
&
\textbf{75.35\%}
&
79.37\%
&
69.70\%
\\

Biomni-500
&
36.75\%
&
3/10
&
\underline{72.79\%}
&
74.92\%
&
75.76\%
\\

Biomni-1k
&
41.50\%
&
3/10
&
65.15\%
&
72.88\%
&
75.76\%
\\

Biomni-2k
&
\underline{46.16\%}
&
3/10
&
60.80\%
&
77.27\%
&
72.73\%
\\

\midrule

\multicolumn{6}{l}{\textit{Ours}} \\

\rowcolor{gray!12}
\textbf{BioManus}
&
\textbf{46.84\%}
&
\textbf{4/10}
&
67.29\%
&
\textbf{90.48\%}
&
\textbf{81.82\%}
\\

\bottomrule
\end{tabular}

\end{table*}
\subsection{Main Results on BioAgentBench and LAB-Bench}
Table~\ref{tab:main_results} summarizes the main results on BioAgentBench and
LAB-Bench. On BioAgentBench, agents are evaluated on 10 realistic biomedical
analysis tasks using an LLM judge that scores final outputs against
ground-truth artifacts on a continuous \(0\!\sim\!1\) scale; pass count reports
tasks exceeding the 80\% correctness threshold. On LAB-Bench, DbQA, SeqQA,
and CloningScenarios are evaluated by exact-match accuracy on the official
test split.
\begin{figure}[!htb]
    \centering
    \includegraphics[width=0.5\textwidth]{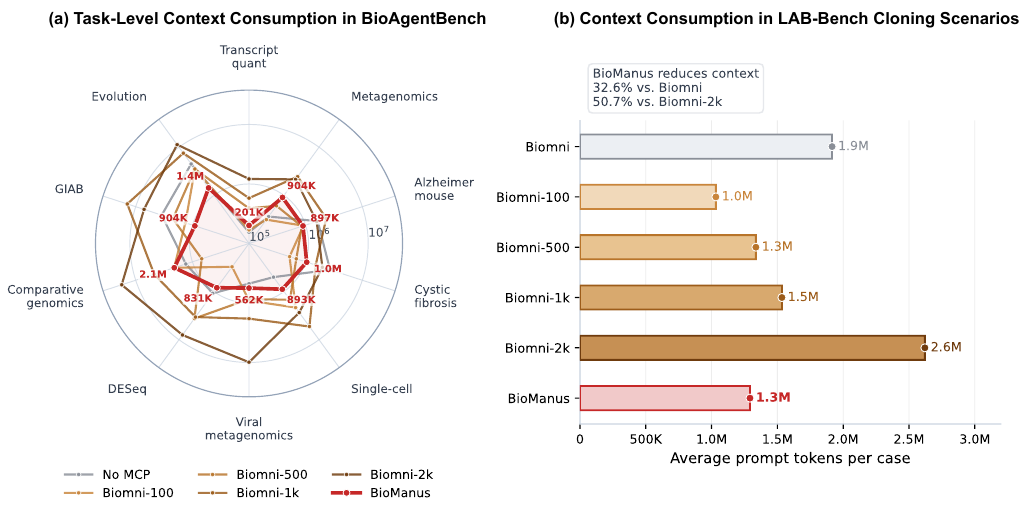}
    \caption{Planning Context Scaling Across Biomedical Agent Benchmarks}
    \label{fig:context}
\end{figure}
BioManus achieves the strongest overall performance across both benchmarks.
On BioAgentBench, it obtains the highest LLM-judge score (\(46.84\%\)),
outperforming the strongest Biomni variant (Biomni-2k) while matching the best
pass count (\(4/10\)). On LAB-Bench, BioManus achieves the best overall
performance on SeqQA (\(90.48\%\)) and CloningScenarios (\(81.82\%\)),
substantially outperforming existing biomedical agent baselines. Although
BioManus does not achieve the highest score on DbQA, this benchmark is
primarily designed around database-centric API interaction and structured
retrieval. Even under this setting, BioManus still achieves competitive
performance (\(67.29\%\)), demonstrating that graph-scaffolded planning
remains effective for database-oriented biomedical workflows.

We further observe that increasing Biomni's MCP inventory does not yield
monotonic gains. While Biomni-100 improves over the original Biomni baseline,
larger tool inventories introduce increasingly unstable behavior, with
Biomni-500 and Biomni-1k degrading on both BioAgentBench and LAB-Bench. This
supports our diagnosis that prompt-based retrieval becomes unstable as the
biomedical action space expands.

Figure~\ref{fig:context} further shows that larger MCP inventories lead to
substantially higher and more variable planning-context consumption for
prompt-retrieval baselines. In contrast, BioManus maintains lower and more
stable context usage while achieving stronger downstream performance,
supporting the advantage of graph-scaffolded planning over compact
task-specific executable subgraphs. Additional statistics are displayed in the Appendix~\ref{app:context_statistics}.
\subsection{Ablation Study: Dissecting the Contributions of Graph-Scaffolded Planning and MCP infrastructure}
\begin{figure}[!htb]
    \centering
    \includegraphics[width=0.5\textwidth]{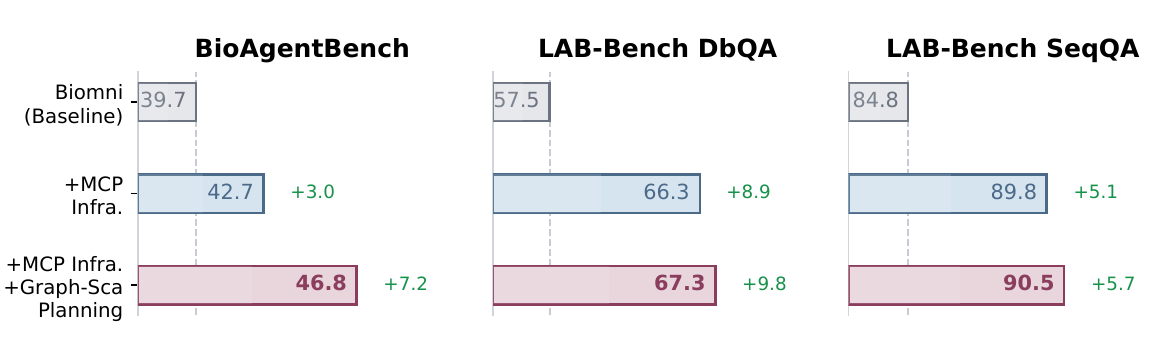}
    \caption{
    Ablation study of MCP-server infrastructure and graph-scaffolded planning.
    }
    \label{fig:ablation_study}
\end{figure}
To better understand the contribution of each component, we conduct an ablation 
study on BioAgentBench and LAB-Bench by progressively introducing MCP-native 
execution infrastructure and graph-scaffolded planning into the baseline.

Figure~\ref{fig:ablation_study} summarizes the results. Adding MCP-server 
infrastructure alone consistently improves performance across all benchmarks, 
increasing the BioAgentBench LLM-judge score from \(39.7\%\) to \(42.7\%\), 
while improving LAB-Bench performance from 
\(57.5\%\rightarrow66.4\%\) on DbQA and 
\(84.8\%\rightarrow89.8\%\) on SeqQA. These gains indicate that standardized 
MCP interfaces improve execution robustness and reduce environment 
fragmentation during multi-step biological analysis. 
Adding graph-scaffolded planning on top of MCP infrastructure yields further 
improvements. Relative to the baseline, this corresponds to 
absolute gains of \(+7.2\%\), \(+9.8\%\), and \(+5.7\%\), respectively.

\section{Conclusion}
We presented BioManus, a graph-scaffolded, MCP-native biomedical agent that addresses a key scalability bottleneck in current biomedical tool-use systems: prompt-based retrieval tightly couples tool coverage with planning-context size. BioManus combines the BioinfoMCP Compiler, which converts heterogeneous bioinformatics software into standardized MCP servers, with a Typed Capability Graph for structured workflow planning over tools, operations, datatypes, and workflow stages. By replacing flat prompt-level retrieval with graph-native executable reasoning, BioManus improves context efficiency, workflow validity, and downstream performance across large-scale biomedical benchmarks. More broadly, our results suggest a paradigm shift for scientific agents, from planning over flat prompt-retrieved tool descriptions to reasoning over structured executable capability graphs.
\section{Limitations}
BioManus represents an initial step toward scalable graph-scaffolded biomedical
agents, and several limitations remain. First, BioinfoMCP Compiler currently
relies on available tool documentation, such as manuals and command-line help
messages, to infer executable MCP interfaces; richer documentation and
execution-based validation could further improve compiler reliability. Second,
the Typed Capability Graph currently uses lightweight semantic annotations over
tools, operations, datatypes, and workflow stages, and could be further
extended with finer-grained biological ontologies and curated workflow
knowledge. Third, although dynamic MCP registration substantially reduces
context overhead and tool confusion, its effectiveness still depends on the
quality of graph retrieval and operation planning. Finally, BioManus is
designed as an assistive system for biomedical research, and expert oversight
remains important for validating scientific interpretations and downstream
biological conclusions.

% \section*{Acknowledgments}

% We thank the members of our research group for their helpful discussions and
% feedback throughout this project. This work was supported by The Chinese
% University of Hong Kong, Shenzhen (CUHK-Shenzhen), under Award
% No.~UDF01004172.

% \paragraph{Use of AI assistance.}
% During the preparation of this manuscript, the authors used ChatGPT to assist
% with language polishing and code comments. The authors reviewed, edited, and
% verified the resulting content as needed, and take full responsibility for the
% final publication.

% \paragraph{Competing interests.}
% The authors declare no competing interests.

% \paragraph{Author contributions.}
% Z.C., F.W., and J.Z. conceived the project. Z.C. and F.W. designed the
% computational framework, implemented the system, and analyzed the experimental
% results. J.Z. supervised the project. Z.C., F.W., and J.Z. wrote the manuscript
% and supplementary materials. All authors discussed the results and contributed
% to the final manuscript.

\nocite{*}
\bibliography{custom}

@book{Aho:72,
    author  = {Alfred V. Aho and Jeffrey D. Ullman},
    title   = {The Theory of Parsing, Translation and Compiling},
    year    = "1972",
    volume  = "1",
    publisher = {Prentice-Hall},
    address = {Englewood Cliffs, NJ}
}

@book{APA:83,
    author  = {{American Psychological Association}},
    title   = {Publications Manual},
    year    = "1983",
    publisher = {American Psychological Association},
    address = {Washington, DC}
}

@article{Chandra:81,
	author = {Ashok K. Chandra and Dexter C. Kozen and Larry J. Stockmeyer},
	year = "1981",
	title = {Alternation},
	journal = {Journal of the Association for Computing Machinery},
	volume = "28",
	number = "1",
	pages = "114--133",
	doi = "10.1145/322234.322243",
}

@inproceedings{andrew2007scalable,
  title={Scalable training of {L1}-regularized log-linear models},
  author={Andrew, Galen and Gao, Jianfeng},
  booktitle={Proceedings of the 24th International Conference on Machine Learning},
  pages={33--40},
  year={2007},
}

@book{Gusfield:97,
    author  = {Dan Gusfield},
    title   = {Algorithms on Strings, Trees and Sequences},
    year    = "1997",
    publisher = {Cambridge University Press},
    address = {Cambridge, UK}
}

@article{rasooli-tetrault-2015,
    author    = {Mohammad Sadegh Rasooli and Joel R. Tetreault},
    title     = {Yara Parser: {A} Fast and Accurate Dependency Parser},
    journal   = {Computing Research Repository},
    volume    = {arXiv:1503.06733},
    year      = {2015},
    url       = {http://arxiv.org/abs/1503.06733},
    note    = {version 2}
}

@article{Ando2005,
	Acmid = {1194905},
	Author = {Ando, Rie Kubota and Zhang, Tong},
	Issn = {1532-4435},
	Issue_Date = {12/1/2005},
	Journal = {Journal of Machine Learning Research},
	Month = dec,
	Numpages = {37},
	Pages = {1817--1853},
	Publisher = {JMLR.org},
	Title = {A Framework for Learning Predictive Structures from Multiple Tasks and Unlabeled Data},
	Volume = {6},
	Year = {2005}
}

@article{leipzig2017review,
	author = {Leipzig, Jeremy},
	title = {A review of bioinformatic pipeline frameworks},
	journal = {Briefings in Bioinformatics},
	volume = {18},
	number = {3},
	pages = {530--536},
	year = {2017},
	doi = {10.1093/bib/bbw020}
}

@article{stephens2015big,
	author = {Stephens, Zachary D. and Lee, Skylar W. and Faghri, Faraz and Campbell, Roy H. and Zhai, Chengxiang and Efron, Miles J. and Iyer, Ravishankar and Schatz, Michael C. and Sinha, Saurabh and Robinson, Gene E.},
	title = {Big Data: Astronomical or Genomical?},
	journal = {PLoS Biology},
	volume = {13},
	number = {7},
	pages = {e1002195},
	year = {2015},
	doi = {10.1371/journal.pbio.1002195}
}

@article{wang2023scientific,
	author = {Wang, Hanchen and Fu, Tianfan and Du, Yuanqi and Gao, Wenhao and Huang, Kexin and Liu, Ziming and Chandak, Payal and Liu, Shengchao and Van Katwyk, Peter and Deac, Andreea and others},
	title = {Scientific discovery in the age of artificial intelligence},
	journal = {Nature},
	volume = {620},
	pages = {47--60},
	year = {2023},
	doi = {10.1038/s41586-023-06221-2}
}

@article{guo2025deepseek,
	author = {Guo, Daya and Yang, Dejian and Zhang, Haowei and Song, Junxiao and Zhang, Ruoyu and Xu, Runxin and Zhu, Qihao and Ma, Shirong and Wang, Peiyi and Bi, Xiao and others},
	title = {{DeepSeek-R1}: Incentivizing reasoning capability in LLMs via reinforcement learning},
	journal = {arXiv preprint},
	year = {2025},
	eprint = {2501.12948},
	archivePrefix = {arXiv},
	primaryClass = {cs.CL}
}

@article{huang2025biomni,
  title={Biomni: A General-Purpose Biomedical AI Agent},
  author={Huang, Kexin and Zhang, Serena and Wang, Hanchen and Qu, Yuanhao and Lu, Yingzhou and Roohani, Yusuf and Li, Ryan and Qiu, Lin and Zhang, Junze and Di, Yin and others},
  journal={bioRxiv},
  pages={2025--05},
  year={2025},
  publisher={Cold Spring Harbor Laboratory}
}

@article{wratten2021reproducible,
  title={Reproducible, scalable, and shareable analysis pipelines with bioinformatics workflow managers},
  author={Wratten, Laura and Wilm, Andreas and G{\"o}ke, Jonathan},
  journal={Nature methods},
  volume={18},
  number={10},
  pages={1161--1168},
  year={2021},
  publisher={Nature Publishing Group US New York}
}

@article{landhuis2016scientific,
  title={Scientific literature: Information overload},
  author={Landhuis, Esther},
  journal={Nature},
  volume={535},
  number={7612},
  pages={457--458},
  year={2016},
  publisher={Nature Publishing Group UK London}
}

@article{brown2020language,
  title={Language models are few-shot learners},
  author={Brown, Tom and Mann, Benjamin and Ryder, Nick and Subbiah, Melanie and Kaplan, Jared D and Dhariwal, Prafulla and Neelakantan, Arvind and Shyam, Pranav and Sastry, Girish and Askell, Amanda and others},
  journal={Advances in neural information processing systems},
  volume={33},
  pages={1877--1901},
  year={2020}
}

@inproceedings{hong2024metagpt,
  title={MetaGPT: Meta programming for a multi-agent collaborative framework},
  author={Hong, Sirui and Zhuge, Mingchen and Chen, Jonathan and Zheng, Xiawu and Cheng, Yuheng and Wang, Jinlin and Zhang, Ceyao and Yau, Steven and Lin, Zijuan and Zhou, Liyang and others},
  booktitle={International Conference on Learning Representations},
  volume={2024},
  pages={23247--23275},
  year={2024}
}

@article{Jin_2024,
   title={GeneGPT: augmenting large language models with domain tools for improved access to biomedical information},
   volume={40},
   ISSN={1367-4811},
   url={http://dx.doi.org/10.1093/bioinformatics/btae075},
   DOI={10.1093/bioinformatics/btae075},
   number={2},
   journal={Bioinformatics},
   publisher={Oxford University Press (OUP)},
   author={Jin, Qiao and Yang, Yifan and Chen, Qingyu and Lu, Zhiyong},
   editor={Wren, Jonathan},
   year={2024},
   month=Feb }

@misc{wang2024geneagentselfverificationlanguageagent,
      title={GeneAgent: Self-verification Language Agent for Gene Set Knowledge Discovery using Domain Databases}, 
      author={Zhizheng Wang and Qiao Jin and Chih-Hsuan Wei and Shubo Tian and Po-Ting Lai and Qingqing Zhu and Chi-Ping Day and Christina Ross and Zhiyong Lu},
      year={2024},
      eprint={2405.16205},
      archivePrefix={arXiv},
      primaryClass={cs.AI},
      url={https://arxiv.org/abs/2405.16205}, 
}

@article{de2024varchat,
  title={VarChat: the generative AI assistant for the interpretation of human genomic variations},
  author={De Paoli, Federica and Berardelli, Silvia and Limongelli, Ivan and Rizzo, Ettore and Zucca, Susanna},
  journal={Bioinformatics},
  volume={40},
  number={4},
  pages={btae183},
  year={2024},
  publisher={Oxford University Press}
}

@article{li2025autopm3,
  title={AutoPM3: enhancing variant interpretation via LLM-driven PM3 evidence extraction from scientific literature},
  author={Li, Shumin and Wang, Yiding and Liu, Chi-Man and Huang, Yuanhua and Lam, Tak-Wah and Luo, Ruibang},
  journal={Bioinformatics},
  volume={41},
  number={7},
  pages={btaf382},
  year={2025},
  publisher={Oxford University Press}
}

@misc{xiao2024cellagentllmdrivenmultiagentframework,
      title={CellAgent: An LLM-driven Multi-Agent Framework for Automated Single-cell Data Analysis}, 
      author={Yihang Xiao and Jinyi Liu and Yan Zheng and Xiaohan Xie and Jianye Hao and Mingzhi Li and Ruitao Wang and Fei Ni and Yuxiao Li and Jintian Luo and Shaoqing Jiao and Jiajie Peng},
      year={2024},
      eprint={2407.09811},
      archivePrefix={arXiv},
      primaryClass={cs.AI},
      url={https://arxiv.org/abs/2407.09811}, 
}

@article{zhou2024ai,
  title={An AI agent for fully automated multi-omic analyses},
  author={Zhou, Juexiao and Zhang, Bin and Li, Guowei and Chen, Xiuying and Li, Haoyang and Xu, Xiaopeng and Chen, Siyuan and He, Wenjia and Xu, Chencheng and Liu, Liwei and others},
  journal={Advanced Science},
  volume={11},
  number={44},
  pages={2407094},
  year={2024},
  publisher={Wiley Online Library}
}

@article{jansen2025leveraging,
  title={Leveraging large language models for data analysis automation},
  author={Jansen, Jacqueline A and Manukyan, Art{\"u}r and Al Khoury, Nour and Akalin, Altuna},
  journal={PloS one},
  volume={20},
  number={2},
  pages={e0317084},
  year={2025},
  publisher={Public Library of Science San Francisco, CA USA}
}

@article{pickard2025automatic,
  title={Automatic biomarker discovery and enrichment with BRAD},
  author={Pickard, Joshua and Prakash, Ram and Choi, Marc Andrew and Oliven, Natalie and Stansbury, Cooper and Cwycyshyn, Jillian and Galioto, Nicholas and Gorodetsky, Alex and Velasquez, Alvaro and Rajapakse, Indika},
  journal={Bioinformatics},
  volume={41},
  number={5},
  pages={btaf159},
  year={2025},
  publisher={Oxford University Press}
}

@article{liang2025scwgbs,
  title={scWGBS-GPT: a foundation model for capturing long-range CPG dependencies in single-cell whole-genome bisulfite sequencing to enhance epigenetic analysis},
  author={Liang, Chaoqi and Ye, Peng and Yan, Hongliang and Zheng, Peng and Sun, Jianle and Wang, Yanni and Li, Yu and Ren, Yuchen and Jiang, Yuanpei and Xiang, Junjia and others},
  journal={bioRxiv},
  pages={2025--02},
  year={2025},
  publisher={Cold Spring Harbor Laboratory}
}

@article{wang2025spatialagent,
  title={SpatialAgent: An autonomous AI agent for spatial biology},
  author={Wang, Hanchen and He, Yichun and Coelho, Paula P and Bucci, Matthew and Nazir, Abbas and Chen, Bob and Trinh, Linh and Zhang, Serena and Huang, Kexin and Chandrasekar, Vineethkrishna and others},
  journal={bioRxiv},
  pages={2025--04},
  year={2025},
  publisher={Cold Spring Harbor Laboratory}
}

@article{garcia2025improving,
  title={Improving automated deep phenotyping through large language models using retrieval-augmented generation},
  author={Garcia, Brandon T and Westerfield, Lauren and Yelemali, Priya and Gogate, Nikhita and Rivera-Munoz, E Andres and Du, Haowei and Dawood, Moez and Jolly, Angad and Lupski, James R and Posey, Jennifer E},
  journal={Genome Medicine},
  volume={17},
  number={1},
  pages={91},
  year={2025},
  publisher={Springer}
}

@article{niu2022assessing,
  title={Assessing and assuring interoperability of a genomics file format},
  author={Niu, Yi Nian and Roberts, Eric G and Denisko, Danielle and Hoffman, Michael M},
  journal={Bioinformatics},
  volume={38},
  number={13},
  pages={3327--3336},
  year={2022},
  publisher={Oxford University Press}
}

@article{reiter2021streamlining,
  title={Streamlining data-intensive biology with workflow systems},
  author={Reiter, Taylor and Brooks, Phillip T and Irber, Luiz and Joslin, Shannon EK and Reid, Charles M and Scott, Camille and Brown, C Titus and Pierce-Ward, N Tessa},
  journal={GigaScience},
  volume={10},
  number={1},
  pages={giaa140},
  year={2021},
  publisher={Oxford University Press}
}

@article{yao2022react,
  title={React: Synergizing reasoning and acting in language models},
  author={Yao, Shunyu and Zhao, Jeffrey and Yu, Dian and Du, Nan and Shafran, Izhak and Narasimhan, Karthik and Cao, Yuan},
  journal={arXiv preprint arXiv:2210.03629},
  year={2022}
}

@misc{wang2024executablecodeactionselicit,
      title={Executable Code Actions Elicit Better LLM Agents}, 
      author={Xingyao Wang and Yangyi Chen and Lifan Yuan and Yizhe Zhang and Yunzhu Li and Hao Peng and Heng Ji},
      year={2024},
      eprint={2402.01030},
      archivePrefix={arXiv},
      primaryClass={cs.CL},
      url={https://arxiv.org/abs/2402.01030}, 
}

@article{cheng2024prompt,
  title={Prompt sapper: A LLM-empowered production tool for building AI chains},
  author={Cheng, Yu and Chen, Jieshan and Huang, Qing and Xing, Zhenchang and Xu, Xiwei and Lu, Qinghua},
  journal={ACM Transactions on Software Engineering and Methodology},
  volume={33},
  number={5},
  pages={1--24},
  year={2024},
  publisher={ACM New York, NY}
}

@misc{schick2023toolformerlanguagemodelsteach,
      title={Toolformer: Language Models Can Teach Themselves to Use Tools}, 
      author={Timo Schick and Jane Dwivedi-Yu and Roberto Dessì and Roberta Raileanu and Maria Lomeli and Luke Zettlemoyer and Nicola Cancedda and Thomas Scialom},
      year={2023},
      eprint={2302.04761},
      archivePrefix={arXiv},
      primaryClass={cs.CL},
      url={https://arxiv.org/abs/2302.04761}, 
}

@misc{patil2023gorillalargelanguagemodel,
      title={Gorilla: Large Language Model Connected with Massive APIs}, 
      author={Shishir G. Patil and Tianjun Zhang and Xin Wang and Joseph E. Gonzalez},
      year={2023},
      eprint={2305.15334},
      archivePrefix={arXiv},
      primaryClass={cs.CL},
      url={https://arxiv.org/abs/2305.15334}, 
}

@misc{fa2026bioagentbenchaiagent,
      title={BioAgent Bench: An AI Agent Evaluation Suite for Bioinformatics}, 
      author={Dionizije Fa and Marko Culjak and Bruno Pandza and Mateo Cupic},
      year={2026},
      eprint={2601.21800},
      archivePrefix={arXiv},
      primaryClass={cs.AI},
      url={https://arxiv.org/abs/2601.21800}, 
}

@misc{laurent2024labbenchmeasuringcapabilitieslanguage,
      title={LAB-Bench: Measuring Capabilities of Language Models for Biology Research}, 
      author={Jon M. Laurent and Joseph D. Janizek and Michael Ruzo and Michaela M. Hinks and Michael J. Hammerling and Siddharth Narayanan and Manvitha Ponnapati and Andrew D. White and Samuel G. Rodriques},
      year={2024},
      eprint={2407.10362},
      archivePrefix={arXiv},
      primaryClass={cs.AI},
      url={https://arxiv.org/abs/2407.10362}, 
}

@article{gruning2018bioconda,
  title={Bioconda: sustainable and comprehensive software distribution for the life sciences},
  author={Gr{\"u}ning, Bj{\"o}rn and Dale, Ryan and Sj{\"o}din, Andreas and Chapman, Brad A and Rowe, Jillian and Tomkins-Tinch, Christopher H and Valieris, Renan and K{\"o}ster, Johannes and Bioconda Team},
  journal={Nature methods},
  volume={15},
  number={7},
  pages={475--476},
  year={2018},
  publisher={Nature Publishing Group US New York}
}

@misc{qin2023toolllmfacilitatinglargelanguage,
      title={ToolLLM: Facilitating Large Language Models to Master 16000+ Real-world APIs}, 
      author={Yujia Qin and Shihao Liang and Yining Ye and Kunlun Zhu and Lan Yan and Yaxi Lu and Yankai Lin and Xin Cong and Xiangru Tang and Bill Qian and Sihan Zhao and Lauren Hong and Runchu Tian and Ruobing Xie and Jie Zhou and Mark Gerstein and Dahai Li and Zhiyuan Liu and Maosong Sun},
      year={2023},
      eprint={2307.16789},
      archivePrefix={arXiv},
      primaryClass={cs.AI},
      url={https://arxiv.org/abs/2307.16789}, 
}

@misc{li2023apibankcomprehensivebenchmarktoolaugmented,
      title={API-Bank: A Comprehensive Benchmark for Tool-Augmented LLMs}, 
      author={Minghao Li and Yingxiu Zhao and Bowen Yu and Feifan Song and Hangyu Li and Haiyang Yu and Zhoujun Li and Fei Huang and Yongbin Li},
      year={2023},
      eprint={2304.08244},
      archivePrefix={arXiv},
      primaryClass={cs.CL},
      url={https://arxiv.org/abs/2304.08244}, 
}

@misc{song2023restgptconnectinglargelanguage,
      title={RestGPT: Connecting Large Language Models with Real-World RESTful APIs}, 
      author={Yifan Song and Weimin Xiong and Dawei Zhu and Wenhao Wu and Han Qian and Mingbo Song and Hailiang Huang and Cheng Li and Ke Wang and Rong Yao and Ye Tian and Sujian Li},
      year={2023},
      eprint={2306.06624},
      archivePrefix={arXiv},
      primaryClass={cs.CL},
      url={https://arxiv.org/abs/2306.06624}, 
}

@misc{wang2025mcpbenchbenchmarkingtoolusingllm,
      title={MCP-Bench: Benchmarking Tool-Using LLM Agents with Complex Real-World Tasks via MCP Servers}, 
      author={Zhenting Wang and Qi Chang and Hemani Patel and Shashank Biju and Cheng-En Wu and Quan Liu and Aolin Ding and Alireza Rezazadeh and Ankit Shah and Yujia Bao and Eugene Siow},
      year={2025},
      eprint={2508.20453},
      archivePrefix={arXiv},
      primaryClass={cs.CL},
      url={https://arxiv.org/abs/2508.20453}, 
}

@article{liu-etal-2024-lost,
    title = "Lost in the Middle: How Language Models Use Long Contexts",
    author = "Liu, Nelson F.  and
      Lin, Kevin  and
      Hewitt, John  and
      Paranjape, Ashwin  and
      Bevilacqua, Michele  and
      Petroni, Fabio  and
      Liang, Percy",
    journal = "Transactions of the Association for Computational Linguistics",
    volume = "12",
    year = "2024",
    address = "Cambridge, MA",
    publisher = "MIT Press",
    url = "https://aclanthology.org/2024.tacl-1.9/",
    doi = "10.1162/tacl_a_00638",
    pages = "157--173"
}

@article{da2017biocontainers,
  title={BioContainers: an open-source and community-driven framework for software standardization},
  author={da Veiga Leprevost, Felipe and Gr{\"u}ning, Bj{\"o}rn A and Alves Aflitos, Saulo and R{\"o}st, Hannes L and Uszkoreit, Julian and Barsnes, Harald and Vaudel, Marc and Moreno, Pablo and Gatto, Laurent and Weber, Jonas and others},
  journal={Bioinformatics},
  volume={33},
  number={16},
  pages={2580--2582},
  year={2017},
  publisher={Oxford University Press}
}

@article{li2026complexmcp,
  title={ComplexMCP: Evaluation of LLM Agents in Dynamic, Interdependent, and Large-Scale Tool Sandbox},
  author={Li, Yuanyang and Yang, Xue and Wang, Longyue and Luo, Weihua and Chen, Hongyang},
  journal={arXiv preprint arXiv:2605.10787},
  year={2026}
}

@article{amstutz2016common,
  title={Common workflow language, v1. 0},
  author={Amstutz, Peter and Crusoe, Michael R and Tijani{\'c}, Neboj{\v{s}}a and Chapman, Brad and Chilton, John and Heuer, Michael and Kartashov, Andrey and Leehr, Dan and M{\'e}nager, Herv{\'e} and Nedeljkovich, Maya and others},
  year={2016},
  publisher={Figshare},
  journal={Figshare}
}

@article{goecks2010galaxy,
  title={Galaxy: a comprehensive approach for supporting accessible, reproducible, and transparent computational research in the life sciences},
  author={Goecks, Jeremy and Nekrutenko, Anton and Taylor, James and Galaxy Team team@ galaxyproject. org},
  journal={Genome biology},
  volume={11},
  number={8},
  pages={R86},
  year={2010},
  publisher={Springer}
}

@article{koster2012snakemake,
  title={Snakemake—a scalable bioinformatics workflow engine},
  author={K{\"o}ster, Johannes and Rahmann, Sven},
  journal={Bioinformatics},
  volume={28},
  number={19},
  pages={2520--2522},
  year={2012},
  publisher={Oxford University Press}
}

@article{di2017nextflow,
  title={Nextflow enables reproducible computational workflows},
  author={Di Tommaso, Paolo and Chatzou, Maria and Floden, Evan W and Barja, Pablo Prieto and Palumbo, Emilio and Notredame, Cedric},
  journal={Nature biotechnology},
  volume={35},
  number={4},
  pages={316--319},
  year={2017},
  publisher={Nature Publishing Group US New York}
}

@misc{deepseekai2026deepseekv4,
      title={DeepSeek-V4: Towards Highly Efficient Million-Token Context Intelligence},
      author={DeepSeek-AI},
      year={2026},
}

@misc{kimiteam2026kimik25visualagentic,
      title={Kimi K2.5: Visual Agentic Intelligence}, 
      author={Kimi Team and Tongtong Bai and Yifan Bai and Yiping Bao and S. H. Cai and Yuan Cao and Y. Charles and H. S. Che and Cheng Chen and Guanduo Chen and Huarong Chen and Jia Chen and Jiahao Chen and Jianlong Chen and Jun Chen and Kefan Chen and Liang Chen and Ruijue Chen and Xinhao Chen and Yanru Chen and Yanxu Chen and Yicun Chen and Yimin Chen and Yingjiang Chen and Yuankun Chen and Yujie Chen and Yutian Chen and Zhirong Chen and Ziwei Chen and Dazhi Cheng and Minghan Chu and Jialei Cui and Jiaqi Deng and Muxi Diao and Hao Ding and Mengfan Dong and Mengnan Dong and Yuxin Dong and Yuhao Dong and Angang Du and Chenzhuang Du and Dikang Du and Lingxiao Du and Yulun Du and Yu Fan and Shengjun Fang and Qiulin Feng and Yichen Feng and Garimugai Fu and Kelin Fu and Hongcheng Gao and Tong Gao and Yuyao Ge and Shangyi Geng and Chengyang Gong and Xiaochen Gong and Zhuoma Gongque and Qizheng Gu and Xinran Gu and Yicheng Gu and Longyu Guan and Yuanying Guo and Xiaoru Hao and Weiran He and Wenyang He and Yunjia He and Chao Hong and Hao Hu and Jiaxi Hu and Yangyang Hu and Zhenxing Hu and Ke Huang and Ruiyuan Huang and Weixiao Huang and Zhiqi Huang and Tao Jiang and Zhejun Jiang and Xinyi Jin and Yu Jing and Guokun Lai and Aidi Li and C. Li and Cheng Li and Fang Li and Guanghe Li and Guanyu Li and Haitao Li and Haoyang Li and Jia Li and Jingwei Li and Junxiong Li and Lincan Li and Mo Li and Weihong Li and Wentao Li and Xinhang Li and Xinhao Li and Yang Li and Yanhao Li and Yiwei Li and Yuxiao Li and Zhaowei Li and Zheming Li and Weilong Liao and Jiawei Lin and Xiaohan Lin and Zhishan Lin and Zichao Lin and Cheng Liu and Chenyu Liu and Hongzhang Liu and Liang Liu and Shaowei Liu and Shudong Liu and Shuran Liu and Tianwei Liu and Tianyu Liu and Weizhou Liu and Xiangyan Liu and Yangyang Liu and Yanming Liu and Yibo Liu and Yuanxin Liu and Yue Liu and Zhengying Liu and Zhongnuo Liu and Enzhe Lu and Haoyu Lu and Zhiyuan Lu and Junyu Luo and Tongxu Luo and Yashuo Luo and Long Ma and Yingwei Ma and Shaoguang Mao and Yuan Mei and Xin Men and Fanqing Meng and Zhiyong Meng and Yibo Miao and Minqing Ni and Kun Ouyang and Siyuan Pan and Bo Pang and Yuchao Qian and Ruoyu Qin and Zeyu Qin and Jiezhong Qiu and Bowen Qu and Zeyu Shang and Youbo Shao and Tianxiao Shen and Zhennan Shen and Juanfeng Shi and Lidong Shi and Shengyuan Shi and Feifan Song and Pengwei Song and Tianhui Song and Xiaoxi Song and Hongjin Su and Jianlin Su and Zhaochen Su and Lin Sui and Jinsong Sun and Junyao Sun and Tongyu Sun and Flood Sung and Yunpeng Tai and Chuning Tang and Heyi Tang and Xiaojuan Tang and Zhengyang Tang and Jiawen Tao and Shiyuan Teng and Chaoran Tian and Pengfei Tian and Ao Wang and Bowen Wang and Chensi Wang and Chuang Wang and Congcong Wang and Dingkun Wang and Dinglu Wang and Dongliang Wang and Feng Wang and Hailong Wang and Haiming Wang and Hengzhi Wang and Huaqing Wang and Hui Wang and Jiahao Wang and Jinhong Wang and Jiuzheng Wang and Kaixin Wang and Linian Wang and Qibin Wang and Shengjie Wang and Shuyi Wang and Si Wang and Wei Wang and Xiaochen Wang and Xinyuan Wang and Yao Wang and Yejie Wang and Yipu Wang and Yiqin Wang and Yucheng Wang and Yuzhi Wang and Zhaoji Wang and Zhaowei Wang and Zhengtao Wang and Zhexu Wang and Zihan Wang and Zizhe Wang and Chu Wei and Ming Wei and Chuan Wen and Zichen Wen and Chengjie Wu and Haoning Wu and Junyan Wu and Rucong Wu and Wenhao Wu and Yuefeng Wu and Yuhao Wu and Yuxin Wu and Zijian Wu and Chenjun Xiao and Jin Xie and Xiaotong Xie and Yuchong Xie and Yifei Xin and Bowei Xing and Boyu Xu and Jianfan Xu and Jing Xu and Jinjing Xu and L. H. Xu and Lin Xu and Suting Xu and Weixin Xu and Xinbo Xu and Xinran Xu and Yangchuan Xu and Yichang Xu and Yuemeng Xu and Zelai Xu and Ziyao Xu and Junjie Yan and Yuzi Yan and Guangyao Yang and Hao Yang and Junwei Yang and Kai Yang and Ningyuan Yang and Ruihan Yang and Xiaofei Yang and Xinlong Yang and Ying Yang and Yi Yang and Yi Yang and Zhen Yang and Zhilin Yang and Zonghan Yang and Haotian Yao and Dan Ye and Wenjie Ye and Zhuorui Ye and Bohong Yin and Chengzhen Yu and Longhui Yu and Tao Yu and Tianxiang Yu and Enming Yuan and Mengjie Yuan and Xiaokun Yuan and Yang Yue and Weihao Zeng and Dunyuan Zha and Haobing Zhan and Dehao Zhang and Hao Zhang and Jin Zhang and Puqi Zhang and Qiao Zhang and Rui Zhang and Xiaobin Zhang and Y. Zhang and Yadong Zhang and Yangkun Zhang and Yichi Zhang and Yizhi Zhang and Yongting Zhang and Yu Zhang and Yushun Zhang and Yutao Zhang and Yutong Zhang and Zheng Zhang and Chenguang Zhao and Feifan Zhao and Jinxiang Zhao and Shuai Zhao and Xiangyu Zhao and Yikai Zhao and Zijia Zhao and Huabin Zheng and Ruihan Zheng and Shaojie Zheng and Tengyang Zheng and Junfeng Zhong and Longguang Zhong and Weiming Zhong and M. Zhou and Runjie Zhou and Xinyu Zhou and Zaida Zhou and Jinguo Zhu and Liya Zhu and Xinhao Zhu and Yuxuan Zhu and Zhen Zhu and Jingze Zhuang and Weiyu Zhuang and Ying Zou and Xinxing Zu},
      year={2026},
      eprint={2602.02276},
      archivePrefix={arXiv},
      primaryClass={cs.CL},
      url={https://arxiv.org/abs/2602.02276}, 
}

@article{zhang2025promptbio,
  title={Promptbio: a multi-agent AI platform for bioinformatics data analysis},
  author={Zhang, Minzhe and Gu, Wenhao and Han, Bowei and Guo, Vincent and Addoni, Chintan and Chen, Jiayu and Ma, Youjia and Leng, Yang and Li, Kai and Lin, Xiaoxi and others},
  journal={bioRxiv},
  pages={2025--07},
  year={2025},
  publisher={Cold Spring Harbor Laboratory}
}

@article{mehandru2025bioagents,
  title={BioAgents: Bridging the gap in bioinformatics analysis with multi-agent systems},
  author={Mehandru, Nikita and Hall, Amanda K and Melnichenko, Olesya and Dubinina, Yulia and Tsirulnikov, Daniel and Bamman, David and Alaa, Ahmed and Saponas, Scott and Malladi, Venkat S},
  journal={Scientific Reports},
  volume={15},
  number={1},
  pages={39036},
  year={2025},
  publisher={Nature Publishing Group UK London}
}

@article{jin2025stella,
  title={Stella: Self-evolving llm agent for biomedical research},
  author={Jin, Ruofan and Zhang, Zaixi and Wang, Mengdi and Cong, Le},
  journal={arXiv preprint arXiv:2507.02004},
  year={2025}
}

@article{suwer2026biomedical,
  title={Biomedical systems biology workflow orchestration and execution with PoSyMed},
  author={S{\"u}wer, Simon and Chervontseva, Zoe and Bagemihl, Kester and Baumbach, Jan and Tsoy, Olga and Maier, Andreas},
  journal={arXiv preprint arXiv:2604.20906},
  year={2026}
}
\appendix

\section{Implementation Details}
\label{app:implementation}

\subsection{Implementation and Hardware}

All experiments are implemented in Python using LangChain/LangGraph, and MCP-based tool execution. BioManus and all
comparison baselines are evaluated with the same benchmark harnesses and  the same LLM backend and execution environment.

The agent LLM is accessed through the Biomni Custom provider interface. In our
experiments, the backend is configured as \texttt{DeepSeek-V4}. We do not perform local model training or
fine-tuning; all neural generation is served through external LLM APIs. 
Experiments are conducted in a Python 3.11 conda environment on a machine with
two Intel Xeon Gold 6430 CPUs, 128 logical CPU cores, and approximately
1.0 TiB RAM.
\subsection{Experimental Setup and Evaluation Protocols}

We use the CodeAct-style agent as the base agent framework for both
BioManus and Biomni baselines. The default generation temperature follows the Biomni configuration
(\(\texttt{temperature}=0.7\)), and the Custom LLM wrapper uses
\(\texttt{max\_tokens}=8192\).

BioManus augments the base agent with an explicit MCP ToolGraph and
GraphRouter. The generated graph snapshot used in our experiments contains
910 servers, 3,500 tools. 

\paragraph{LAB-Bench Protocol}

We evaluate on the DbQA and SeqQA subsets of LAB-Bench. For each subset, we use
seed \(20260514\), reserve 45 examples as the development portion, and evaluate
on 315 test examples. 
The agent is instructed to return final answers in the LAB-Bench format:
\[
\texttt{<solution>[ANSWER]X[/ANSWER]</solution>}.
\]

\paragraph{BioAgentBench Protocol}
We evaluate BioAgentBench using the official task metadata and dataset root.
Each task is executed in an isolated run directory under explicit data-access
constraints: the agent can read only the task input and reference directories,
and write only to the run directory. Access to the results directory and
sibling benchmark tasks is blocked. 
LLM-based(DeepSeek-V4-Pro) evaluation is used and the final generated artifacts are compared against ground-truth
artifacts.

\section{Prompt Templates}
\label{app:prompts}

BioManus uses six categories of prompts throughout tool onboarding, graph
planning, execution, and evaluation. All prompts are implemented as structured
templates with task-specific fields filled at runtime. For compactness, we
visualize the \textbf{BioinfoMCP Compiler Generation Prompt} using a structured
diagram and provide the remaining prompts as boxed templates.

\subsection{BioinfoMCP Compiler Generation Prompt}
\label{app:prompt_compiler_generation}

\begin{figure*}[!htb]
    \centering
    \includegraphics[width=1\textwidth, alt={Give a detail explanation for each part of the system prompt structure for BioinfoMCP Converter, from the Role, Task, Requirements, and Instructions.}]{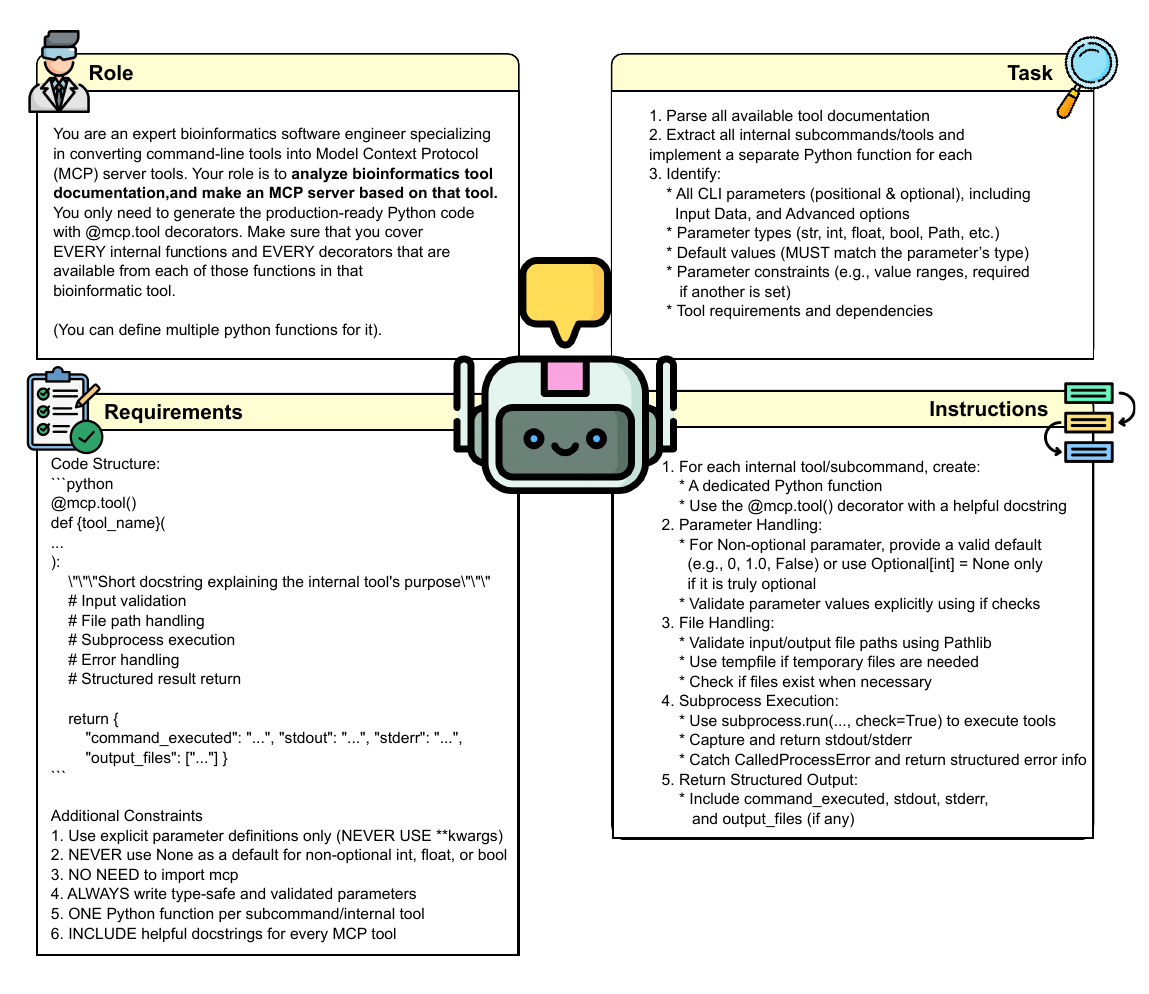}
    \caption{Structured system prompt template for BioinfoMCP Compiler.}
    \label{fig:rtir}
\end{figure*}

The BioinfoMCP Compiler uses a structured system prompt to convert
heterogeneous bioinformatics tools into executable MCP servers. Rather than
showing the full prompt verbatim, we summarize its organization in
Figure~\ref{fig:rtir}. The prompt is designed around four components:
\textbf{Role}, which defines the model as an MCP compiler for bioinformatics
tools; \textbf{Task}, which specifies the goal of generating MCP-compatible
server code; \textbf{Requirements}, which enforce interface correctness,
robustness, and structured outputs; and \textbf{Instructions}, which provide
step-by-step guidance for parameter handling, subprocess execution, and final
code formatting.

\subsection{Task Schema Extraction Prompt}
\label{app:prompt_schema_extraction}

This prompt maps a natural-language biological request to a structured task
specification. The extracted schema provides semantic anchors for downstream
GraphRAG retrieval and graph-scaffolded planning.

\begin{promptbox}{Task Schema Extraction Prompt}
\begin{Verbatim}[breaklines=true,breakanywhere=true,fontsize=\scriptsize]
System:
You are a biomedical workflow schema extractor.
Convert natural-language biological analysis requests into structured JSON.

User:
User query:
{user_query}

Available biological domains:
{domain_list}

Known datatype vocabulary:
{datatype_vocabulary}

Known operation vocabulary:
{operation_vocabulary}

Task:
Extract the structured task specification needed for graph-based biomedical
workflow planning.

Return strict JSON only with the following fields:
{
  "task_summary": string,
  "subtasks": list of strings,
  "biological_entities": list of strings,
  "domain_keywords": list of strings,
  "biological_domains": list of strings,
  "input_datatypes": list of strings,
  "output_datatypes": list of strings,
  "operation_hints": list of strings,
  "capability_hints": list of strings,
  "workflow_stages": list of strings,
  "organism": string or null,
  "constraints": list of strings,
  "success_criteria": list of strings
}

Rules:
- Use only information supported by the query.
- Prefer typed biological datatypes when they can be inferred.
- Do not invent tools.
- Do not produce prose outside the JSON object.
\end{Verbatim}
\end{promptbox}

\subsection{Graph-Scaffolded Planning Prompt}
\label{app:prompt_graph_planning}

This prompt is used by the LLM planner after GraphRAG retrieves a compact
task-specific subgraph. The planner reasons over operations and datatypes
rather than over the full raw tool inventory.

\begin{promptbox}{Graph-Scaffolded Planning Prompt}
\begin{Verbatim}[breaklines=true,breakanywhere=true,fontsize=\scriptsize]
System:
You are a biomedical workflow planner.
Plan only over the provided graph context and candidate MCP tools.

User:
User query:
{user_query}

Structured task specification:
{task_specification_json}

Compact GraphRAG subgraph:
{compact_subgraph_json}

Candidate operations:
{candidate_operations_json}

Candidate MCP tools:
{candidate_tools_json}

Operation path hints:
{operation_path_hint_json}

Task:
Generate an operation-level workflow scaffold and bind operations to executable
MCP tools.

Planning rules:
1. Plan at the operation level before selecting concrete tools.
2. Use only tools appearing in the candidate MCP tool list.
3. Respect datatype compatibility between consecutive operations.
4. Prefer workflow-stage orderings that are biologically plausible.
5. Avoid redundant or unrelated tools.
6. If multiple tools can implement the same operation, choose the most relevant
   one based on datatype compatibility and task semantics.
7. Do not invent MCP servers or tool names.

Return strict JSON only:
{
  "operation_path": [
    {
      "step_id": integer,
      "operation": string,
      "input_datatypes": list of strings,
      "output_datatypes": list of strings,
      "purpose": string
    }
  ],
  "tool_plan": [
    {
      "step_id": integer,
      "operation": string,
      "selected_tool": string,
      "selected_server": string,
      "arguments_hint": object
    }
  ],
  "selected_mcp_servers": list of strings,
  "dag_call_chain": list of objects,
  "planning_notes": string
}
\end{Verbatim}
\end{promptbox}

\subsection{Execution Agent Prompt}
\label{app:prompt_execution_agent}

This prompt guides the runtime agent after BioManus dynamically registers the
selected MCP servers. The execution agent receives a compact workflow scaffold
rather than the full global tool inventory.

\begin{promptbox}{Execution Agent Prompt}
\begin{Verbatim}[breaklines=true,breakanywhere=true,fontsize=\scriptsize]
System:
You are BioManus, an MCP-native biomedical execution agent.
Execute the provided biological workflow using only the registered tools and
available files.

User:
User query:
{user_query}

Structured task specification:
{task_specification_json}

Operation-level workflow scaffold:
{operation_path_json}

DAG call chain:
{dag_call_chain_json}

Registered MCP servers and tools:
{registered_mcp_tools_json}

Runtime constraints:
{runtime_constraints_json}

Task:
Execute the workflow, inspect intermediate outputs, recover from tool errors
when possible, and return the final biological result.

Execution rules:
1. Follow the operation-level workflow scaffold unless observations indicate a
   necessary correction.
2. Use only registered MCP tools and permitted local code execution.
3. Inspect tool outputs before proceeding to downstream steps.
4. If a tool call fails, diagnose the error and retry with corrected parameters
   when appropriate.
5. Maintain intermediate artifacts in the permitted run directory.
6. Do not access benchmark results, sibling tasks, or disallowed directories.
7. Produce a concise final answer and list generated result artifacts.

Final response format:
- Summary of executed workflow
- Final answer
- Important output files
- Any errors encountered and how they were resolved
\end{Verbatim}
\end{promptbox}

\subsection{BioAgentBench LLM-Judge Prompt}
\label{app:prompt_llm_judge}

BioAgentBench uses an LLM-based evaluator to assess the correctness of final
agent-produced artifacts. The evaluator follows an \emph{answer-first} design:
predicted artifacts are directly compared against ground-truth artifacts, while
pipeline traces, path mentions, and step-completion evidence are used only as
supporting context. This prevents agents from receiving high scores merely for
executing many workflow steps without producing correct biological outputs.

The main evaluation score is \texttt{results\_match}, a continuous
\(0\!\sim\!1\) correctness score measuring semantic agreement between predicted
and reference artifacts. The judge also returns
\texttt{results\_match\_pass}, which indicates whether the prediction satisfies
the task-specific correctness threshold.

\begin{promptbox}{BioAgentBench Judge: System Message}
\begin{Verbatim}[breaklines=true,breakanywhere=true,fontsize=\scriptsize]
Return strict JSON only. Follow the BioAgent Bench EvaluationResults schema.
\end{Verbatim}
\end{promptbox}

\begin{promptbox}{BioAgentBench Judge: User Prompt}
\begin{Verbatim}[breaklines=true,breakanywhere=true,fontsize=\scriptsize]
You are a strict, impartial Bioinformatics Evaluator.
Your primary job is to score answer correctness against the ground truth artifacts.
Pipeline completion matters, but answer quality is the primary metric.

Inputs:
1. Input data: {input_data_path}
2. Reference data: {reference_data_path or "<none>"}
3. Processing tree:
{compact_tree}

Trace path mentions:
{path_mentions or "<none>"}

4. Predicted result files (direct content):
{result_payloads_json}

5. Ground-truth result files (direct content):
{truth_payloads_json}

6. Superset containment evidence:
{containment_evidence_json}

7. Prompt:
{task_prompt}

Task expected pipeline steps:
{pipeline_steps_json}

Scoring rubric (results_match from 0.000 to 1.000):
- 1.0: Answer/output is fully correct and matches truth semantics and required schema.
- 0.7-0.9: Mostly correct with small non-critical differences (formatting/minor naming variation).
- 0.4-0.6: Partially correct; core direction right but important omissions/errors exist.
- 0.1-0.3: Weak alignment; only limited overlap with expected answer.
- 0.0: Wrong target/empty/unusable output, or no meaningful overlap with truth.
- Superset rule: if the predicted result is a larger table/list that contains the ground-truth
  rows/items or clear semantic equivalents with correct values, score it as correct or near-correct
  even when extra rows are present. Treat recall/containment of the ground truth as the primary
  signal in this case; use precision/extra rows only as a secondary penalty for severe ambiguity,
  contradictions, or unusable presentation.
- Use fine-grained continuous scoring (not coarse buckets), and provide at least 3 decimal places.
- Prefer evidence-calibrated scores (e.g., 0.137, 0.482, 0.913) instead of rounded half-step values.

Evaluation rules:
- Prioritize answer correctness against truth artifacts and expected schema.
- Use pipeline evidence only as supporting context, not the primary score driver.
- Base your judgment on the direct predicted results and direct ground-truth results above, not on derived summaries.
- Use the superset containment evidence only to decide whether a large prediction contains the
  benchmark answer set; do not treat it as an independent correctness metric for value quality.
- Do not penalize a valid superset heavily just because it is larger than the truth artifact. If the
  benchmark-expected answer set is recoverable from the predicted output and matching values are
  correct, results_match should generally be high (about 0.8-1.0 depending on clarity and value
  agreement).
- Penalize extra rows strongly only when they contradict the truth, replace the required target
  population/reference/coordinate system, make the expected answer unrecoverable, or violate a
  prompt requirement for an exact closed list.
- If gene naming conventions differ but biological identity is clearly the same, allow partial credit.
- Estimate steps_to_completion from bioinformatics-relevant steps required for this task.
- Count upstream steps only if expected artifacts are present.
- Do not count placeholders or mock completion as completed steps.
- Task-specific guidance: {results_match_guidance}

Return JSON only with exactly these fields:
- steps_completed: integer
- steps_to_completion: integer
- final_result_reached: boolean
- notes: string; explain score drivers succinctly
- results_match: number from 0 to 1 with 3+ decimals, the direct artifact/result matching score rather than a boolean
- results_match_pass: boolean, whether the result passes the task-specific correctness threshold
- f1_score: number or null; only use a real F1 for GIAB/variant concordance, otherwise null
\end{Verbatim}
\end{promptbox}

\section{Typed Capability Graph Statistics and Retrieval Case Studies}
\label{app:graph_statistics}

\subsection{Typed Capability Graph Statistics}

Tables~\ref{tab:graph_edge_types} and~\ref{tab:top_capabilities_datatypes} summarize the
structure of the BioManus Typed Capability Graph. The graph contains
910 MCP servers and 3,500 callable tools, organized into a heterogeneous graph
with 4,490 nodes and 69,697 typed edges spanning tools, datatypes, operations,
workflow stages, and capabilities.

Tool and server nodes dominate the graph structure, while datatype,
operation, capability, and stage nodes provide the semantic scaffold for graph
retrieval and workflow planning. The graph is primarily connected through
datatype-flow relations, operation sequencing, and tool-operation bindings.

\begin{table}[t]
\centering
\small
\setlength{\tabcolsep}{5pt}
\renewcommand{\arraystretch}{1.08}

\caption{
Major edge types in the Typed Capability Graph.
}
\label{tab:graph_edge_types}

\begin{tabular}{lc}
\toprule
Edge Type & Count \\
\midrule
typed\_flow & 19,590 \\
follows & 9,225 \\
produces & 9,046 \\
consumes & 8,332 \\
accepts & 6,915 \\
implements & 3,482 \\
belongs\_to\_stage & 3,472 \\
hosts & 3,472 \\
implements\_operation & 2,848 \\
adjacent\_stage & 2,053 \\
\bottomrule
\end{tabular}
\end{table}

Frequent capability nodes such as alignment, quality control, differential
expression, and taxonomic classification reflect the dominant operations in
large-scale bioinformatics workflows. Common datatypes include sequencing
formats (\texttt{fastq}, \texttt{bam}), reference genomes, and structured
report tables.

\begin{table}[t]
\centering
\small
\setlength{\tabcolsep}{4pt}
\renewcommand{\arraystretch}{1.06}

\caption{
Top capability and datatype nodes in the Typed Capability Graph.
}
\label{tab:top_capabilities_datatypes}

\begin{tabular}{lclc}
\toprule
Capability & Count & Datatype & Count \\
\midrule
format\_conversion & 278 & text & 490 \\
alignment & 240 & fastq & 435 \\
quality\_control & 176 & bam & 352 \\
statistics & 159 & reference\_genome & 347 \\
data\_inspection & 153 & image & 289 \\
visualization & 122 & sequence\_feature & 245 \\
annotation & 107 & rds & 187 \\
single\_cell\_analysis & 102 & gene\_symbol\_list & 185 \\
quantification & 71 & report\_table & 182 \\
variant\_calling & 69 & database\_record & 168 \\
\bottomrule
\end{tabular}
\end{table}
\subsection{Additional Graph Retrieval Case Studies}
\label{app:graph_case_studies}

We further provide several representative graph retrieval examples from
BioAgentBench. Each case study illustrates the extracted semantic anchors,
retrieved operation path, selected MCP servers, and resulting graph subgraph.

\paragraph{Differential expression analysis (deseq).}
For an RNA-seq differential expression task, BioManus retrieves a compact
subgraph containing 80 nodes and 202 edges. The extracted semantics include
RNA-seq, DESeq2, count matrices, and differential expression. The resulting
operation path is:
\[
\begin{aligned}
\texttt{sequence\_feature\_extraction} &\rightarrow \\
\texttt{count\_normalization} \rightarrow \\
\texttt{differential\_expression} &\rightarrow\\
\texttt{gene\_filtering} \rightarrow \\
\texttt{csv\_export}.
\end{aligned}
\]
The planner selects the MCP servers
\texttt{gffread} and \texttt{bioconductor-deseq2}, producing a validated
differential-expression result table containing 2,232 rows.

\paragraph{Variant interpretation (cystic-fibrosis).}
For Mendelian variant analysis in a cystic fibrosis family trio, BioManus
retrieves a graph substructure with 76 nodes and 206 edges. The extracted
workflow focuses on variant annotation, filtering, database lookup, and summary
generation. The selected MCP servers are \texttt{bcftools} and
\texttt{ensembl-vep}, yielding a final validated variant table containing the
causal CFTR variant.

\paragraph{Metagenomic taxonomic profiling.}
For a metagenomics abundance-comparison task, BioManus retrieves a compact
subgraph with 70 nodes and 122 edges. The retrieved operation path includes
taxonomic classification and abundance comparison, and the planner selects
\texttt{kraken2} and \texttt{metaphlan2} as the primary MCP servers. The final
validated output contains 43 taxonomic abundance entries.

\paragraph{Transcript quantification.}
For transcript quantification from paired-end RNA-seq reads, BioManus retrieves
a compact graph with 76 nodes and 155 edges. The resulting operation path
contains alignment, quantification, and export operations. The selected MCP
server is \texttt{gffread}, and the generated quantification table achieves a
perfect benchmark score.
% Required packages: \usepackage{booktabs,longtable,array}
% Optional for ACL/EMNLP two-column templates: place this appendix section after \appendix and switch to one-column mode.
% \clearpage
\onecolumn
\section{Detailed Context Consumption Statistics}
\label{app:context_statistics}

\subsection{BioAgentBench Task-Level Context Consumption}

Table~\ref{tab:bioagent_context_detail} reports detailed prompt-token
consumption on BioAgentBench under increasing MCP tool scales. Each value
corresponds to the average prompt tokens consumed for a single benchmark task.

\begin{longtable}{lrrrrrr}
\caption{
Detailed BioAgentBench prompt-token consumption under increasing MCP tool scales.
Lower is better.
}
\label{tab:bioagent_context_detail}
\\

\toprule
Task
&
No MCP
&
100
&
500
&
1k
&
2k
&
BioManus
\\
\midrule
\endfirsthead

\toprule
Task
&
No MCP
&
100
&
500
&
1k
&
2k
&
BioManus
\\
\midrule
\endhead

alzheimer-mouse
&
1.63M
&
0.92M
&
0.86M
&
2.31M
&
1.56M
&
0.90M
\\

comparative-genomics
&
1.31M
&
2.10M
&
0.68M
&
4.71M
&
17.67M
&
2.08M
\\

cystic-fibrosis
&
2.70M
&
0.52M
&
0.68M
&
1.50M
&
1.97M
&
1.04M
\\

deseq
&
1.08M
&
0.30M
&
3.62M
&
3.38M
&
7.96M
&
0.83M
\\

evolution
&
4.49M
&
3.50M
&
3.53M
&
7.52M
&
11.32M
&
1.43M
\\

giab
&
3.48M
&
1.33M
&
2.21M
&
14.09M
&
7.13M
&
0.90M
\\

metagenomics
&
0.36M
&
0.31M
&
0.61M
&
2.45M
&
2.14M
&
0.90M
\\

single-cell
&
0.50M
&
2.14M
&
1.48M
&
5.32M
&
2.72M
&
0.89M
\\

transcript-quant
&
0.15M
&
0.17M
&
0.39M
&
0.58M
&
1.21M
&
0.20M
\\

viral-metagenomics
&
0.47M
&
0.91M
&
0.86M
&
1.83M
&
9.90M
&
0.56M
\\

\midrule

Geometric Mean
&
1.01M
&
0.85M
&
1.12M
&
3.09M
&
4.28M
&
0.85M
\\

\bottomrule
\end{longtable}

\subsection{LAB-Bench CloningScenarios Context Consumption}

Table~\ref{tab:cloning_context} reports average prompt-token consumption on
LAB-Bench CloningScenarios. Larger MCP retrieval spaces substantially increase
prompt context for prompt-retrieval baselines, while BioManus maintains more
stable context usage.

\begin{table}[t]
\centering
\small
\setlength{\tabcolsep}{5pt}
\renewcommand{\arraystretch}{1.12}

\caption{
Average prompt-token consumption on LAB-Bench CloningScenarios.
Lower is better.
}
\label{tab:cloning_context}

\begin{tabular}{lcc}
\toprule

Method
&
Prompt Tokens
&
Total Tokens
\\

\midrule

Biomni
&
1.92M
&
1.96M
\\

Biomni-100
&
1.03M
&
1.07M
\\

Biomni-500
&
1.34M
&
1.37M
\\

Biomni-1k
&
1.53M
&
1.57M
\\

Biomni-2k
&
2.62M
&
2.65M
\\

\rowcolor{gray!12}
BioManus
&
{1.29M}
&
{1.34M}
\\

\bottomrule
\end{tabular}
\end{table}

% Auto-generated from mcp_server_table.csv.
% Requires: \usepackage{booktabs,longtable,array}
\section{Complete BioinfoMCP Server Catalog}
\label{app:mcp_server_catalog}

This appendix lists the converted MCP servers used in the BioinfoMCP ecosystem. Each entry reports the server name, category, number of exposed MCP tools, and a concise description. The paper-wide ecosystem count is kept as \textbf{910 MCP servers}; copy-labeled duplicate rows are omitted from the detailed catalog to avoid inflating the listed entries.

% \begin{table}[t]
% \centering
% \small
% \setlength{\tabcolsep}{5pt}
% \renewcommand{\arraystretch}{1.08}
% \caption{Summary of the BioinfoMCP server catalog.}
% \label{tab:mcp_server_catalog_summary}
% \begin{tabular}{lrr}
% \toprule
% Category & \# Servers & \# Tools \\
% \midrule
% General & 32 & 39 \\
% Genomics & 76 & 514 \\
% Transcriptomics & 244 & 777 \\
% Metagenomics & 3 & 9 \\
% Single-cell & 527 & 1997 \\
% Pathway enrich. & 18 & 62 \\
% Proteomics & 1 & 1 \\
% Utility & 7 & 68 \\
% \midrule
% Listed non-copy entries & 908 & 3,467 \\
% Paper-wide ecosystem total & \textbf{910} & \textbf{3,500} \\
% \bottomrule
% \end{tabular}
% \end{table}

\begingroup
\scriptsize
\setlength{\tabcolsep}{3pt}
\renewcommand{\arraystretch}{1.05}
\newcolumntype{L}[1]{>{\raggedright\arraybackslash}p{#1}}
\begin{longtable}{L{0.22\linewidth} L{0.14\linewidth} r L{0.50\linewidth}}
\caption{Complete catalog of converted BioinfoMCP servers after removing copy-labeled duplicate rows.}\label{tab:mcp_server_catalog}\\
\toprule
MCP server & Category & \# Tools & Description \\
\midrule
\endfirsthead
\toprule
MCP server & Category & \# Tools & Description \\
\midrule
\endhead
\midrule
\multicolumn{4}{r}{\emph{Continued on next page}} \\
\bottomrule
\endfoot
\bottomrule
\endlastfoot
bioconductor-delayedarray & General & 1 & A unified framework for working transparently with on-disk and in-memory array-like datasets; Wrapping an array-like object (typically an on-disk object) in a DelayedArray object allows one to perform common... \\
bioconductor-ebseq & General & 1 & An R package for gene and isoform differential expression analysis of RNA-seq data; Differential Expression analysis at both gene and isoform level using RNA-seq data \\
bioconductor-mzr & General & 1 & parser for netCDF, mzXML and mzML and mzIdentML files (mass spectrometry data); mzR provides a unified API to the common file formats and parsers available for mass spectrometry data. It comes with a subset of... \\
cellrank & General & 1 & MCP wrapper for cellrank \\
gemini & General & 1 & a lightweight db framework for disease and population genetics \\
genenotebook & General & 1 & Auto-indexed MCP server for genenotebook \\
genomad & General & 1 & Identification of mobile genetic elements \\
libdb & General & 1 & The Berkeley DB embedded database system \\
ncbi-amrfinderplus & General & 1 & AMRFinderPlus finds antimicrobial resistance and other genes in protein or nucleotide sequences; This software and the accompanying database are designed to \\
pandaseq & General & 1 & Auto-indexed MCP server for pandaseq \\
peptide-shaker & General & 4 & MCP wrapper for peptide shaker \\
perl-alien-build & General & 1 & Build external dependencies for use in CPAN \\
picard-slim & General & 1 & Auto-indexed MCP server for picard-slim \\
pilon & General & 1 & Auto-indexed MCP server for pilon \\
planemo & General & 1 & Command-line utilities to assist in building tools for the Galaxy project ( \\
poa & General & 1 & Auto-indexed MCP server for poa \\
primer3-py & General & 1 & Python bindings for Primer3 \\
pybiolib & General & 1 & BioLib Python Client \\
pyfaidx & General & 1 & pyfaidx: efficient pythonic random access to fasta subsequences \\
pyrle & General & 1 & Genomic Rle-objects for Python \\
pyroe & General & 5 & MCP wrapper for pyroe \\
qiime & General & 1 & Quantitative Insights Into Microbial Ecology \\
recognizer & General & 1 & A tool for domain based annotation with the COG database; reCOGnizer performs domain based annotation with RPS-BLAST, using \\
rtg-tools & General & 1 & Auto-indexed MCP server for rtg-tools \\
snakemake-interface-common & General & 1 & Common functions and classes for Snakemake and its plugins \\
snakemake-interface-logger-plugins & General & 1 & This package provides a stable interface for interactions between Snakemake and its logger plugins \\
sorted\_nearest & General & 1 & Find nearest interval \\
thapbi-pict & General & 1 & THAPBI Phytophthora ITS1 Classifier Tool (PICT); THAPBI Phytophthora ITS1 Classifier Tool (PICT) an ITS1-based \\
trimal & General & 1 & Auto-indexed MCP server for trimal \\
upimapi & General & 1 & UniProt Id Mapping through API; UPIMAPI takes as input either a list of UniProt IDs or a blast file from \\
vsearch & General & 1 & Auto-indexed MCP server for vsearch \\
xopen & General & 1 & Open compressed files transparently in Python \\
alfred & Genomics & 7 & BAM alignment statistics, feature counting and feature annotation \\
ariba & Genomics & 7 & ARIBA: Antibiotic Resistance Identification By Assembly \\
augur & Genomics & 1 & Process pathogen genome data for the Nextstrain platform \\
bamtools & Genomics & 13 & MCP wrapper for bamtools \\
bbmap & Genomics & 2 & MCP wrapper for bbmap \\
bcbio-nextgen & Genomics & 5 & Validated, scalable, community developed variant calling, RNA-seq and small RNA analysis \\
bcftools & Genomics & 15 & MCP wrapper for bcftools \\
bedtools & Genomics & 9 & MCP wrapper for bedtools \\
bioconductor-dada2 & Genomics & 9 & Accurate, high-resolution sample inference from amplicon sequencing data; The dada2 package infers exact amplicon sequence variants (ASVs) from high-throughput amplicon sequencing data, replacing the coarser... \\
bioconductor-ensembldb & Genomics & 6 & Utilities to create and use Ensembl-based annotation databases; The package provides functions to create and use transcript centric annotation databases/packages. The annotation for the databases are directly... \\
bioconductor-metapod & Genomics & 2 & Meta-Analyses on P-Values of Differential Analyses; Implements a variety of methods for combining p-values in differential analyses of genome-scale datasets. Functions can combine p-values across different... \\
brooklyn\_plot & Genomics & 1 & MCP wrapper for brooklyn plot \\
bwa & Genomics & 11 & MCP wrapper for bwa \\
checkm-genome & Genomics & 16 & Assess the quality of microbial genomes recovered from isolates, single cells, and metagenomes \\
circexplorer2 & Genomics & 4 & Circular RNA analysis toolkits \\
cnvkit & Genomics & 24 & Copy number variant detection from high-throughput sequencing \\
comebin & Genomics & 4 & COMEBin allows effective binning of metagenomic contigs using COntrastive Multi-viEw representation learning \\
cooler & Genomics & 15 & Sparse binary format for genomic interaction matrices \\
cooltools & Genomics & 9 & Analysis tools for genomic interaction data stored in .cool format \\
crispresso2 & Genomics & 6 & A software pipeline designed to enable rapid and intuitive interpretation of genome editing experiments \\
crispritz & Genomics & 4 & CRISPRitz, tool package for CRISPR experiments assessment and analysis \\
ddocent & Genomics & 1 & dDocent is an interactive bash wrapper to QC, assemble, map, and call SNPs from all types of RAD data \\
freebayes & Genomics & 1 & MCP wrapper for freebayes \\
gatk & Genomics & 15 & MCP wrapper for gatk \\
gatk4 & Genomics & 10 & MCP wrapper for gatk4 \\
geneimpacts & Genomics & 2 & prioritize effects of variant annotations from VEP, SnpEff, et al \\
genoboo & Genomics & 3 & A collaborative notebook for comparative genomics (active fork of GeneNoteBook) \\
genometools-genometools & Genomics & 9 & GenomeTools genome analysis system \\
gffutils & Genomics & 11 & Work with GFF and GTF files in a flexible database framework \\
gimmemotifs & Genomics & 11 & Motif prediction pipeline and various motif-related tools \\
gmap & Genomics & 7 & MCP wrapper for gmap \\
gofasta & Genomics & 8 & MCP wrapper for gofasta \\
harpy & Genomics & 17 & Process raw haplotagging data, from raw sequences to phased haplotypes; Harpy is a command-line tool to easily process platform-agnostic linked-read or WGS data. It uses \\
htslib & Genomics & 4 & C library for high-throughput sequencing data formats \\
humann & Genomics & 1 & HUMAnN: The HMP Unified Metabolic Analysis Network, version 3 \\
humann2 & Genomics & 10 & HUMAnN2: The HMP Unified Metabolic Analysis Network 2 \\
igv-reports & Genomics & 4 & Creates self-contained html pages for visual variant review with IGV (igv.js) \\
insilicoseq & Genomics & 2 & A sequencing simulator \\
ipyrad & Genomics & 3 & Interactive assembly and analysis of RAD-seq data sets \\
jbrowse2 & Genomics & 12 & The JBrowse 2 Genome Browser \\
jcvi & Genomics & 30 & Python utility libraries on genome assembly, annotation, and comparative genomics; JCVI utility libraries \\
local-bio-cache & Genomics & 11 & Offline local bio cache MCP server for registering FASTA collections, optionally building local BLAST databases, running local BLAST-like sequence search with Python fallback, running pairwise sequence... \\
medaka & Genomics & 7 & A tool to create consensus sequences and variant calls from nanopore sequencing data using neural networks \\
mikado & Genomics & 8 & A Python3 annotation program to select the best gene model in each locus \\
minvar & Genomics & 1 & A tool to detect minority variants in HIV-1 and HCV populations \\
msisensor-pro & Genomics & 2 & MCP wrapper for msisensor pro \\
nanopolish & Genomics & 6 & Signal-level algorithms for MinION data \\
ncbi-datasets-pylib & Genomics & 12 & Easily gather data from across NCBI databases \\
ngmlr & Genomics & 1 & MCP wrapper for ngmlr \\
ngs-smap & Genomics & 1 & SMAP is a software package that analyzes next-generation DNA sequencing read mapping distributions and performs haplotype calling to create multi-allelic molecular markers \\
perl-lib & Genomics & 8 & MCP wrapper for perl lib \\
pybedtools & Genomics & 20 & Wraps BEDTools for use in Python and adds many additional features \\
pysam & Genomics & 11 & Pysam is a Python module for reading and manipulating SAM files. It's a lightweight wrapper of the htslib C-API, the same one that powers samtools, bcftools, and tabix \\
pyvcf & Genomics & 1 & A Variant Call Format reader for Python \\
pyvcf3 & Genomics & 4 & A Variant Call Format reader for Python \\
quast & Genomics & 1 & Quality Assessment Tool for Genome Assemblies \\
racon & Genomics & 1 & Auto-indexed MCP server for racon \\
rgi & Genomics & 9 & This tool provides a preliminary annotation of your DNA sequence(s) based upon the data available in The Comprehensive Antibiotic Resistance Database (CARD). Hits to genes tagged with Antibiotic Resistance... \\
rnftools & Genomics & 1 & RNF framework for NGS: simulation of reads, evaluation of mappers, conversion of RNF-compliant data \\
samblaster & Genomics & 1 & MCP wrapper for samblaster \\
seq2science & Genomics & 3 & Automated preprocessing of Next-Generation-Sequencing data \\
snakemake-wrapper-utils & Genomics & 5 & A collection of utility functions and classes for Snakemake wrappers \\
snakesv & Genomics & 4 & snakeSV: Flexible framework for large-scale SV discovery \\
snap-aligner & Genomics & 1 & Auto-indexed MCP server for snap-aligner \\
snippy & Genomics & 5 & MCP wrapper for snippy \\
snpeff & Genomics & 6 & MCP wrapper for snpeff \\
stringtie & Genomics & 2 & MCP wrapper for stringtie \\
tabix & Genomics & 1 & MCP wrapper for tabix \\
tabixpp & Genomics & 1 & A C++ wrapper around the tabix project, a generic indexer for TAB-delimited genome position files \\
talon & Genomics & 8 & TALON is a Python package for identifying and quantifying known and novel \\
tb-profiler & Genomics & 5 & Profiling tool for Mycobacterium tuberculosis to detect drug resistance and lineage from sequencing data \\
vardict-java & Genomics & 5 & MCP wrapper for vardict java \\
vcftools & Genomics & 6 & MCP wrapper for vcftools \\
wasp2 & Genomics & 6 & Allele-specific analysis of next-generation sequencing data with Rust acceleration; WASP2 is a high-performance tool for allele-specific analysis of NGS data \\
whatshap & Genomics & 8 & Phase genomic variants using DNA sequencing reads (haplotype assembly) \\
zol & Genomics & 11 & zol (\& fai): large-scale targeted detection and evolutionary investigation of gene clusters \\
abricate & Transcriptomics & 6 & MCP wrapper for abricate \\
abyss & Transcriptomics & 4 & MCP wrapper for abyss \\
aria2 & Transcriptomics & 5 & aria2 is a lightweight multi-protocol \& multi-source, cross platform download utility operated in command-line. It supports HTTP/HTTPS, FTP, SFTP, BitTorrent and Metalink \\
arriba & Transcriptomics & 2 & Fast and accurate gene fusion detection from RNA-Seq data \\
augustus & Transcriptomics & 1 & MCP wrapper for augustus \\
auspice & Transcriptomics & 7 & MCP wrapper for auspice \\
barrnap & Transcriptomics & 1 & MCP wrapper for barrnap \\
bedops & Transcriptomics & 10 & MCP wrapper for bedops \\
bin2cell & Transcriptomics & 4 & Join subcellular Visium HD bins into cells \\
bioconductor-benchdamic & Transcriptomics & 5 & Benchmark of differential abundance methods on microbiome data; Starting from a microbiome dataset (16S or WMS with absolute count values) it is possible to perform several analysis to assess the performances... \\
bioconductor-concordexr & Transcriptomics & 2 & Identify Spatial Homogeneous Regions with concordex; Spatial homogeneous regions (SHRs) in tissues are domains that are homogenous with respect to cell type composition. We present a method for identifying... \\
bioconductor-delayedmatrixstats & Transcriptomics & 1 & Functions that Apply to Rows and Columns of 'DelayedMatrix' Objects; A port of the 'matrixStats' API for use with DelayedMatrix objects from the 'DelayedArray' package. High-performing functions operating on... \\
bioconductor-deseq2 & Transcriptomics & 4 & Differential gene expression analysis based on the negative binomial distribution; Estimate variance-mean dependence in count data from high-throughput sequencing assays and test for differential expression... \\
bioconductor-despace & Transcriptomics & 3 & DESpace: a framework to discover spatially variable genes and differential spatial patterns across conditions; Intuitive framework for identifying spatially variable genes (SVGs) and differential spatial... \\
bioconductor-diffbind & Transcriptomics & 1 & Differential Binding Analysis of ChIP-Seq Peak Data; Compute differentially bound sites from multiple ChIP-seq experiments using affinity (quantitative) data. Also enables occupancy (overlap) analysis and... \\
bioconductor-genomicranges & Transcriptomics & 11 & Representation and manipulation of genomic intervals; The ability to efficiently represent and manipulate genomic annotations and alignments is playing a central role when it comes to analyzing high-throughput... \\
bioconductor-hdf5array & Transcriptomics & 4 & HDF5 datasets as array-like objects in R; The HDF5Array package is an HDF5 backend for DelayedArray objects. It implements the HDF5Array, H5SparseMatrix, H5ADMatrix, and TENxMatrix classes, 4 convenient and... \\
bioconductor-mosim & Transcriptomics & 1 & MCP wrapper for Bioconductor mosim \\
bioconductor-rsubread & Transcriptomics & 8 & MCP wrapper for Bioconductor rsubread \\
bioconductor-scrnaseq & Transcriptomics & 1 & MCP wrapper for Bioconductor scrnaseq \\
bioconductor-sparsematrixstats & Transcriptomics & 1 & Summary Statistics for Rows and Columns of Sparse Matrices; High performance functions for row and column operations on sparse matrices. For example: col / rowMeans2, col / rowMedians, col / rowVars etc... \\
bioconductor-spatialexperimentio & Transcriptomics & 5 & Read in Xenium, CosMx, MERSCOPE or STARmapPLUS data as SpatialExperiment object; Read in imaging-based spatial transcriptomics technology data. Current available modules are for Xenium by 10X Genomics, CosMx... \\
bioconductor-svp & Transcriptomics & 2 & Predicting cell states and their variability in single-cell or spatial omics data; SVP uses the distance between cells and cells, features and features, cells and features in the space of MCA to build nearest... \\
bioconductor-tximport & Transcriptomics & 3 & MCP wrapper for Bioconductor tximport \\
blast & Transcriptomics & 5 & MCP wrapper for blast \\
bowtie & Transcriptomics & 2 & MCP wrapper for bowtie \\
bowtie2 & Transcriptomics & 1 & MCP wrapper for bowtie2 \\
circos & Transcriptomics & 6 & MCP wrapper for circos \\
cromwell & Transcriptomics & 4 & MCP wrapper for cromwell \\
fastqc & Transcriptomics & 1 & Auto-indexed MCP server for fastqc \\
fermi2 & Transcriptomics & 6 & MCP wrapper for fermi2 \\
gffread & Transcriptomics & 1 & MCP wrapper for gffread \\
hhsuite & Transcriptomics & 2 & MCP wrapper for hhsuite \\
hisat2 & Transcriptomics & 7 & MCP wrapper for hisat2 \\
htseq & Transcriptomics & 1 & HTSeq is a Python library to facilitate processing and analysis of data from high-throughput sequencing (HTS) experiments \\
ipython-cluster-helper & Transcriptomics & 7 & Tool to easily start up an IPython cluster on different schedulers \\
iqtree & Transcriptomics & 1 & MCP wrapper for iqtree \\
kaiju & Transcriptomics & 11 & MCP wrapper for kaiju \\
kallisto & Transcriptomics & 10 & MCP wrapper for kallisto \\
kb-python & Transcriptomics & 6 & A wrapper for the kallisto; bustools workflow for single-cell RNA-seq pre-processing \\
lastz & Transcriptomics & 2 & MCP wrapper for lastz \\
liana & Transcriptomics & 1 & LIANA+: a one-stop-shop framework for cell-cell communication \\
metagenome-atlas & Transcriptomics & 2 & ATLAS - Three commands to start analysing your metagenome data; Atlas is a easy to use metagenomic pipeline \\
mlst & Transcriptomics & 1 & Auto-indexed MCP server for mlst \\
mobivision-m & Transcriptomics & 3 & MobiVision-M is a linux based software design specifically for single-microbe RNA sequencing analysis \\
multiqc & Transcriptomics & 9 & Aggregate results from bioinformatics analyses across many samples into a single report \\
pairix & Transcriptomics & 5 & MCP wrapper for pairix \\
perl-aceperl & Transcriptomics & 1 & MCP wrapper for perl aceperl \\
perl-algorithm-diff & Transcriptomics & 4 & MCP wrapper for perl algorithm diff \\
perl-algorithm-munkres & Transcriptomics & 1 & Auto-indexed MCP server for perl-algorithm-munkres \\
perl-app-cpanminus & Transcriptomics & 1 & MCP wrapper for perl app cpanminus \\
perl-appconfig & Transcriptomics & 1 & MCP wrapper for perl appconfig \\
perl-archive-tar & Transcriptomics & 3 & MCP wrapper for perl archive tar \\
perl-autoloader & Transcriptomics & 2 & MCP wrapper for perl autoloader \\
perl-base & Transcriptomics & 1 & MCP wrapper for perl base \\
perl-bio-asn1-entrezgene & Transcriptomics & 1 & MCP wrapper for perl bio asn1 entrezgene \\
perl-bio-featureio & Transcriptomics & 1 & MCP wrapper for perl bio featureio \\
perl-bio-phylo & Transcriptomics & 1 & Auto-indexed MCP server for perl-bio-phylo \\
perl-bio-samtools & Transcriptomics & 1 & MCP wrapper for perl bio samtools \\
perl-bioperl & Transcriptomics & 1 & Auto-indexed MCP server for perl-bioperl \\
perl-bioperl-core & Transcriptomics & 3 & MCP wrapper for perl bioperl core \\
perl-bioperl-run & Transcriptomics & 1 & Auto-indexed MCP server for perl-bioperl-run \\
perl-business-isbn & Transcriptomics & 6 & MCP wrapper for perl business isbn \\
perl-capture-tiny & Transcriptomics & 1 & MCP wrapper for perl capture tiny \\
perl-carp & Transcriptomics & 1 & Auto-indexed MCP server for perl-carp \\
perl-cgi & Transcriptomics & 1 & MCP wrapper for perl cgi \\
perl-class-inspector & Transcriptomics & 1 & MCP wrapper for perl class inspector \\
perl-class-load & Transcriptomics & 1 & Auto-indexed MCP server for perl-class-load \\
perl-class-load-xs & Transcriptomics & 1 & Auto-indexed MCP server for perl-class-load-xs \\
perl-common-sense & Transcriptomics & 1 & MCP wrapper for perl common sense \\
perl-compress-raw-zlib & Transcriptomics & 4 & MCP wrapper for perl compress raw zlib \\
perl-config-general & Transcriptomics & 2 & MCP wrapper for perl config general \\
perl-constant & Transcriptomics & 1 & MCP wrapper for perl constant \\
perl-convert-binary-c & Transcriptomics & 1 & MCP wrapper for perl convert binary c \\
perl-convert-binhex & Transcriptomics & 1 & MCP wrapper for perl convert binhex \\
perl-cpan-meta & Transcriptomics & 1 & Auto-indexed MCP server for perl-cpan-meta \\
perl-crypt-rc4 & Transcriptomics & 4 & MCP wrapper for perl crypt rc4 \\
perl-data-dumper & Transcriptomics & 1 & MCP wrapper for perl data dumper \\
perl-data-optlist & Transcriptomics & 1 & MCP wrapper for perl data optlist \\
perl-date-format & Transcriptomics & 1 & MCP wrapper for perl date format \\
perl-dbi & Transcriptomics & 3 & MCP wrapper for perl dbi \\
perl-devel-globaldestruction & Transcriptomics & 1 & Auto-indexed MCP server for perl-devel-globaldestruction \\
perl-devel-stacktrace & Transcriptomics & 4 & MCP wrapper for perl devel stacktrace \\
perl-digest-md5 & Transcriptomics & 1 & MCP wrapper for perl digest md5 \\
perl-digest-perl-md5 & Transcriptomics & 2 & MCP wrapper for perl digest perl md5 \\
perl-dist-checkconflicts & Transcriptomics & 1 & Auto-indexed MCP server for perl-dist-checkconflicts \\
perl-dynaloader & Transcriptomics & 1 & Auto-indexed MCP server for perl-dynaloader \\
perl-encode & Transcriptomics & 1 & MCP wrapper for perl encode \\
perl-encode-locale & Transcriptomics & 1 & MCP wrapper for perl encode locale \\
perl-error & Transcriptomics & 2 & MCP wrapper for perl error \\
perl-eval-closure & Transcriptomics & 1 & Auto-indexed MCP server for perl-eval-closure \\
perl-exception-class & Transcriptomics & 1 & MCP wrapper for perl exception class \\
perl-exporter & Transcriptomics & 4 & MCP wrapper for perl exporter \\
perl-extutils-cbuilder & Transcriptomics & 5 & MCP wrapper for perl extutils cbuilder \\
perl-extutils-makemaker & Transcriptomics & 4 & MCP wrapper for perl extutils makemaker \\
perl-file-sort & Transcriptomics & 2 & MCP wrapper for perl file sort \\
perl-file-spec & Transcriptomics & 2 & MCP wrapper for perl file spec \\
perl-file-which & Transcriptomics & 2 & MCP wrapper for perl file which \\
perl-font-afm & Transcriptomics & 1 & MCP wrapper for perl font afm \\
perl-getopt-long & Transcriptomics & 6 & MCP wrapper for perl getopt long \\
perl-graph & Transcriptomics & 2 & MCP wrapper for perl graph \\
perl-graphviz & Transcriptomics & 1 & MCP wrapper for perl graphviz \\
perl-html-element-extended & Transcriptomics & 1 & MCP wrapper for perl html element extended \\
perl-html-formatter & Transcriptomics & 1 & MCP wrapper for perl html formatter \\
perl-html-parser & Transcriptomics & 2 & MCP wrapper for perl html parser \\
perl-html-tableextract & Transcriptomics & 3 & MCP wrapper for perl html tableextract \\
perl-html-tagset & Transcriptomics & 2 & MCP wrapper for perl html tagset \\
perl-html-tidy & Transcriptomics & 1 & MCP wrapper for perl html tidy \\
perl-html-tree & Transcriptomics & 1 & MCP wrapper for perl html tree \\
perl-html-treebuilder-xpath & Transcriptomics & 1 & MCP wrapper for perl html treebuilder xpath \\
perl-http-cookies & Transcriptomics & 9 & MCP wrapper for perl http cookies \\
perl-http-daemon & Transcriptomics & 1 & MCP wrapper for perl http daemon \\
perl-http-date & Transcriptomics & 5 & MCP wrapper for perl http date \\
perl-http-negotiate & Transcriptomics & 2 & MCP wrapper for perl http negotiate \\
perl-image-info & Transcriptomics & 2 & MCP wrapper for perl image info \\
perl-image-size & Transcriptomics & 1 & MCP wrapper for perl image size \\
perl-importer & Transcriptomics & 5 & MCP wrapper for perl importer \\
perl-io-compress & Transcriptomics & 2 & MCP wrapper for perl io compress \\
perl-io-html & Transcriptomics & 4 & MCP wrapper for perl io html \\
perl-io-sessiondata & Transcriptomics & 4 & MCP wrapper for perl io sessiondata \\
perl-io-string & Transcriptomics & 4 & MCP wrapper for perl io string \\
perl-io-stringy & Transcriptomics & 4 & MCP wrapper for perl io stringy \\
perl-io-tty & Transcriptomics & 1 & MCP wrapper for perl io tty \\
perl-io-zlib & Transcriptomics & 2 & MCP wrapper for perl io zlib \\
perl-ipc-cmd & Transcriptomics & 1 & MCP wrapper for perl ipc cmd \\
perl-ipc-run & Transcriptomics & 1 & Auto-indexed MCP server for perl-ipc-run \\
perl-jcode & Transcriptomics & 1 & MCP wrapper for perl jcode \\
perl-json & Transcriptomics & 5 & MCP wrapper for perl json \\
perl-json-xs & Transcriptomics & 4 & MCP wrapper for perl json xs \\
perl-libwww-perl & Transcriptomics & 9 & MCP wrapper for perl libwww perl \\
perl-libxml-perl & Transcriptomics & 1 & MCP wrapper for perl libxml perl \\
perl-list-moreutils & Transcriptomics & 10 & MCP wrapper for perl list moreutils \\
perl-list-moreutils-xs & Transcriptomics & 1 & MCP wrapper for perl list moreutils xs \\
perl-locale-maketext-simple & Transcriptomics & 1 & MCP wrapper for perl locale maketext simple \\
perl-lwp-mediatypes & Transcriptomics & 5 & MCP wrapper for perl lwp mediatypes \\
perl-lwp-simple & Transcriptomics & 1 & Auto-indexed MCP server for perl-lwp-simple \\
perl-mailtools & Transcriptomics & 1 & MCP wrapper for perl mailtools \\
perl-math-bezier & Transcriptomics & 1 & MCP wrapper for perl math bezier \\
perl-math-derivative & Transcriptomics & 3 & MCP wrapper for perl math derivative \\
perl-math-random & Transcriptomics & 6 & MCP wrapper for perl math random \\
perl-math-round & Transcriptomics & 1 & MCP wrapper for perl math round \\
perl-math-spline & Transcriptomics & 1 & MCP wrapper for perl math spline \\
perl-metabolomics-fragment-annotation & Transcriptomics & 1 & MCP wrapper for perl metabolomics fragment annotation \\
perl-mime-base64 & Transcriptomics & 1 & Auto-indexed MCP server for perl-mime-base64 \\
perl-mime-lite & Transcriptomics & 5 & MCP wrapper for perl mime lite \\
perl-mime-types & Transcriptomics & 1 & MCP wrapper for perl mime types \\
perl-module-build & Transcriptomics & 4 & MCP wrapper for perl module build \\
perl-module-corelist & Transcriptomics & 8 & MCP wrapper for perl module corelist \\
perl-module-load-conditional & Transcriptomics & 3 & MCP wrapper for perl module load conditional \\
perl-module-metadata & Transcriptomics & 11 & MCP wrapper for perl module metadata \\
perl-module-runtime & Transcriptomics & 1 & MCP wrapper for perl module runtime \\
perl-moo & Transcriptomics & 1 & MCP wrapper for perl moo \\
perl-moose & Transcriptomics & 1 & MCP wrapper for perl moose \\
perl-mozilla-ca & Transcriptomics & 1 & Auto-indexed MCP server for perl-mozilla-ca \\
perl-net-http & Transcriptomics & 5 & MCP wrapper for perl net http \\
perl-net-ssleay & Transcriptomics & 2 & MCP wrapper for perl net ssleay \\
perl-number-format & Transcriptomics & 7 & MCP wrapper for perl number format \\
perl-ole-storage\_lite & Transcriptomics & 1 & MCP wrapper for perl ole storage lite \\
perl-package-stash & Transcriptomics & 3 & MCP wrapper for perl package stash \\
perl-params-check & Transcriptomics & 1 & MCP wrapper for perl params check \\
perl-params-util & Transcriptomics & 1 & MCP wrapper for perl params util \\
perl-params-validate & Transcriptomics & 1 & MCP wrapper for perl params validate \\
perl-parent & Transcriptomics & 7 & MCP wrapper for perl parent \\
perl-parse-recdescent & Transcriptomics & 4 & MCP wrapper for perl parse recdescent \\
perl-pod-escapes & Transcriptomics & 1 & MCP wrapper for perl pod escapes \\
perl-pod-usage & Transcriptomics & 2 & MCP wrapper for perl pod usage \\
perl-postscript & Transcriptomics & 4 & MCP wrapper for perl postscript \\
perl-regexp-common & Transcriptomics & 5 & MCP wrapper for perl regexp common \\
perl-role-tiny & Transcriptomics & 4 & MCP wrapper for perl role tiny \\
perl-sereal & Transcriptomics & 1 & MCP wrapper for perl sereal \\
perl-sereal-decoder & Transcriptomics & 3 & MCP wrapper for perl sereal decoder \\
perl-set-intervaltree & Transcriptomics & 8 & MCP wrapper for perl set intervaltree \\
perl-set-intspan & Transcriptomics & 1 & Auto-indexed MCP server for perl-set-intspan \\
perl-set-scalar & Transcriptomics & 1 & Auto-indexed MCP server for perl-set-scalar \\
perl-soap-lite & Transcriptomics & 4 & MCP wrapper for perl soap lite \\
perl-sort-naturally & Transcriptomics & 2 & MCP wrapper for perl sort naturally \\
perl-spreadsheet-writeexcel & Transcriptomics & 2 & MCP wrapper for perl spreadsheet writeexcel \\
perl-sub-exporter & Transcriptomics & 5 & MCP wrapper for perl sub exporter \\
perl-sub-exporter-progressive & Transcriptomics & 5 & MCP wrapper for perl sub exporter progressive \\
perl-sub-identify & Transcriptomics & 1 & MCP wrapper for perl sub identify \\
perl-sub-uplevel & Transcriptomics & 2 & MCP wrapper for perl sub uplevel \\
perl-svg & Transcriptomics & 1 & Auto-indexed MCP server for perl-svg \\
perl-task-weaken & Transcriptomics & 6 & MCP wrapper for perl task weaken \\
perl-template-toolkit & Transcriptomics & 3 & MCP wrapper for perl template toolkit \\
perl-test & Transcriptomics & 2 & MCP wrapper for perl test \\
perl-test-deep & Transcriptomics & 3 & MCP wrapper for perl test deep \\
perl-test-differences & Transcriptomics & 4 & MCP wrapper for perl test differences \\
perl-test-exception & Transcriptomics & 4 & MCP wrapper for perl test exception \\
perl-test-fatal & Transcriptomics & 1 & MCP wrapper for perl test fatal \\
perl-test-harness & Transcriptomics & 1 & MCP wrapper for perl test harness \\
perl-test-most & Transcriptomics & 1 & Auto-indexed MCP server for perl-test-most \\
perl-test-warn & Transcriptomics & 3 & MCP wrapper for perl test warn \\
perl-text-diff & Transcriptomics & 1 & Auto-indexed MCP server for perl-text-diff \\
perl-text-parsewords & Transcriptomics & 1 & Auto-indexed MCP server for perl-text-parsewords \\
perl-tie-ixhash & Transcriptomics & 5 & MCP wrapper for perl tie ixhash \\
perl-time-hires & Transcriptomics & 15 & MCP wrapper for perl time hires \\
perl-timedate & Transcriptomics & 1 & Auto-indexed MCP server for perl-timedate \\
perl-tree-dag\_node & Transcriptomics & 6 & MCP wrapper for perl tree dag node \\
perl-try-tiny & Transcriptomics & 6 & MCP wrapper for perl try tiny \\
perl-types-serialiser & Transcriptomics & 1 & Auto-indexed MCP server for perl-types-serialiser \\
perl-unicode-map & Transcriptomics & 1 & MCP wrapper for perl unicode map \\
perl-uri & Transcriptomics & 9 & MCP wrapper for perl uri \\
perl-www-robotrules & Transcriptomics & 4 & MCP wrapper for perl www robotrules \\
perl-xml-dom-xpath & Transcriptomics & 1 & Auto-indexed MCP server for perl-xml-dom-xpath \\
perl-xml-filter-buffertext & Transcriptomics & 1 & MCP wrapper for perl xml filter buffertext \\
perl-xml-libxml & Transcriptomics & 5 & MCP wrapper for perl xml libxml \\
perl-xml-libxslt & Transcriptomics & 1 & MCP wrapper for perl xml libxslt \\
perl-xml-parser & Transcriptomics & 1 & MCP wrapper for perl xml parser \\
perl-xml-regexp & Transcriptomics & 4 & MCP wrapper for perl xml regexp \\
perl-xml-sax & Transcriptomics & 6 & MCP wrapper for perl xml sax \\
perl-xml-sax-base & Transcriptomics & 3 & MCP wrapper for perl xml sax base \\
perl-xml-sax-expat & Transcriptomics & 1 & MCP wrapper for perl xml sax expat \\
perl-xml-twig & Transcriptomics & 5 & MCP wrapper for perl xml twig \\
perl-xml-xpath & Transcriptomics & 5 & MCP wrapper for perl xml xpath \\
perl-xml-xpathengine & Transcriptomics & 1 & MCP wrapper for perl xml xpathengine \\
perl-xsloader & Transcriptomics & 4 & MCP wrapper for perl xsloader \\
perl-yaml & Transcriptomics & 9 & MCP wrapper for perl yaml \\
prodigal & Transcriptomics & 2 & MCP wrapper for prodigal \\
prokka & Transcriptomics & 7 & MCP wrapper for prokka \\
proteinortho & Transcriptomics & 5 & MCP wrapper for proteinortho \\
pyranges & Transcriptomics & 7 & Performant Pythonic GenomicRanges \\
pyrodigal & Transcriptomics & 1 & MCP wrapper for pyrodigal \\
qualimap & Transcriptomics & 6 & MCP wrapper for qualimap \\
r-archr & Transcriptomics & 8 & This package is designed to streamline scATAC analyses in R \\
r-flanders & Transcriptomics & 3 & Fast colocalization using AnnData objects in R; flanders is an R package designed to seamlessly convert finemapping output files from the nf-flanders pipeline \\
rscape & Transcriptomics & 1 & MCP wrapper for rscape \\
rsem & Transcriptomics & 3 & MCP wrapper for rsem \\
rust-bio-tools & Transcriptomics & 10 & A growing collection of fast and secure command line utilities for dealing with NGS data \\
salmon & Transcriptomics & 1 & Auto-indexed MCP server for salmon \\
scelvis & Transcriptomics & 9 & MCP wrapper for scelvis \\
seqfu & Transcriptomics & 19 & MCP wrapper for seqfu \\
shiba & Transcriptomics & 2 & A versatile method for systematic identification of differential RNA splicing across platforms; A versatile computational method for systematic identification of differential RNA splicing \\
snpsift & Transcriptomics & 1 & MCP wrapper for snpsift \\
sortmerna & Transcriptomics & 3 & MCP wrapper for sortmerna \\
sra-tools & Transcriptomics & 6 & MCP wrapper for sra tools \\
star & Transcriptomics & 4 & An RNA-seq read aligner \\
star-fusion & Transcriptomics & 4 & MCP wrapper for star fusion \\
subread & Transcriptomics & 8 & High-performance read alignment, quantification, and mutation discovery \\
t-coffee & Transcriptomics & 5 & MCP wrapper for t coffee \\
trinity & Transcriptomics & 4 & MCP wrapper for trinity \\
umis & Transcriptomics & 5 & Tools for processing UMI RNA-tag data \\
vardict & Transcriptomics & 1 & Auto-indexed MCP server for vardict \\
vcflib & Transcriptomics & 1 & Auto-indexed MCP server for vcflib \\
zdb & Transcriptomics & 6 & MCP wrapper for zdb \\
kraken2 & Metagenomics & 3 & MCP wrapper for kraken2 \\
metaphlan & Metagenomics & 2 & Metagenomic Phylogenetic Analysis; MetaPhlAn is a computational tool for profiling the composition of microbial \\
metaphlan2 & Metagenomics & 4 & Metagenomic Phylogenetic Analysis; MetaPhlAn is a computational tool for profiling the composition of microbial \\
abnumber & Single-cell & 1 & AbNumber - Antibody numbering using ANARCI \\
abundancebin & Single-cell & 1 & MCP wrapper for abundancebin \\
adapterremoval & Single-cell & 3 & The AdapterRemoval v2 tool for merging and clipping reads \\
anansescanpy & Single-cell & 4 & implementation of scANANSE for scanpy objects in Python \\
anarci & Single-cell & 3 & ANARCI: Antibody Numbering and Antigen Receptor ClassIfication \\
aragorn & Single-cell & 1 & MCP wrapper for aragorn \\
art & Single-cell & 4 & MCP wrapper for art \\
arvados-cwl-runner & Single-cell & 5 & Arvados Common Workflow Language runner \\
arvados-python-client & Single-cell & 5 & Arvados client library; Python API for Arvados, an open source platform for managing and \\
atropos & Single-cell & 1 & trim adapters from high-throughput sequencing reads \\
bactopia & Single-cell & 5 & Bactopia is a flexible pipeline for complete analysis of bacterial genomes \\
bcbio-gff & Single-cell & 4 & A Python library to read and write Generic Feature Format (GFF) \\
bioawk & Single-cell & 1 & MCP wrapper for bioawk \\
biobambam & Single-cell & 11 & Tools for early stage alignment file processing \\
bioconductor-affy & Single-cell & 2 & MCP wrapper for Bioconductor affy \\
bioconductor-affyio & Single-cell & 5 & MCP wrapper for Bioconductor affyio \\
bioconductor-alabaster & Single-cell & 1 & Umbrella for the Alabaster Framework; Umbrella for the alabaster suite, providing a single-line import for all alabaster.* packages. Installing this package ensures that all known alabaster.* packages are also... \\
bioconductor-alabaster.sfe & Single-cell & 2 & Language agnostic on disk serialization of SpatialFeatureExperiment; Builds upon the existing ArtifactDB project, expending alabaster.spatial for language agnostic on disk serialization of... \\
bioconductor-alabaster.spatial & Single-cell & 2 & Save and Load Spatial 'Omics Data to/from File; Save SpatialExperiment objects and their images into file artifacts, and load them back into memory. This is a more portable alternative to serialization of such... \\
bioconductor-ancombc & Single-cell & 2 & Microbiome differential abudance and correlation analyses with bias correction; ANCOMBC is a package containing differential abundance (DA) and correlation analyses for microbiome data. Specifically, the... \\
bioconductor-annotate & Single-cell & 1 & MCP wrapper for Bioconductor annotate \\
bioconductor-annotationdbi & Single-cell & 5 & MCP wrapper for Bioconductor annotationdbi \\
bioconductor-annotationfilter & Single-cell & 6 & Facilities for Filtering Bioconductor Annotation Resources; This package provides class and other infrastructure to implement filters for manipulating Bioconductor annotation resources. The filters will be... \\
bioconductor-annotationhub & Single-cell & 1 & Client to access AnnotationHub resources; This package provides a client for the Bioconductor AnnotationHub web resource. The AnnotationHub web resource provides a central location where genomic files (e.g... \\
bioconductor-apeglm & Single-cell & 1 & MCP wrapper for Bioconductor apeglm \\
bioconductor-apl & Single-cell & 1 & Association Plots; APL is a package developed for computation of Association Plots (AP), a method for visualization and analysis of single cell transcriptomics data. The main focus of APL is the identification... \\
bioconductor-awaggregator & Single-cell & 3 & Attribute-Weighted Aggregation; This package implements an attribute-weighted aggregation algorithm which leverages peptide-spectrum match (PSM) attributes to provide a more accurate estimate of protein... \\
bioconductor-banksy & Single-cell & 4 & Spatial transcriptomic clustering; Banksy is an R package that incorporates spatial information to cluster cells in a feature space (e.g. gene expression). To incorporate spatial information, BANKSY computes... \\
bioconductor-beachmat & Single-cell & 4 & MCP wrapper for Bioconductor beachmat \\
bioconductor-biobase & Single-cell & 2 & MCP wrapper for Bioconductor biobase \\
bioconductor-biocbaseutils & Single-cell & 1 & General utility functions for developing Bioconductor packages; The package provides utility functions related to package development. These include functions that replace slots, and selectors for show... \\
bioconductor-biocfilecache & Single-cell & 9 & Manage Files Across Sessions; This package creates a persistent on-disk cache of files that the user can add, update, and retrieve. It is useful for managing resources (such as custom Txdb objects) that are... \\
bioconductor-biocgenerics & Single-cell & 2 & MCP wrapper for Bioconductor biocgenerics \\
bioconductor-biocio & Single-cell & 2 & Standard Input and Output for Bioconductor Packages; The `BiocIO` package contains high-level abstract classes and generics used by developers to build IO funcionality within the Bioconductor suite of... \\
bioconductor-biocneighbors & Single-cell & 1 & MCP wrapper for Bioconductor biocneighbors \\
bioconductor-biocparallel & Single-cell & 7 & MCP wrapper for Bioconductor biocparallel \\
bioconductor-biocsingular & Single-cell & 3 & MCP wrapper for Bioconductor biocsingular \\
bioconductor-biomart & Single-cell & 3 & MCP wrapper for Bioconductor biomart \\
bioconductor-biomformat & Single-cell & 2 & MCP wrapper for Bioconductor biomformat \\
bioconductor-biostrings & Single-cell & 1 & MCP wrapper for Bioconductor biostrings \\
bioconductor-biovizbase & Single-cell & 7 & Basic graphic utilities for visualization of genomic data; The biovizBase package is designed to provide a set of utilities, color schemes and conventions for genomic data. It serves as the base for various... \\
bioconductor-blase & Single-cell & 1 & Bulk Linking Analysis for Single-cell Experiments; BLASE is a method for finding where bulk RNA-seq data lies on a single-cell pseudotime trajectory. It uses a fast and understandable approach based on... \\
bioconductor-bluster & Single-cell & 5 & Clustering Algorithms for Bioconductor; Wraps common clustering algorithms in an easily extended S4 framework. Backends are implemented for hierarchical, k-means and graph-based clustering. Several utilities... \\
bioconductor-bsgenome & Single-cell & 2 & MCP wrapper for Bioconductor bsgenome \\
bioconductor-bulksignalr & Single-cell & 3 & Infer Ligand-Receptor Interactions from bulk expression (transcriptomics/proteomics) data, or spatial transcriptomics; Inference of ligand-receptor (LR) interactions from bulk expression... \\
bioconductor-cardspa & Single-cell & 1 & Spatially Informed Cell Type Deconvolution for Spatial Transcriptomics; CARD is a reference-based deconvolution method that estimates cell type composition in spatial transcriptomics based on cell type... \\
bioconductor-catscradle & Single-cell & 4 & This package provides methods for analysing spatial transcriptomics data and for discovering gene clusters; This package addresses two broad areas. It allows for in-depth analysis of spatial transcriptomic... \\
bioconductor-cdi & Single-cell & 4 & Clustering Deviation Index (CDI); Single-cell RNA-sequencing (scRNA-seq) is widely used to explore cellular variation. The analysis of scRNA-seq data often starts from clustering cells into subpopulations... \\
bioconductor-cellhashr & Single-cell & 6 & An R package designed to demultiplex cell hashing data \\
bioconductor-cellid & Single-cell & 6 & Unbiased Extraction of Single Cell gene signatures using Multiple Correspondence Analysis; CelliD is a clustering-free multivariate statistical method for the robust extraction of per-cell gene signatures from... \\
bioconductor-cellmigration & Single-cell & 1 & Track Cells, Analyze Cell Trajectories and Compute Migration Statistics; Import TIFF images of fluorescently labeled cells, and track cell movements over time. Parallelization is supported for image processing... \\
bioconductor-clusterfoldsimilarity & Single-cell & 2 & Calculate similarity of clusters from different single cell samples using foldchanges; This package calculates a similarity coefficient using the fold changes of shared features (e.g. genes) among clusters of... \\
bioconductor-clustifyr & Single-cell & 1 & MCP wrapper for Bioconductor clustifyr \\
bioconductor-clustsignal & Single-cell & 4 & ClustSIGNAL: a spatial clustering method; clustSIGNAL: clustering of Spatially Informed Gene expression with Neighbourhood Adapted Learning. A tool for adaptively smoothing and clustering gene expression data... \\
bioconductor-complexheatmap & Single-cell & 2 & Make Complex Heatmaps; Complex heatmaps are efficient to visualize associations between different sources of data sets and reveal potential patterns. Here the ComplexHeatmap package provides a highly flexible... \\
bioconductor-cotan & Single-cell & 1 & COexpression Tables ANalysis; Statistical and computational method to analyze the co-expression of gene pairs at single cell level. It provides the foundation for single-cell gene interactome analysis. The... \\
bioconductor-csoa & Single-cell & 1 & Calculate per-cell gene signature scores in scRNA-seq data using cell set overlaps; Cell Set Overlap Analysis (CSOA) is a tool for calculating per-cell gene signature scores in an scRNA-seq dataset. CSOA... \\
bioconductor-ctsv & Single-cell & 2 & Identification of cell-type-specific spatially variable genes accounting for excess zeros; The R package CTSV implements the CTSV approach developed by Jinge Yu and Xiangyu Luo that detects cell-type-specific... \\
bioconductor-curatedatlasqueryr & Single-cell & 6 & Queries the Human Cell Atlas; Provides access to a copy of the Human Cell Atlas, but with harmonised metadata. This allows for uniform querying across numerous datasets within the Atlas using common fields... \\
bioconductor-cytomapper & Single-cell & 2 & MCP wrapper for Bioconductor cytomapper \\
bioconductor-data-packages & Single-cell & 6 & A package to enable downloading and installation of Bioconductor data packages \\
bioconductor-decipher & Single-cell & 5 & Tools for curating, analyzing, and manipulating biological sequences; A toolset for deciphering and managing biological sequences \\
bioconductor-decontam & Single-cell & 1 & Identify Contaminants in Marker-gene and Metagenomics Sequencing Data; Simple statistical identification of contaminating sequence features in marker-gene or metagenomics data. Works on any kind of feature... \\
bioconductor-decontx & Single-cell & 2 & Decontamination of single cell genomics data; This package contains implementation of DecontX (Yang et al. 2020), a decontamination algorithm for single-cell RNA-seq, and DecontPro (Yin et al. 2023), a... \\
bioconductor-deconvobuddies & Single-cell & 6 & Helper Functions for LIBD Deconvolution; Funtions helpful for LIBD deconvolution project. Includes tools for marker finding with mean ratio, expression plotting, and plotting deconvolution results. Working to... \\
bioconductor-dino & Single-cell & 1 & Normalization of Single-Cell mRNA Sequencing Data; Dino normalizes single-cell, mRNA sequencing data to correct for technical variation, particularly sequencing depth, prior to downstream analysis. The... \\
bioconductor-dirichletmultinomial & Single-cell & 2 & MCP wrapper for Bioconductor dirichletmultinomial \\
bioconductor-dnacopy & Single-cell & 2 & MCP wrapper for Bioconductor dnacopy \\
bioconductor-edger & Single-cell & 2 & MCP wrapper for Bioconductor edger \\
bioconductor-erma & Single-cell & 1 & epigenomic road map adventures; Software and data to support epigenomic road map adventures \\
bioconductor-escher & Single-cell & 4 & Unified multi-dimensional visualizations with Gestalt principles; The creation of effective visualizations is a fundamental component of data analysis. In biomedical research, new challenges are emerging to... \\
bioconductor-experimentsubset & Single-cell & 2 & Manages subsets of data with Bioconductor Experiment objects; Experiment objects such as the SummarizedExperiment or SingleCellExperiment are data containers for one or more matrix-like assays along with the... \\
bioconductor-genefilter & Single-cell & 3 & MCP wrapper for Bioconductor genefilter \\
bioconductor-geneplotter & Single-cell & 3 & MCP wrapper for Bioconductor geneplotter \\
bioconductor-genomeinfodb & Single-cell & 4 & Utilities for manipulating chromosome names, including modifying them to follow a particular naming style; Contains data and functions that define and allow translation between different chromosome sequence... \\
bioconductor-genomeinfodbdata & Single-cell & 2 & MCP wrapper for Bioconductor genomeinfodbdata \\
bioconductor-genomicalignments & Single-cell & 1 & MCP wrapper for Bioconductor genomicalignments \\
bioconductor-genomicfeatures & Single-cell & 1 & MCP wrapper for Bioconductor genomicfeatures \\
bioconductor-geomxtools & Single-cell & 7 & NanoString GeoMx Tools; Tools for NanoString Technologies GeoMx Technology. Package provides functions for reading in DCC and PKC files based on an ExpressionSet derived object. Normalization and QC functions... \\
bioconductor-ggsc & Single-cell & 5 & Visualizing Single Cell and Spatial Transcriptomics; Useful functions to visualize single cell and spatial data. It supports visualizing 'Seurat', 'SingleCellExperiment' and 'SpatialExperiment' objects through... \\
bioconductor-ggspavis & Single-cell & 4 & Visualization functions for spatial transcriptomics data; Visualization functions for spatial transcriptomics data. Includes functions to generate several types of plots, including spot plots, feature... \\
bioconductor-ggtree & Single-cell & 1 & an R package for visualization of tree and annotation data; 'ggtree' extends the 'ggplot2' plotting system which implemented the grammar of graphics. 'ggtree' is designed for visualization and annotation of... \\
bioconductor-glmgampoi & Single-cell & 4 & Fit a Gamma-Poisson Generalized Linear Model; Fit linear models to overdispersed count data. The package can estimate the overdispersion and fit repeated models for matrix input. It is designed to handle large... \\
bioconductor-gosemsim & Single-cell & 7 & MCP wrapper for Bioconductor gosemsim \\
bioconductor-graph & Single-cell & 1 & MCP wrapper for Bioconductor graph \\
bioconductor-hcatonsildata & Single-cell & 1 & Provide programmatic access to the tonsil cell atlas datasets; This package provides access to the scRNA-seq, scATAC-seq, multiome, CITE-seq and spatial transcriptomics (Visium) data generated by the tonsil... \\
bioconductor-hoodscanr & Single-cell & 6 & Spatial cellular neighbourhood scanning in R; hoodscanR is an user-friendly R package providing functions to assist cellular neighborhood analysis of any spatial transcriptomics data with single-cell... \\
bioconductor-humanhippocampus2024 & Single-cell & 2 & Access to SRT and snRNA-seq data from spatial\_HPC project; This is an ExperimentHub Data package that helps to access the spatially-resolved transcriptomics and single-nucleus RNA sequencing data. The datasets... \\
bioconductor-imcdatasets & Single-cell & 2 & Collection of publicly available imaging mass cytometry (IMC) datasets; The imcdatasets package provides access to publicly available IMC datasets. IMC is a technology that enables measurement of > 40 proteins... \\
bioconductor-imcrtools & Single-cell & 4 & Methods for imaging mass cytometry data analysis; This R package supports the handling and analysis of imaging mass cytometry and other highly multiplexed imaging data. The main functionality includes reading... \\
bioconductor-impute & Single-cell & 2 & MCP wrapper for Bioconductor impute \\
bioconductor-infercnv & Single-cell & 3 & MCP wrapper for Bioconductor infercnv \\
bioconductor-interactivedisplaybase & Single-cell & 1 & Base package for enabling powerful shiny web displays of Bioconductor objects; The interactiveDisplayBase package contains the the basic methods needed to generate interactive Shiny based display methods for... \\
bioconductor-iranges & Single-cell & 5 & MCP wrapper for Bioconductor iranges \\
bioconductor-irisfgm & Single-cell & 1 & Comprehensive Analysis of Gene Interactivity Networks Based on Single-Cell RNA-Seq; Single-cell RNA-Seq data is useful in discovering cell heterogeneity and signature genes in specific cell populations in... \\
bioconductor-jazzpanda & Single-cell & 3 & Finding spatially relevant marker genes in image based spatial transcriptomics data; This package contains the function to find marker genes for image-based spatial transcriptomics data. There are functions to... \\
bioconductor-keggrest & Single-cell & 6 & MCP wrapper for Bioconductor keggrest \\
bioconductor-limma & Single-cell & 1 & MCP wrapper for Bioconductor limma \\
bioconductor-lisaclust & Single-cell & 5 & lisaClust: Clustering of Local Indicators of Spatial Association; lisaClust provides a series of functions to identify and visualise regions of tissue where spatial associations between cell-types is similar... \\
bioconductor-matrixgenerics & Single-cell & 2 & S4 Generic Summary Statistic Functions that Operate on Matrix-Like Objects; S4 generic functions modeled after the 'matrixStats' API for alternative matrix implementations. Packages with alternative matrix... \\
bioconductor-merfishdata & Single-cell & 2 & Collection of public MERFISH datasets; MerfishData is an ExperimentHub package that serves publicly available datasets obtained with Multiplexed Error-Robust Fluorescence in situ Hybridization (MERFISH)... \\
bioconductor-metabolomicsworkbenchr & Single-cell & 1 & MCP wrapper for Bioconductor metabolomicsworkbenchr \\
bioconductor-mia & Single-cell & 1 & Microbiome analysis; mia implements tools for microbiome analysis based on the SummarizedExperiment, SingleCellExperiment and TreeSummarizedExperiment infrastructure. Data wrangling and analysis in the context... \\
bioconductor-mousegastrulationdata & Single-cell & 3 & Single-Cell -omics Data across Mouse Gastrulation and Early Organogenesis; Provides processed and raw count data for single-cell RNA sequencing, single-cell ATAC-seq, and seqFISH (spatial transcriptomic)... \\
bioconductor-mspurity & Single-cell & 6 & Automated Evaluation of Precursor Ion Purity for Mass Spectrometry Based Fragmentation in Metabolomics; msPurity R package was developed to: 1) Assess the spectral quality of fragmentation spectra by... \\
bioconductor-multiassayexperiment & Single-cell & 5 & Software for the integration of multi-omics experiments in Bioconductor; Harmonize data management of multiple experimental assays performed on an overlapping set of specimens. It provides a familiar... \\
bioconductor-multtest & Single-cell & 2 & MCP wrapper for Bioconductor multtest \\
bioconductor-nebulosa & Single-cell & 2 & MCP wrapper for Bioconductor nebulosa \\
bioconductor-nnsvg & Single-cell & 1 & Scalable identification of spatially variable genes in spatially-resolved transcriptomics data; Method for scalable identification of spatially variable genes (SVGs) in spatially-resolved transcriptomics data... \\
bioconductor-noiseq & Single-cell & 2 & MCP wrapper for Bioconductor noiseq \\
bioconductor-orfhunter & Single-cell & 5 & Predict open reading frames in nucleotide sequences; The ORFhunteR package is a R and C++ library for an automatic determination and annotation of open reading frames (ORF) in a large set of RNA molecules. It... \\
bioconductor-org.ce.eg.db & Single-cell & 6 & MCP wrapper for Bioconductor org.ce.eg.db \\
bioconductor-org.hs.eg.db & Single-cell & 8 & MCP wrapper for Bioconductor org.hs.eg.db \\
bioconductor-org.mm.eg.db & Single-cell & 5 & MCP wrapper for Bioconductor org.mm.eg.db \\
bioconductor-partcnv & Single-cell & 3 & Infer locally aneuploid cells using single cell RNA-seq data; This package uses a statistical framework for rapid and accurate detection of aneuploid cells with local copy number deletion or amplification. Our... \\
bioconductor-phemd & Single-cell & 1 & MCP wrapper for Bioconductor phemd \\
bioconductor-phyloseq & Single-cell & 5 & Handling and analysis of high-throughput microbiome census data; phyloseq provides a set of classes and tools to facilitate the import, storage, analysis, and graphical display of microbiome census data \\
bioconductor-pipecomp & Single-cell & 1 & pipeComp pipeline benchmarking framework; A simple framework to facilitate the comparison of pipelines involving various steps and parameters. The `pipelineDefinition` class represents pipelines as, minimally... \\
bioconductor-poem & Single-cell & 1 & POpulation-based Evaluation Metrics; This package provides a comprehensive set of external and internal evaluation metrics. It includes metrics for assessing partitions or fuzzy partitions derived from... \\
bioconductor-preprocesscore & Single-cell & 8 & MCP wrapper for Bioconductor preprocesscore \\
bioconductor-proteomicsannotationhubdata & Single-cell & 2 & MCP wrapper for Bioconductor proteomicsannotationhubdata \\
bioconductor-protgenerics & Single-cell & 1 & MCP wrapper for Bioconductor protgenerics \\
bioconductor-qvalue & Single-cell & 2 & Q-value estimation for false discovery rate control; This package takes a list of p-values resulting from the simultaneous testing of many hypotheses and estimates their q-values and local FDR values. The... \\
bioconductor-rbgl & Single-cell & 1 & MCP wrapper for Bioconductor rbgl \\
bioconductor-reactomegsa.data & Single-cell & 1 & Companion data package for the ReactomeGSA package; Companion data sets to showcase the functionality of the ReactomeGSA package. This package contains proteomics and RNA-seq data of the melanoma B-cell... \\
bioconductor-rforproteomics & Single-cell & 1 & MCP wrapper for Bioconductor rforproteomics \\
bioconductor-rgraphviz & Single-cell & 1 & MCP wrapper for Bioconductor rgraphviz \\
bioconductor-rhdf5 & Single-cell & 1 & MCP wrapper for Bioconductor rhdf5 \\
bioconductor-rhdf5filters & Single-cell & 1 & HDF5 Compression Filters; Provides a collection of additional compression filters for HDF5 datasets. The package is intended to provide seemless integration with rhdf5, however the compiled filters can also be... \\
bioconductor-rhdf5lib & Single-cell & 1 & hdf5 library as an R package; Provides C and C++ hdf5 libraries \\
bioconductor-rhtslib & Single-cell & 1 & MCP wrapper for Bioconductor rhtslib \\
bioconductor-rsamtools & Single-cell & 5 & MCP wrapper for Bioconductor rsamtools \\
bioconductor-rtracklayer & Single-cell & 4 & MCP wrapper for Bioconductor rtracklayer \\
bioconductor-s4vectors & Single-cell & 1 & Auto-indexed MCP server for bioconductor-s4vectors \\
bioconductor-scaledmatrix & Single-cell & 6 & Creating a DelayedMatrix of Scaled and Centered Values; Provides delayed computation of a matrix of scaled and centered values. The result is equivalent to using the scale function but avoids explicit... \\
bioconductor-scalign & Single-cell & 1 & An alignment and integration method for single cell genomics; An unsupervised deep learning method for data alignment, integration and estimation of per-cell differences in -omic data (e.g. gene expression)... \\
bioconductor-scarray.sat & Single-cell & 1 & Large-scale single-cell RNA-seq data analysis using GDS files and Seurat; Extends the Seurat classes and functions to support Genomic Data Structure (GDS) files as a DelayedArray backend for data... \\
bioconductor-scater & Single-cell & 1 & Single-Cell Analysis Toolkit for Gene Expression Data in R; A collection of tools for doing various analyses of single-cell RNA-seq gene expression data, with a focus on quality control and visualization \\
bioconductor-scbfa & Single-cell & 2 & MCP wrapper for Bioconductor scbfa \\
bioconductor-scbubbletree & Single-cell & 1 & Quantitative visual exploration of scRNA-seq data; scBubbletree is a quantitative method for the visual exploration of scRNA-seq data, preserving key biological properties such as local and global cell... \\
bioconductor-sccb2 & Single-cell & 1 & MCP wrapper for Bioconductor sccb2 \\
bioconductor-scclassifr & Single-cell & 3 & Pretrained learning models for cell type prediction on single cell RNA-sequencing data; The package comprises a set of pretrained machine learning models to predict basic immune cell types. This enables all... \\
bioconductor-scdataviz & Single-cell & 3 & MCP wrapper for Bioconductor scdataviz \\
bioconductor-scdotplot & Single-cell & 3 & Cluster a Single-cell RNA-seq Dot Plot; Dot plots of single-cell RNA-seq data allow for an examination of the relationships between cell groupings (e.g. clusters) and marker gene expression. The scDotPlot... \\
bioconductor-scider & Single-cell & 5 & Spatial cell-type inter-correlation by density in R; scider is an user-friendly R package providing functions to model the global density of cells in a slide of spatial transcriptomics data. All functions in... \\
bioconductor-scmageck & Single-cell & 4 & MCP wrapper for Bioconductor scmageck \\
bioconductor-scqtltools & Single-cell & 1 & scQTLtools: an R/Bioconductor package for comprehensive identification and visualization of single-cell eQTLs; scQTLtools is a comprehensive R/Bioconductor package that facilitates end-to-end single-cell eQTL... \\
bioconductor-scran & Single-cell & 1 & Methods for Single-Cell RNA-Seq Data Analysis; Implements miscellaneous functions for interpretation of single-cell RNA-seq data. Methods are provided for assignment of cell cycle phase, detection of highly... \\
bioconductor-screpertoire & Single-cell & 1 & MCP wrapper for Bioconductor screpertoire \\
bioconductor-sctreeviz & Single-cell & 2 & R/Bioconductor package to interactively explore and visualize single cell RNA-seq datasets with hierarhical annotations; scTreeViz provides classes to support interactive data aggregation and visualization of... \\
bioconductor-scuttle & Single-cell & 3 & Single-Cell RNA-Seq Analysis Utilities; Provides basic utility functions for performing single-cell analyses, focusing on simple normalization, quality control and data transformations. Also provides some... \\
bioconductor-scvir & Single-cell & 5 & experimental inferface from R to scvi-tools; This package defines interfaces from R to scvi-tools. A vignette works through the totalVI tutorial for analyzing CITE-seq data. Another vignette compares outputs... \\
bioconductor-seraster & Single-cell & 3 & Rasterization Preprocessing Framework for Scalable Spatial Omics Data Analysis; SEraster is a rasterization preprocessing framework that aggregates cellular information into spatial pixels to reduce resource... \\
bioconductor-shortread & Single-cell & 6 & MCP wrapper for Bioconductor shortread \\
bioconductor-singlecellexperiment & Single-cell & 1 & S4 Classes for Single Cell Data; Defines a S4 class for storing data from single-cell experiments. This includes specialized methods to store and retrieve spike-in information, dimensionality reduction... \\
bioconductor-singlecellmultimodal & Single-cell & 1 & MCP wrapper for Bioconductor singlecellmultimodal \\
bioconductor-smoothclust & Single-cell & 1 & smoothclust; Method for identification of spatial domains and spatially-aware clustering in spatial transcriptomics data. The method generates spatial domains with smooth boundaries by smoothing gene... \\
bioconductor-smoppix & Single-cell & 4 & Analyze Single Molecule Spatial Omics Data Using the Probabilistic Index; Test for univariate and bivariate spatial patterns in spatial omics data with single-molecule resolution. The tests implemented allow... \\
bioconductor-sosta & Single-cell & 4 & A package for the analysis of anatomical tissue structures in spatial omics data; sosta (Spatial Omics STructure Analysis) is a package for analyzing spatial omics data to explore tissue organization at the... \\
bioconductor-spacetrooper & Single-cell & 5 & SpaceTrooper performs Quality Control analysis of Image-Based spatial; SpaceTrooper performs Quality Control analysis using data driven GLM models of Image-Based spatial data, providing exploration plots, QC... \\
bioconductor-spaniel & Single-cell & 2 & MCP wrapper for Bioconductor spaniel \\
bioconductor-spari & Single-cell & 3 & Spatially Aware Adjusted Rand Index for Evaluating Spatial Transcritpomics Clustering; The R package used in the manuscript "Spatially Aware Adjusted Rand Index for Evaluating Spatial Transcritpomics... \\
bioconductor-spatialcpie & Single-cell & 1 & MCP wrapper for Bioconductor spatialcpie \\
bioconductor-spatialdatasets & Single-cell & 3 & Collection of spatial omics datasets; This is a collection of publically available spatial omics datasets. Where possible we have curated these datasets as either SpatialExperiments, MoleculeExperiments or... \\
bioconductor-spatialdecon & Single-cell & 1 & MCP wrapper for Bioconductor spatialdecon \\
bioconductor-spatialdmelxsim & Single-cell & 4 & Spatial allelic expression counts for fly cross embryo; Spatial allelic expression counts from Combs \& Fraser (2018), compiled into a SummarizedExperiment object. This package contains data of allelic... \\
bioconductor-spatialexperiment & Single-cell & 2 & MCP wrapper for Bioconductor spatialexperiment \\
bioconductor-spatialfda & Single-cell & 5 & A Tool for Spatial Multi-sample Comparisons; spatialFDA is a package to calculate spatial statistics metrics. The package takes a SpatialExperiment object and calculates spatial statistics metrics using the... \\
bioconductor-spatialfeatureexperiment & Single-cell & 2 & Integrating SpatialExperiment with Simple Features in sf; A new S4 class integrating Simple Features with the R package sf to bring geospatial data analysis methods based on vector data to spatial... \\
bioconductor-spatialheatmap & Single-cell & 5 & MCP wrapper for Bioconductor spatialheatmap \\
bioconductor-spatialomicsoverlay & Single-cell & 1 & Spatial Overlay for Omic Data from Nanostring GeoMx Data; Tools for NanoString Technologies GeoMx Technology. Package to easily graph on top of an OME-TIFF image. Plotting annotations can range from tissue... \\
bioconductor-speckle & Single-cell & 6 & Statistical methods for analysing single cell RNA-seq data; The speckle package contains functions for the analysis of single cell RNA-seq data. The speckle package currently contains functions to analyse... \\
bioconductor-spoon & Single-cell & 1 & Address the Mean-variance Relationship in Spatial Transcriptomics Data; This package addresses the mean-variance relationship in spatially resolved transcriptomics data. Precision weights are generated for... \\
bioconductor-spotclean & Single-cell & 2 & SpotClean adjusts for spot swapping in spatial transcriptomics data; SpotClean is a computational method to adjust for spot swapping in spatial transcriptomics data. Recent spatial transcriptomics experiments... \\
bioconductor-spotsweeper & Single-cell & 4 & Spatially-aware quality control for spatial transcriptomics; Spatially-aware quality control (QC) software for both spot-level and artifact-level QC in spot-based spatial transcripomics, such as 10x Visium... \\
bioconductor-standr & Single-cell & 1 & Spatial transcriptome analyses of Nanostring's DSP data in R; standR is an user-friendly R package providing functions to assist conducting good-practice analysis of Nanostring's GeoMX DSP data. All functions... \\
bioconductor-statial & Single-cell & 5 & A package to identify changes in cell state relative to spatial associations; Statial is a suite of functions for identifying changes in cell state. The functionality provided by Statial provides robust... \\
bioconductor-stjoincount & Single-cell & 4 & stJoincount - Join count statistic for quantifying spatial correlation between clusters; stJoincount facilitates the application of join count analysis to spatial transcriptomic data generated from the 10x... \\
bioconductor-summarizedexperiment & Single-cell & 1 & A container (S4 class) for matrix-like assays; The SummarizedExperiment container contains one or more assays, each represented by a matrix-like object of numeric or other mode. The rows typically represent... \\
bioconductor-tenxvisiumdata & Single-cell & 1 & Visium spatial gene expression data by 10X Genomics; Collection of Visium spatial gene expression datasets by 10X Genomics, formatted into objects of class SpatialExperiment. Data cover various organisms and... \\
bioconductor-tenxxeniumdata & Single-cell & 1 & Collection of Xenium spatial data by 10X genomics; Collection of Xenium spatial transcriptomics datasets provided by 10x Genomics, formatted into the Bioconductor classes, the SpatialExperiment or... \\
bioconductor-tidyomics & Single-cell & 3 & Easily install and load the tidyomics ecosystem; The tidyomics ecosystem is a set of packages for ’omic data analysis that work together in harmony; they share common data representations and API design... \\
bioconductor-tidyspatialexperiment & Single-cell & 2 & SpatialExperiment with tidy principles; tidySpatialExperiment provides a bridge between the SpatialExperiment package and the tidyverse ecosystem. It creates an invisible layer that allows you to interact with... \\
bioconductor-treesummarizedexperiment & Single-cell & 1 & TreeSummarizedExperiment: a S4 Class for Data with Tree Structures; TreeSummarizedExperiment has extended SingleCellExperiment to include hierarchical information on the rows or columns of the rectangular data \\
bioconductor-variantannotation & Single-cell & 3 & MCP wrapper for Bioconductor variantannotation \\
bioconductor-vectrapolarisdata & Single-cell & 4 & Vectra Polaris and Vectra 3 multiplex single-cell imaging data; Provides two multiplex imaging datasets collected on Vectra instruments at the University of Colorado Anschutz Medical Campus. Data are provided... \\
bioconductor-visiumio & Single-cell & 2 & Import Visium data from the 10X Space Ranger pipeline; The package allows users to readily import spatial data obtained from either the 10X website or from the Space Ranger pipeline. Supported formats include... \\
bioconductor-weberdivechalcdata & Single-cell & 2 & Spatially-resolved transcriptomics and single-nucleus RNA-sequencing data from the locus coeruleus (LC) in postmortem human brain samples; Spatially-resolved transcriptomics (SRT) and single-nucleus... \\
bioconductor-xcms & Single-cell & 1 & LC-MS and GC-MS Data Analysis; Framework for processing and visualization of chromatographically separated and single-spectra mass spectral data. Imports from AIA/ANDI NetCDF, mzXML, mzData and mzML files... \\
bioconductor-xenlite & Single-cell & 5 & Simple classes and methods for managing Xenium datasets; Define a relatively light class for managing Xenium data using Bioconductor. Address use of parquet for coordinates, SpatialExperiment for assay and... \\
bioconductor-xvector & Single-cell & 8 & MCP wrapper for Bioconductor xvector \\
bioconductor-zlibbioc & Single-cell & 1 & MCP wrapper for Bioconductor zlibbioc \\
biopython & Single-cell & 1 & Freely available tools for computational molecular biology \\
blast-legacy & Single-cell & 2 & The Basic Local Alignment Search Tool (BLAST) finds regions of local similarity between sequences \\
bpipe & Single-cell & 10 & MCP wrapper for bpipe \\
busco & Single-cell & 4 & Assessment of assembly completeness using Universal Single Copy Orthologs; BUSCO provides measures for quantitative assessment of genome assembly, gene set, and transcriptome completeness based on... \\
bx-python & Single-cell & 5 & Tools for manipulating biological data, particularly multiple sequence alignments \\
c-ares & Single-cell & 2 & c-ares is a C library for asynchronous DNS requests (including name resolves) \\
cascade-reg & Single-cell & 7 & Causal discovery of gene regulatory programs from single-cell genomics; CASCADE stands for Causality-Aware Single-Cell Adaptive \\
cd-hit & Single-cell & 1 & MCP wrapper for cd hit \\
cdbtools & Single-cell & 3 & CDB (Constant DataBase) indexing and retrieval tools for FASTA files \\
cellitac & Single-cell & 1 & Cell type identification using Transcription factor Analysis and Chromatin accessibility \\
cellqc & Single-cell & 1 & Cellqc standardizes the qualiy control of single-cell RNA-Seq (scRNA) data to render clean feature count matrices \\
cellsnake & Single-cell & 5 & cellsnake, a user-friendly tool for single cell RNA sequencing analysis \\
celltypist & Single-cell & 3 & MCP wrapper for celltypist \\
celltypist-so & Single-cell & 2 & Fork of CellTypist without leidenalg in the package requirements \\
checkatlas & Single-cell & 2 & One liner tool to check the quality of your single-cell atlases \\
clustalo & Single-cell & 1 & MCP wrapper for clustalo \\
clustalw & Single-cell & 2 & MCP wrapper for clustalw \\
cmappy & Single-cell & 5 & Assorted tools for interacting with .gct, .gctx, .grp, and .gmt files as well as other Connectivity Map (Broad Institute) data/tools \\
cnmf & Single-cell & 5 & MCP wrapper for cnmf \\
comet-ms & Single-cell & 1 & Comet is an open source tandem mass spectrometry (MS/MS) sequence database search tool \\
constellations & Single-cell & 1 & MCP wrapper for constellations \\
coreutils & Single-cell & 12 & The GNU Core Utilities are the basic file, shell and text manipulation utilities of the GNU operating system. These are the core utilities which are expected to exist on every operating system \\
cosg & Single-cell & 1 & Accurate and fast cell marker gene identification with COSG \\
crisprme & Single-cell & 1 & CRISPRme, tool package for CRISPR experiments assessment and analysis \\
cutadapt & Single-cell & 3 & Trim adapters from high-throughput sequencing reads \\
cwltool & Single-cell & 6 & Common Workflow Language reference implementation \\
cytoscape & Single-cell & 1 & MCP wrapper for cytoscape \\
cytotrace2-python & Single-cell & 1 & MCP wrapper for cytotrace2 python \\
cyvcf2 & Single-cell & 1 & A cython wrapper around htslib built for fast parsing of Variant Call Format (VCF) files \\
dca & Single-cell & 1 & Count autoencoder for scRNA-seq denoising \\
deblur & Single-cell & 2 & Deblur is a greedy deconvolution algorithm based on known read error profiles \\
deeptools & Single-cell & 1 & A set of user-friendly tools for normalization and visualzation of deep-sequencing data \\
deeptoolsintervals & Single-cell & 1 & A python module creating/accessing GTF-based interval trees with associated meta-data \\
delly & Single-cell & 3 & MCP wrapper for delly \\
dendropy & Single-cell & 5 & MCP wrapper for dendropy \\
diamond & Single-cell & 17 & MCP wrapper for diamond \\
disease-gene-qa & Single-cell & 3 & MCP wrapper for disease gene qa \\
dnaio & Single-cell & 5 & Read and write FASTA and FASTQ files efficiently \\
doubletdetection & Single-cell & 3 & Method to detect and enable removal of doublets from single-cell RNA-sequencing \\
dropkick & Single-cell & 2 & Automated scRNA-seq filtering \\
dsh-bio & Single-cell & 2 & MCP wrapper for dsh bio \\
easy\_vitessce & Single-cell & 4 & A package to easily use Vitessce to create interactive plots for single-cell data \\
echidna & Single-cell & 2 & Mapping genotype to phenotype through joint probabilistic modeling of single-cell gene expression and chromosomal copy number variation \\
emboss & Single-cell & 13 & The European Molecular Biology Open Software Suite \\
ena-webin-cli & Single-cell & 2 & MCP wrapper for ena webin cli \\
entrez-direct & Single-cell & 10 & Entrez Direct (EDirect) - Access to NCBI's Entrez databases; Entrez Direct (EDirect) provides access to Entrez, the NCBI's suite of interconnected databases (publication, sequence, structure, gene, variation... \\
epic & Single-cell & 3 & Chip-Seq broad peak/domain finder \\
episcanpy & Single-cell & 1 & Epigenomics Single-Cell Analysis in Python \\
eva-sub-cli & Single-cell & 6 & EVA Submission Command Line Interface \\
f5c & Single-cell & 6 & MCP wrapper for f5c \\
famsa & Single-cell & 1 & MCP wrapper for famsa \\
fast5 & Single-cell & 2 & A C++ header-only library for reading Oxford Nanopore Fast5 files \\
fastani & Single-cell & 1 & FastANI is developed for fast alignment-free computation of whole-genome Average Nucleotide Identity (ANI) \\
fastdtw & Single-cell & 2 & MCP wrapper for fastdtw \\
fastp & Single-cell & 1 & MCP wrapper for fastp \\
fasttree & Single-cell & 1 & MCP wrapper for fasttree \\
fgbio & Single-cell & 11 & MCP wrapper for fgbio \\
filechunkio & Single-cell & 1 & FileChunkIO represents a chunk of an OS-level file containing bytes data \\
flagx & Single-cell & 2 & FLAG-X: FLow cytometry Automated Gating toolboX \\
flye & Single-cell & 1 & MCP wrapper for flye \\
foldseek & Single-cell & 1 & Auto-indexed MCP server for foldseek \\
ftputil & Single-cell & 7 & High-level FTP client library (virtual file system and more) \\
fwdpy11 & Single-cell & 3 & MCP wrapper for fwdpy11 \\
galaxy-lib & Single-cell & 5 & Subset of Galaxy (http: core code base designed to be used a library \\
gatk4-spark & Single-cell & 5 & MCP wrapper for gatk4 spark \\
gdk-pixbuf & Single-cell & 4 & MCP wrapper for gdk pixbuf \\
gecode & Single-cell & 2 & Generic constraint development environment \\
gene-trajectory-python & Single-cell & 1 & Compute gene trajectories; Gene Trajectory is a Python package that computes and analyzes gene trajectories in single-cell data \\
genomepy & Single-cell & 11 & Install and use genomes \& gene annotations the easy way! \\
ghostscript & Single-cell & 7 & An interpreter for the PostScript language and for PDF \\
glimmerhmm & Single-cell & 1 & MCP wrapper for glimmerhmm \\
gneiss & Single-cell & 6 & Compositional data analysis tools and visualizations \\
gnuplot & Single-cell & 1 & MCP wrapper for gnuplot \\
goleft & Single-cell & 6 & MCP wrapper for goleft \\
gridss & Single-cell & 1 & Auto-indexed MCP server for gridss \\
gromacs & Single-cell & 9 & GROMACS is a versatile package to perform molecular dynamics \\
gsmap & Single-cell & 9 & MCP wrapper for gsmap \\
gtdbtk & Single-cell & 8 & MCP wrapper for gtdbtk \\
hicexplorer & Single-cell & 9 & Set of programs to process, analyze and visualize Hi-C and capture Hi-C data \\
hifiasm & Single-cell & 1 & MCP wrapper for hifiasm \\
hmmer & Single-cell & 11 & Biosequence analysis using profile hidden Markov models \\
hyphy & Single-cell & 1 & Auto-indexed MCP server for hyphy \\
igv & Single-cell & 5 & MCP wrapper for igv \\
illumina-interop & Single-cell & 7 & The Illumina InterOp libraries are a set of common routines used for reading and writing InterOp metric files. These metric files are binary files produced during a run providing detailed statistics about a... \\
infernal & Single-cell & 11 & Infernal is for searching DNA sequence databases for RNA structure and sequence similarities; Infernal ("INFERence of RNA ALignment") is for searching DNA sequence databases for RNA structure and sequence... \\
intarna & Single-cell & 1 & Efficient RNA-RNA interaction prediction incorporating seeding and accessibility of interacting sites \\
involucro & Single-cell & 3 & MCP wrapper for involucro \\
iow & Single-cell & 2 & Implementation of Balanced Parentheses; An implementation of the balanced parentheses tree structure as described by \\
itsxpress & Single-cell & 1 & ITSxpress: Software to rapidly trim the Internally Transcribed Spacer (ITS) region from FASTQ files \\
ivar & Single-cell & 7 & MCP wrapper for ivar \\
jalview & Single-cell & 5 & MCP wrapper for jalview \\
java-jdk & Single-cell & 3 & MCP wrapper for java jdk \\
jellyfish & Single-cell & 11 & MCP wrapper for jellyfish \\
k8 & Single-cell & 3 & MCP wrapper for k8 \\
kalign2 & Single-cell & 1 & Kalign is a fast and accurate multiple sequence alignment algorithm designed to align large numbers of protein sequences \\
khipu-metabolomics & Single-cell & 1 & Python library for generalized, low-level annotation of MS metabolomics \\
kma & Single-cell & 1 & Auto-indexed MCP server for kma \\
kmc & Single-cell & 2 & MCP wrapper for kmc \\
kmer-jellyfish & Single-cell & 8 & Jellyfish is a tool for fast, memory-efficient counting of k-mers in DNA. A k-mer is a substring of length k, and counting the occurrences of all such substrings is a central step in many analyses of DNA... \\
krona & Single-cell & 6 & Krona Tools is a set of scripts to create Krona charts from several Bioinformatics tools as well as from text and XML files \\
last & Single-cell & 1 & MCP wrapper for last \\
libcifpp & Single-cell & 5 & Library containing code to manipulate mmCIF and PDB files; This library, libcifpp, is a generic CIF library with some specific additions to work with mmCIF files \\
libdeflate & Single-cell & 4 & libdeflate is a library for fast, whole-buffer DEFLATE-based compression and decompression \\
libsequence & Single-cell & 1 & A C++ class library for evolutionary genetics \\
locarna & Single-cell & 6 & MCP wrapper for locarna \\
macs2 & Single-cell & 12 & Model Based Analysis for ChIP-Seq data \\
mafft & Single-cell & 3 & MCP wrapper for mafft \\
mameshiba & Single-cell & 3 & mameshiba installs only the dependencies needed to run MameShiba; mameshiba is a minimal conda meta-package that installs all dependencies required \\
mappy & Single-cell & 2 & MCP wrapper for mappy \\
markerrepo & Single-cell & 4 & A tool for marker list management and annotation in the single cell context \\
mash & Single-cell & 8 & MCP wrapper for mash \\
maxquant & Single-cell & 3 & MCP wrapper for maxquant \\
mcl & Single-cell & 1 & Auto-indexed MCP server for mcl \\
megahit & Single-cell & 2 & MCP wrapper for megahit \\
meme & Single-cell & 2 & Motif-based sequence analysis tools \\
mentalist & Single-cell & 4 & The MLST pipeline developed by the PathOGiST research group \\
minced & Single-cell & 1 & MCP wrapper for minced \\
miniasm & Single-cell & 2 & MCP wrapper for miniasm \\
minimap2 & Single-cell & 3 & MCP wrapper for minimap2 \\
mitos & Single-cell & 4 & MITOS is a tool for the annotation of metazoan mitochondrial genomes \\
mmtf-python & Single-cell & 2 & A decoding libary for the PDB mmtf format \\
moments & Single-cell & 6 & Evolutionary inference using SFS and LD statistics \\
mosdepth & Single-cell & 1 & MCP wrapper for mosdepth \\
mothur & Single-cell & 5 & MCP wrapper for mothur \\
msproteomicstools & Single-cell & 1 & msproteomicstools is a Python library that can be used in LC-MS/MS based proteomics. It features a core library called \\
msstitch & Single-cell & 22 & MS proteomics post processing utilities \\
multiqc-bcbio & Single-cell & 4 & MultiQC plugin for bcbio report visualization \\
multiqc-xenium-extra & Single-cell & 1 & MultiQC plugin for extra Xenium spatial transcriptomics analysis \\
mummer & Single-cell & 1 & MCP wrapper for mummer \\
munkres & Single-cell & 1 & MCP wrapper for munkres \\
muscle & Single-cell & 2 & MCP wrapper for muscle \\
mysql-connector-c & Single-cell & 1 & MySQL Connector/C, the C interface for communicating with MySQL servers \\
mztosqlite & Single-cell & 1 & MCP wrapper for mztosqlite \\
nanocomp & Single-cell & 2 & Comparing runs of Oxford Nanopore sequencing data and alignments \\
nanoget & Single-cell & 4 & Functions to extract information from Oxford Nanopore sequencing data and alignments \\
nanomath & Single-cell & 3 & A few simple math function for other Oxford Nanopore processing scripts \\
nanoplot & Single-cell & 1 & Plotting suite for long read sequencing data and alignments \\
ncbi-genome-download & Single-cell & 2 & Download genome files from the NCBI FTP server \\
ncbi-ngs-sdk & Single-cell & 1 & NGS is a new, domain-specific API for accessing reads, alignments and pileups produced from Next Generation Sequencing \\
ncbi-vdb & Single-cell & 1 & SRA tools database engine; "VDB is the database engine that all SRA tools use. It is a columnar database \\
ncls & Single-cell & 4 & A fast interval tree-like implementation in C, wrapped for the Python ecosystem. Basically a static interval-tree that is silly fast for both construction and lookups \\
nextalign & Single-cell & 3 & MCP wrapper for nextalign \\
nextclade & Single-cell & 7 & MCP wrapper for nextclade \\
nextflow & Single-cell & 9 & MCP wrapper for nextflow \\
nf-core & Single-cell & 15 & Python package with helper tools for the nf-core community \\
nglview & Single-cell & 6 & An IPython widget to interactively view molecular structures and trajectories. Utilizes the embeddable NGL Viewer for rendering \\
ngs-tools & Single-cell & 1 & Reusable tools for working with next-generation sequencing (NGS) data \\
novae & Single-cell & 4 & Graph-based foundation model for spatial transcriptomics data \\
novoalign & Single-cell & 2 & MCP wrapper for novoalign \\
ont-fast5-api & Single-cell & 5 & Oxford Nanopore Technologies fast5 API software \\
openms-thirdparty & Single-cell & 1 & A helper package to install OpenMS TOPP tools with all their compatible and conda-available versions of adapted thirdparty tools \\
opticlust & Single-cell & 5 & Single cell clustering and recommendations at a glance \\
orthofinder & Single-cell & 1 & MCP wrapper for orthofinder \\
ourotools & Single-cell & 1 & A comprehensive toolkit for quality control and analysis of single-cell long-read RNA-seq data; Ouro-Tools is a novel, comprehensive computational pipeline for long-read scRNA-seq with the following key... \\
palantir & Single-cell & 11 & Palantir for modeling continuous cell state and cell fate choices in single cell data \\
paml & Single-cell & 5 & A package of programs for phylogenetic analyses of DNA or protein sequences using maximum likelihood \\
pango-designation & Single-cell & 1 & MCP wrapper for pango designation \\
pangolearn & Single-cell & 2 & Store of the trained model for pangolin to access \\
pangolin & Single-cell & 1 & Phylogenetic Assignment of Named Global Outbreak LINeages \\
parasail-python & Single-cell & 1 & Python bindings for the parasail C library containing implementations of pairwise sequence alignment algorithms \\
pasta & Single-cell & 1 & MCP wrapper for pasta \\
paste-bio & Single-cell & 2 & A computational method to align and integrate spatial transcriptomics experiments \\
pcdl & Single-cell & 18 & physicell data loader (pcdl) provides a platform independent, python3 based, pip installable interface to transform output, generated with the PhysiCell agent based modeling framework, into standard formats \\
peakqc & Single-cell & 1 & Quality control of single cell ATAC-seq data based on fragment length distributions \\
perl-alien-libxml2 & Single-cell & 1 & Installs the C libxml2 library on your system \\
perl-bio-coordinate & Single-cell & 3 & Methods for dealing with genomic coordinates \\
perl-bio-searchio-hmmer & Single-cell & 2 & A parser for HMMER2 and HMMER3 output (hmmscan, hmmsearch, hmmpfam) \\
perl-bio-tools-phylo-paml & Single-cell & 5 & Parses output from the PAML programs codeml, baseml, basemlg, codemlsites and yn00 \\
perl-bio-tools-run-alignment-clustalw & Single-cell & 1 & Object for the calculation of a multiple sequence alignment from a set of unaligned sequences or alignments using the Clustalw program \\
perl-business-isbn-data & Single-cell & 7 & data pack for Business::ISBN \\
perl-cpan-meta-requirements & Single-cell & 4 & A set of version requirements for a CPAN dist \\
perl-devel-overloadinfo & Single-cell & 1 & introspect overloaded operators \\
perl-digest-sha1 & Single-cell & 3 & Perl interface to the SHA-1 algorithm \\
perl-ffi-checklib & Single-cell & 10 & Check that a library is available for FFI \\
perl-ipc-sharelite & Single-cell & 1 & Lightweight interface to shared memory \\
perl-json-pp & Single-cell & 2 & JSON::XS compatible pure-Perl module \\
perl-module-runtime-conflicts & Single-cell & 1 & Provide information on conflicts for Module::Runtime \\
perl-package-deprecationmanager & Single-cell & 1 & Manage deprecation warnings for your distribution \\
perl-sub-name & Single-cell & 1 & MCP wrapper for perl sub name \\
perl-sub-quote & Single-cell & 8 & Efficient generation of subroutines via string eval \\
perl-url-encode & Single-cell & 6 & Encoding and decoding of application/x-www-form-urlencoded encoding \\
perl-version & Single-cell & 7 & Structured version objects \\
perl-xml-namespacesupport & Single-cell & 1 & MCP wrapper for perl xml namespacesupport \\
phyml & Single-cell & 1 & MCP wrapper for phyml \\
piaso & Single-cell & 2 & PIASO: Precise Integrative Analysis of Single-cell Omics; PIASO is a Python toolkit for precise integrative analysis of single-cell omics data \\
picard & Single-cell & 10 & MCP wrapper for picard \\
pixelator & Single-cell & 10 & A command-line tool and library to process and analyze sequencing data from Molecular Pixelation (MPX) assays \\
plink & Single-cell & 1 & Auto-indexed MCP server for plink \\
portcullis & Single-cell & 9 & Splice junction analysis and filtering from BAM files \\
pplacer & Single-cell & 1 & Auto-indexed MCP server for pplacer \\
prank & Single-cell & 4 & MCP wrapper for prank \\
primer3 & Single-cell & 5 & Design PCR primers from DNA sequence. From mispriming libraries to sequence quality data to the generation of internal oligos, primer3 does it \\
prophyle & Single-cell & 8 & ProPhyle is an accurate, resource-frugal and deterministic phylogeny-based metagenomic classifier \\
proteomiqon-peptidedb & Single-cell & 1 & MCP wrapper for proteomiqon peptidedb \\
proteomiqon-peptidespectrummatching & Single-cell & 1 & MCP wrapper for proteomiqon peptidespectrummatching \\
pubchempy & Single-cell & 6 & MCP wrapper for pubchempy \\
py2bit & Single-cell & 5 & MCP wrapper for py2bit \\
pybigwig & Single-cell & 7 & A python extension written in C for quick access to bigWig files \\
pycistopic & Single-cell & 6 & MCP wrapper for pycistopic \\
pyfastx & Single-cell & 7 & MCP wrapper for pyfastx \\
pyhmmer & Single-cell & 6 & Cython bindings and Python interface to HMMER3 \\
pyrovelocity & Single-cell & 1 & Probabilistic RNA velocity for cell fate uncertainty estimation \\
pysftp & Single-cell & 6 & A friendly face on SFTP \\
pyspoa & Single-cell & 1 & Python binding to spoa library \\
pytabix & Single-cell & 3 & Fast random access to sorted files compressed with bgzip and indexed by tabix \\
pyteomics & Single-cell & 8 & A framework for proteomics data analysis \\
python-edlib & Single-cell & 1 & Lightweight, super fast C/C++ (\& Python) library for sequence alignment using edit (Levenshtein) distance \\
qcatch & Single-cell & 2 & QCatch: Quality Control downstream of alevin-fry / simpleaf \\
r-abdiv & Single-cell & 10 & Alpha and Beta Diversity Measures; 'A collection of measures for measuring ecological diversity \\
r-acidbase & Single-cell & 11 & Low-level base functions imported by Acid Genomics packages \\
r-azimuth & Single-cell & 1 & Azimuth is a Shiny app demonstrating a query-reference mapping algorithm for single-cell data \\
r-basejump & Single-cell & 1 & Base functions for bioinformatics and R package development \\
r-beyondcell & Single-cell & 7 & Tool for the Analysis of tumour therapeutic heterogeneity in single-cell RNA-seq; Beyondcell is a methodology for the identification of drug vulnerabilities in single-cell RNA-seq (scRNA-seq) data. To this... \\
r-cdseq & Single-cell & 3 & Estimate cell-type-specific gene expression profiles and sample-specific cell-type proportions simultaneously using bulk sequencing data. Kang et al. (2019) <doi:10.1371/journal.pcbi.1007510> \\
r-dwls & Single-cell & 6 & Deconvolution of bulk mRNA data using single-cell RNAseq to provide cell type specific signatures \\
r-epitrace & Single-cell & 6 & Inference of cell age and phylogeny from single cell ATAC data \\
r-goalie & Single-cell & 11 & Assertive check functions for defensive R programming \\
r-mams & Single-cell & 1 & R package for Matrix and Analysis Metadata Standards \\
r-redeemr & Single-cell & 1 & R package for Regulatory multi-omics with Deep Mitochondrial mutation profiling; Introduce a new approach for single-cell Regulatory multi-omics (transcriptomics and chromatin accessibility) with Deep... \\
r-restfulr & Single-cell & 5 & Models a RESTful service as if it were a nested R list \\
r-saige & Single-cell & 1 & SAIGE is an R package with Scalable and Accurate Implementation of Generalized mixed model (Chen, H. et al. 2016); SAIGE is an R package with Scalable and Accurate Implementation of Generalized \\
r-sceasy & Single-cell & 1 & A package providing functions to convert between different single-cell data formats \\
r-scopfunctions & Single-cell & 8 & An R package of functions for single cell -omics analysis \\
r-scpred & Single-cell & 3 & MCP wrapper for r scpred \\
r-seurat & Single-cell & 9 & MCP wrapper for r seurat \\
r-seurat-data & Single-cell & 5 & Single cell RNA sequencing datasets can be large, consisting of matrices that contain expression data for several thousand features across several thousand cells. This package is designed to easily install... \\
r-seurat-disk & Single-cell & 4 & The h5Seurat file format is specifically designed for the storage and analysis of multi-modal single-cell and spatially-resolved expression experiments, for example, from CITE-seq or 10X Visium technologies... \\
r-seurat-scripts & Single-cell & 1 & MCP wrapper for r seurat scripts \\
r-signac & Single-cell & 3 & MCP wrapper for r signac \\
raxml & Single-cell & 5 & Phylogenetics - Randomized Axelerated Maximum Likelihood \\
repeatmasker & Single-cell & 2 & RepeatMasker is a program that screens DNA sequences for interspersed repeats and low complexity DNA sequences \\
rnastructure & Single-cell & 10 & RNAstructure is a complete package for RNA and DNA secondary structure prediction and analysis. It includes algorithms for secondary structure prediction, including facility to predict base pairing... \\
rpsbproc & Single-cell & 1 & Auto-indexed MCP server for rpsbproc \\
rsa & Single-cell & 6 & Pure-Python RSA implementation \\
rseqc & Single-cell & 18 & QC package for RNA-seq data \\
samap & Single-cell & 3 & The SAMap algorithm \\
sambamba & Single-cell & 11 & MCP wrapper for sambamba \\
samsift & Single-cell & 1 & Advanced filtering and tagging of SAM/BAM alignments using Python expressions \\
samtools & Single-cell & 29 & MCP wrapper for samtools \\
sc-musketeers & Single-cell & 1 & A tri-partite modular autoencoder for addressing imbalanced cell type annotation and batch effect reduction \\
scaden & Single-cell & 6 & Cell type deconvolution using single cell data \\
scanpy & Single-cell & 1 & MCP wrapper for scanpy \\
scanpy-cli & Single-cell & 13 & CLI for Scanpy \\
scanpy-scripts & Single-cell & 16 & Scripts for using scanpy from the command line \\
scar & Single-cell & 1 & MCP wrapper for scar \\
scarches & Single-cell & 6 & MCP wrapper for scarches \\
sccaf & Single-cell & 1 & MCP wrapper for sccaf \\
scepia & Single-cell & 2 & Single Cell Epigenome-based Inference of Activity \\
scglue & Single-cell & 11 & Graph-linked unified embedding for unpaired single-cell multi-omics data integration; GLUE is a flexible framework that utilizes prior knowledge about feature \\
scib & Single-cell & 1 & Evaluating single-cell data integration methods \\
scirpy & Single-cell & 5 & A Scanpy extension for analyzing single-cell T-cell and B-cell receptor (TCR/BCR) sequencing data \\
scmeta & Single-cell & 8 & MCP wrapper for scmeta \\
scmidas & Single-cell & 3 & A torch-based integration method for single-cell multi-omic data \\
scpred-cli & Single-cell & 2 & MCP wrapper for scpred cli \\
scrnasim-toolz & Single-cell & 7 & Tools used by scRNAsim workflow \\
scspectra & Single-cell & 6 & Supervised discovery of interpretable gene programs from single-cell data \\
scstem & Single-cell & 1 & A method for mapping single-cell and spatial transcriptomics data with transfer learning \\
scvelo & Single-cell & 1 & MCP wrapper for scvelo \\
scvi & Single-cell & 7 & Single-cell Variational Inference \\
scvi-tools & Single-cell & 1 & Deep probabilistic analysis of single-cell omics data \\
scvis & Single-cell & 2 & scvis is a python package for dimension reduction of high-dimensional biological data, especially single-cell RNA-sequencing (scRNA-seq) data \\
scvis\_galaxy & Single-cell & 3 & scvis is a python package for dimension reduction of high-dimensional biological data, especially single-cell RNA-sequencing (scRNA-seq) data \\
scxmatch & Single-cell & 1 & Python implementation for single-cell cross match test, an efficient implementation of Rosenbaum's test \\
sdeper & Single-cell & 2 & Spatial Deconvolution method with Platform Effect Removal; SDePER (Spatial Deconvolution method with Platform Effect Removal) is a hybrid machine learning and regression method to deconvolve Spatial... \\
seacells & Single-cell & 5 & SEACells algorithm for Inference of transcriptional and epigenomic cellular states from single-cell genomics data \\
searchgui & Single-cell & 4 & MCP wrapper for searchgui \\
segemehl & Single-cell & 2 & Short read mapping with gaps \\
sentieon & Single-cell & 1 & Accelerated performance bioinformatics tools for mapping and variant calling \\
sepp & Single-cell & 3 & SATe-enabled phylogenetic placement \\
seqcluster & Single-cell & 1 & small RNA analysis from NGS data \\
seqkit & Single-cell & 27 & MCP wrapper for seqkit \\
seqtk & Single-cell & 15 & MCP wrapper for seqtk \\
sequence-operations & Single-cell & 7 & MCP wrapper for sequence operations \\
seurat-scripts & Single-cell & 11 & MCP wrapper for seurat scripts \\
sevenbridges-python & Single-cell & 4 & SBG API python client bindings; sevenbridges-python is a Python library that provides an interface for the Seven Bridges Platform the Cancer Genomics Cloud and Cavatica public APIs. It works with Python... \\
simo-omics & Single-cell & 1 & Spatial integration of multi-omics single-cell datasets through probabilistic alignment \\
simplejson & Single-cell & 2 & Simple, fast, extensible JSON encoder/decoder for Python \\
sina & Single-cell & 3 & MCP wrapper for sina \\
sincei & Single-cell & 6 & A user-friendly toolkit for QC, counting, clustering and plotting of single-cell (epi)genomics data \\
slow5tools & Single-cell & 11 & MCP wrapper for slow5tools \\
snakemake & Single-cell & 6 & A popular workflow management system aiming at full in-silico reproducibility; Snakemake is a workflow management system that aims to reduce the complexity of creating \\
snakemake-interface-executor-plugins & Single-cell & 1 & This package provides a stable interface for interactions between Snakemake and its executor plugins \\
snakemake-interface-report-plugins & Single-cell & 5 & This package provides a stable interface for interactions between Snakemake and its report plugins \\
snakemake-interface-storage-plugins & Single-cell & 5 & This package provides a stable interface for interactions between Snakemake and its storage plugins \\
snakemake-minimal & Single-cell & 8 & A popular workflow management system aiming at full in-silico reproducibility; Snakemake is a workflow management system that aims to reduce the complexity \\
snp2cell & Single-cell & 8 & MCP wrapper for snp2cell \\
sopa & Single-cell & 8 & Spatial-omics pipeline and analysis \\
sourmash & Single-cell & 8 & Quickly search, compare, and analyze genomic and metagenomic data sets \\
spades & Single-cell & 10 & SPAdes (St. Petersburg genome assembler) is intended for both standard isolates and single-cell MDA bacteria assemblies; SPAdes (St. Petersburg genome assembler) is a genome assembly algorithm which was... \\
spagrn & Single-cell & 1 & A comprehensive tool to infer TF-centered, spatial gene regulatory networks for the spatially resolved transcriptomics (SRT) data \\
spapros & Single-cell & 5 & Probe set selection for targeted spatial transcriptomics \\
spatialleiden & Single-cell & 1 & Implementation of multiplex Leiden for analysis of spatial omics data \\
stacks & Single-cell & 15 & MCP wrapper for stacks \\
stream & Single-cell & 8 & STREAM-Single-cell Trajectories Reconstruction, Exploration And Mapping \\
switchtfi & Single-cell & 1 & Implementation of the SwitchTFI method as presented in: https: \\
t\_coffee & Single-cell & 5 & A collection of tools for Computing, Evaluating and Manipulating Multiple Alignments of DNA, RNA, Protein Sequences and Structures \\
taxonkit & Single-cell & 12 & MCP wrapper for taxonkit \\
tidyp & Single-cell & 2 & MCP wrapper for tidyp \\
tmalign & Single-cell & 5 & TM-align sequence-order independent protein structure alignment \\
tobias & Single-cell & 1 & Transcription factor Occupancy prediction By Investigation of ATAC-seq Signal; TOBIAS (Transcription factor Occupancy prediction By Investigation of ATAC-seq Signal) is a collection \\
toil & Single-cell & 1 & A scalable, efficient, cross-platform and easy-to-use workflow engine in pure Python \\
transdecoder & Single-cell & 1 & Auto-indexed MCP server for transdecoder \\
treetime & Single-cell & 5 & Maximum-Likelihood dating and ancestral inference for phylogenetic trees \\
trf & Single-cell & 1 & Auto-indexed MCP server for trf \\
trim-galore & Single-cell & 3 & Trim Galore! is a wrapper script to automate quality and adapter trimming as well as quality control \\
trimadap & Single-cell & 1 & Fast but inaccurate adapter trimmer for Illumina reads \\
trimmomatic & Single-cell & 2 & MCP wrapper for trimmomatic \\
trnascan-se & Single-cell & 4 & tRNA detection in large-scale genomic sequences \\
ucsc-bedgraphtobigwig & Single-cell & 1 & Convert a bedGraph file to bigWig format \\
ucsc-bedtobigbed & Single-cell & 1 & Convert bed file to bigBed. (BigBed version: 4) \\
ucsc-cell-browser & Single-cell & 10 & A browser for single-cell data, main site at UCSC Cellbrowser, an interactive browser for single cell data. Includes importers and basic pipelines for text files, Seurat, Scanpy and Cellranger. All Javascript... \\
ucsc-fatotwobit & Single-cell & 1 & Convert DNA from fasta to 2bit format \\
ucsc-gtftogenepred & Single-cell & 1 & Convert a GTF file to a genePred \\
ucsc-liftover & Single-cell & 2 & Move annotations from one assembly to another \\
ucsc-nibfrag & Single-cell & 1 & Extract part of a nib file as .fa (all bases/gaps lower case by default) \\
ucsc-twobitinfo & Single-cell & 1 & Get information about sequences in a .2bit file \\
ucsc-twobittofa & Single-cell & 1 & Convert all or part of .2bit file to fasta \\
ucsc-wigtobigwig & Single-cell & 1 & Convert ascii format wig file (in fixedStep, variableStep) \\
umi\_tools & Single-cell & 1 & Tools for dealing with Unique Molecular Identifiers (UMIs) / Random Molecular Tags (RMTs) \\
unicycler & Single-cell & 3 & MCP wrapper for unicycler \\
unifrac & Single-cell & 2 & Fast phylogenetic diversity calculations; UniFrac is a commonly phylogenetic diversity distance metric used in \\
unifrac-binaries & Single-cell & 1 & Fast phylogenetic diversity calculations; UniFrac is a commonly phylogenetic diversity distance metric used in \\
urllib3 & Single-cell & 4 & HTTP library with thread-safe connection pooling, file post, and more \\
usher & Single-cell & 4 & Ultrafast Sample Placement on Existing Trees (UShER) \\
varlociraptor & Single-cell & 1 & Auto-indexed MCP server for varlociraptor \\
varscan & Single-cell & 1 & Auto-indexed MCP server for varscan \\
vcfpy & Single-cell & 6 & Python 3 VCF library with good support for both reading and writing \\
vcontact2 & Single-cell & 4 & Viral Contig Automatic Clustering and Taxonomy \\
viennarna & Single-cell & 23 & ViennaRNA package -- RNA secondary structure prediction and comparison \\
vitessce-python & Single-cell & 1 & Jupyter widget facilitating interactive visualization of spatial single-cell data with Vitessce \\
vpt & Single-cell & 11 & Command line tool for highly parallelized processing of Vizgen data \\
vt & Single-cell & 11 & A tool set for short variant discovery in genetic sequence data \\
womtool & Single-cell & 6 & MCP wrapper for womtool \\
xclone & Single-cell & 5 & Inference of clonal Copy Number Alterations in single cells \\
yacrd & Single-cell & 5 & MCP wrapper for yacrd \\
bioconductor-clusterprofiler & Pathway enrich. & 5 & A universal enrichment tool for interpreting omics data; This package supports functional characteristics of both coding and non-coding genomics data for thousands of species with up-to-date gene annotation... \\
bioconductor-dose & Pathway enrich. & 6 & Disease Ontology Semantic and Enrichment analysis; This package implements five methods proposed by Resnik, Schlicker, Jiang, Lin and Wang respectively for measuring semantic similarities among DO terms and... \\
bioconductor-fgsea & Pathway enrich. & 3 & Fast Gene Set Enrichment Analysis; The package implements an algorithm for fast gene set enrichment analysis. Using the fast algorithm allows to make more permutations and get more fine grained p-values, which... \\
bioconductor-go.db & Pathway enrich. & 2 & MCP wrapper for Bioconductor go.db \\
bioconductor-gsva & Pathway enrich. & 1 & MCP wrapper for Bioconductor gsva \\
bioconductor-mastr & Pathway enrich. & 3 & Markers Automated Screening Tool in R; mastR is an R package designed for automated screening of signatures of interest for specific research questions. The package is developed for generating refined lists of... \\
bioconductor-regionalst & Pathway enrich. & 4 & Investigating regions of interest and performing regional cell type-specific analysis with spatial transcriptomics data; This package analyze spatial transcriptomics data through cross-regional cell... \\
bioconductor-scfeatures & Pathway enrich. & 2 & scFeatures: Multi-view representations of single-cell and spatial data for disease outcome prediction; scFeatures constructs multi-view representations of single-cell and spatial data. scFeatures is a tool... \\
bioconductor-signifinder & Pathway enrich. & 1 & Collection and implementation of public transcriptional cancer signatures; signifinder is an R package for computing and exploring a compendium of tumor signatures. It allows to compute a variety of signatures... \\
cell2cell & Pathway enrich. & 2 & Inferring cell-cell interactions from transcriptomes with cell2cell \\
cospar & Pathway enrich. & 6 & A toolkit for dynamic inference of cell fate by integrating state and lineage information \\
decoupler & Pathway enrich. & 7 & MCP wrapper for decoupler \\
genecircuitry & Pathway enrich. & 2 & GeneCircuitry: TRN analysis from single-cell data (Scanpy, CellOracle, Hotspot); A modular, checkpoint-enabled pipeline for TRN analysis from \\
gseapy & Pathway enrich. & 5 & Gene Set Enrichment Analysis in Python \\
mageck & Pathway enrich. & 6 & MCP wrapper for mageck \\
sccellfie & Pathway enrich. & 4 & A tool for inferring metabolic activities from single-cell and spatial transcriptomics \\
sctriangulate & Pathway enrich. & 1 & A Python package to mix-and-match conflicting clustering results in single cell analysis, and generate reconciled clustering solutions \\
tirank & Pathway enrich. & 2 & A comprehensive analysis tool for transferring phenotype of bulk transcriptomic data to single-cell or spatial transcriptomic data; TiRank integrates deep learning and statistical analysis to infer phenotype \\
sage-proteomics & Proteomics & 1 & Proteomics searching so fast it feels like magic \\
bioconductor-moleculeexperiment & Utility & 7 & Prioritising a molecule-level storage of Spatial Transcriptomics Data; MoleculeExperiment contains functions to create and work with objects from the new MoleculeExperiment class. We introduce this class for... \\
csvtk & Utility & 40 & MCP wrapper for csvtk \\
dxpy & Utility & 12 & DNAnexus Platform API bindings for Python." \\
ensembl-vep & Utility & 3 & Ensembl Variant Effect Predictor; The VEP determines the effect of your variants (SNPs, insertions, deletions, CNVs or structural variants) on genes, transcripts, and protein sequence, as well as regulatory... \\
jq & Utility & 1 & MCP wrapper for jq \\
multivelo & Utility & 1 & MCP wrapper for multivelo \\
perl-spreadsheet-parseexcel & Utility & 4 & MCP wrapper for perl spreadsheet parseexcel \\
\end{longtable}
\endgroup

% Removed copied duplicate servers: bioconductor-affyio copy, bioconductor-rtracklayer copy, csvtk copy, diamond copy, picard copy
% If additional appendix sections follow and the paper should return to two-column layout, uncomment the next line.
% \twocolumn
\end{document}